\documentclass[10pt,twocolumn,letterpaper]{article}
\usepackage[pagenumbers]{cvpr}

\usepackage{graphicx}
\usepackage{amsmath,amssymb}
\usepackage{booktabs}
\usepackage{tabularx}
\newcolumntype{Y}{>{\centering\arraybackslash}X}   %
\usepackage{multirow}
\usepackage[table]{xcolor}
\usepackage{capt-of}
\usepackage{xspace}
\usepackage{microtype}
\usepackage{placeins}
\usepackage{algorithm}
\usepackage{algpseudocode}
\definecolor{oursrow}{HTML}{EAF2FC}
\renewcommand{\topfraction}{0.9}
\renewcommand{\dbltopfraction}{0.9}
\renewcommand{\textfraction}{0.1}
\renewcommand{\floatpagefraction}{0.85}
\renewcommand{\dblfloatpagefraction}{0.85}
\definecolor{cvprblue}{rgb}{0.21,0.49,0.74}
\usepackage[pagebackref,breaklinks,colorlinks,allcolors=cvprblue]{hyperref}
\usepackage[capitalize]{cleveref}
\crefname{section}{Sec.}{Secs.}
\Crefname{section}{Section}{Sections}
\crefname{table}{Tab.}{Tabs.}
\Crefname{table}{Table}{Tables}
\hypersetup{pdftitle={When \textquotedblleft A or B\textquotedblright\space Is Less Likely than A: Coherent Probabilities for Vision Decision Models},pdfauthor={Seongrae Noh, SeongEun Hong, HyeongYeop Kang}}
\makeatletter
\let\suppmaketitle@cvpr\@maketitle   %
\newcommand{\supptoc}{\begingroup\def\numberline##1\hskip -1em.~{\hbox to\@tempdima{##1.\hfil}}\tableofcontents\endgroup}
\makeatother

\newcommand{\method}{\textsc{Core}\xspace}

\DeclareMathOperator{\KL}{KL}

\newcommand{\NBaseAcc}{60.8}

\newcommand{\NBaseEce}{5.6}
\newcommand{\NBasePart}{14.8}
\newcommand{\NBaseHard}{25.7}

\newcommand{\NOursAcc}{67.1}

\newcommand{\NOursEce}{1.7}
\newcommand{\NOursPart}{5.8}
\newcommand{\NOursHard}{8.5}

\newcommand{\NOursBrier}{42.4}
\newcommand{\NSftAcc}{66.6}

\newcommand{\NSftPart}{19.3}

\newcommand{\NObsEce}{6.4}
\newcommand{\NObsPart}{11.6}

\newcommand{\NTsEce}{1.9}
\newcommand{\NTsPart}{14.6}

\newcommand{\NLjevDPart}{9.4}
\newcommand{\NLjevDHard}{12.2}

\newcommand{\NLjevDBrier}{40.7}
\newcommand{\NOursSeedsLite}{5}
\newcommand{\NAccGapLjevD}{1.8}
\newcommand{\NAccGapSftD}{2.1}

\newcommand{\NPartMargin}{3.5}
\newcommand{\NHardMargin}{3.6}
\newcommand{\NBestPubPartDrop}{0.4}
\newcommand{\NTfMaxDrop}{0.9}

\newcommand{\NFlTfMaxDrop}{1.1}
\newcommand{\NFlObsPart}{11.6}

\newcommand{\NFlObsEce}{11.7}

\newcommand{\NAccGain}{6.3}
\newcommand{\NSftAccGain}{5.8}

\newcommand{\NGapAbsBaseLite}{11.5}
\newcommand{\NMonoBaseLite}{31}

\newcommand{\NGapAbsSftLite}{12.3}

\newcommand{\NGapAbsLjevLite}{11.3}

\newcommand{\NGapAbsBaseNano}{9.0}

\newcommand{\NGapAbsBaseMini}{9.3}

\newcommand{\NGapAbsBaseFlash}{8.7}

\newcommand{\NGapAbsBaseBig}{5.4}

\newcommand{\NFlBaseEce}{3.5}
\newcommand{\NFlBasePart}{15.0}
\newcommand{\NFlBaseHard}{35.7}

\newcommand{\NFlOursAcc}{77.2}

\newcommand{\NFlOursEce}{2.3}
\newcommand{\NFlOursPart}{4.8}
\newcommand{\NFlOursHard}{9.2}

\newcommand{\NFlOursBrier}{30.2}

\newcommand{\NFlLjevAcc}{78.0}

\newcommand{\NFlLjevPart}{11.5}

\newcommand{\NFlLjevBrier}{29.3}
\newcommand{\NFlSftDAcc}{80.7}

\newcommand{\NFlSftDEce}{12.0}
\newcommand{\NFlSftDPart}{7.3}
\newcommand{\NFlSftDHard}{14.9}

\newcommand{\NFlLjevDPart}{12.1}

\newcommand{\NFlLjevDBrier}{28.1}

\newcommand{\NFlExactAcc}{77.4}

\newcommand{\NFlExactPart}{4.6}

\newcommand{\NFlAccGain}{4.6}

\newcommand{\NBigBaseEce}{1.7}

\newcommand{\NBigOursEce}{3.2}

\newcommand{\NMiniBaseHard}{38.3}

\newcommand{\NAblAnchorAcc}{66.9}

\newcommand{\NAblAnchorPart}{5.8}

\newcommand{\NAblUAcc}{67.1}

\newcommand{\NAblUPart}{6.1}

\newcommand{\NAblWzeroAcc}{64.3}

\newcommand{\NAblWzeroPart}{5.5}

\newcommand{\NAblWhalfAcc}{68.6}
\newcommand{\NAblWhalfEce}{2.5}
\newcommand{\NAblWhalfPart}{6.4}

\newcommand{\NAblWoneAcc}{68.2}
\newcommand{\NAblWoneEce}{13.3}
\newcommand{\NAblWonePart}{11.8}

\newcommand{\NAblSchedAcc}{68.2}
\newcommand{\NAblSchedEce}{3.2}

\newcommand{\NAblExactAcc}{67.3}

\newcommand{\NAblExactPart}{5.8}

\newcommand{\NAblCeAcc}{67.8}

\newcommand{\NAblCePart}{5.7}

\newcommand{\NSizeBasePartMin}{10.2}
\newcommand{\NSizeBasePartMax}{16.4}

\newcommand{\NSizeOursPartMin}{4.1}
\newcommand{\NSizeOursPartMax}{5.8}
\newcommand{\NSizeOursHardMin}{8.1}
\newcommand{\NSizeOursHardMax}{10.2}
\newcommand{\NSizeAccGainMin}{3.2}
\newcommand{\NSizeAccGainMax}{6.3}

\newcommand{\NWordBaseOneof}{14.1}

\newcommand{\NWordBaseAny}{15.1}
\newcommand{\NWordBaseSlash}{17.9}
\newcommand{\NWordOursOr}{5.8}
\newcommand{\NWordOursOneof}{6.6}
\newcommand{\NWordOursEither}{6.7}
\newcommand{\NWordOursAny}{7.0}
\newcommand{\NWordOursSlash}{11.7}
\newcommand{\NWordRedSlash}{35}
\newcommand{\NWordRedMin}{53}
\newcommand{\NWordRedMax}{61}
\newcommand{\NAokBaseAcc}{60.8}

\newcommand{\NAokBasePart}{14.8}
\newcommand{\NAokBaseHard}{25.7}

\newcommand{\NAokSftAcc}{61.9}

\newcommand{\NAokSftEce}{17.5}
\newcommand{\NAokSftPart}{12.0}

\newcommand{\NAokOursAcc}{62.2}

\newcommand{\NAokOursEce}{2.4}
\newcommand{\NAokOursPart}{8.3}
\newcommand{\NAokOursHard}{14.8}

\newcommand{\NVisBaseBlankPart}{12.6}
\newcommand{\NVisBaseSwapAcc}{33.3}
\newcommand{\NVisBaseSwapPart}{14.7}
\newcommand{\NVisBaseBlankDrop}{27.3}
\newcommand{\NVisOursOrigAcc}{67.8}
\newcommand{\NVisOursOrigPart}{5.8}
\newcommand{\NVisOursBlankAcc}{38.3}
\newcommand{\NVisOursBlankPart}{7.6}
\newcommand{\NVisOursSwapAcc}{37.9}
\newcommand{\NVisOursSwapPart}{6.8}
\newcommand{\NVisOursBlankDrop}{29.5}

\newcommand{\NConfBaseB}{17.4}

\newcommand{\NConfBaseD}{9.3}
\newcommand{\NConfBaseTop}{63.7}
\newcommand{\NConfOursA}{7.8}

\newcommand{\NConfOursD}{1.9}
\newcommand{\NConfOursTop}{63.8}

\newcommand{\NDecBase}{18.6}
\newcommand{\NFloorOurs}{2.4}
\newcommand{\NDecOurs}{7.4}

\newcommand{\NDecLjevD}{8.4}

\newcommand{\NVlmTwoPart}{21.7}

\newcommand{\NVlmTwoMono}{22.9}

\newcommand{\NVlmTwoOrder}{24.5}
\newcommand{\NVlmFourPart}{23.8}

\newcommand{\NVlmFourMono}{31.2}

\newcommand{\NVlmEightOrder}{22.8}

\newcommand{\suppref}[1]{\ref{#1}}

\providecommand{\FlashDaggerList}{SFT and LLM-as-Jev}

\title{When ``A or B'' Is Less Likely than A:\\Coherent Probabilities for Vision Decision Models}

\author{
Seongrae Noh \quad SeongEun Hong \quad HyeongYeop Kang$^{\dagger}$\\[0.8em]
Korea University\\[0.3em]
{\tt\small \{rhosunr99, seong\_eun, siamiz\_hkang\}@korea.ac.kr}\\[0.3em]
{\small\url{https://seongraenoh.github.io/core/}}
}

\begin{document}
\addtocontents{toc}{\protect\setcounter{tocdepth}{-1}}
\twocolumn[{%
\renewcommand\twocolumn[1][]{#1}%
\maketitle
\vspace{-2.2em}
\begin{center}
\includegraphics[width=\textwidth]{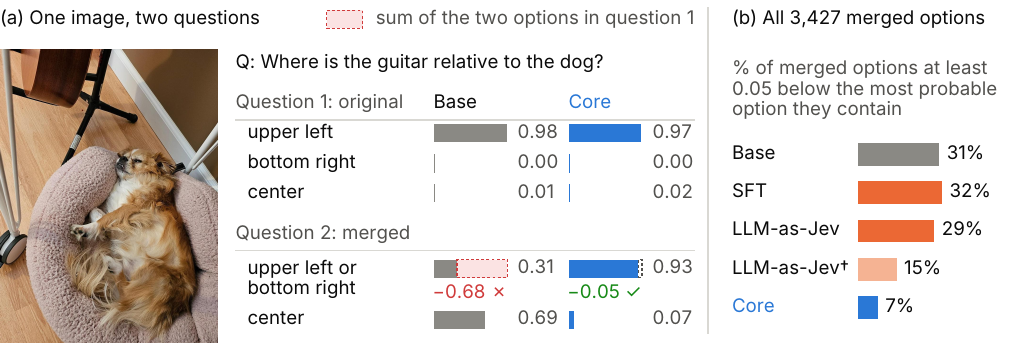}
\vspace{-1.6em}
\captionof{figure}{\textbf{A vision decision model assigns a merged option less probability than one of the options it contains.}
(a) d3-lite on a RealWorldQA item (correct answer: upper left; option texts abbreviated; question 1 averaged over three
option orders). Dashed boxes: the sum of the two options in question 1, the probability that coherence requires.
(b) Share of the 3,427 merged options of the evaluation set that receive at least 0.05 less than the most probable option
they contain (mean over seeds). LLM-as-Jev$^\dagger$: trained on our regrouped questions.}
\label{fig:teaser}
\vspace{1.2em}
\end{center}
}]
\begingroup\renewcommand{\thefootnote}{\fnsymbol{footnote}}\footnotetext[2]{Corresponding author.}\endgroup

\begin{abstract}
Vision decision models answer a multiple-choice question about an image with one probability per option, and software
uses them directly. We show that the probabilities depend on how the options are grouped: a merged option
``A~or~B'' receives on average \NGapAbsBaseLite{} points less probability than the sum of A and B. This under-additivity appears at every model size from 0.85B to 26B and remains after
fine-tuning on labels. We propose \method, which asks each training question with its options regrouped and trains
the answers to all these questions toward one target distribution over the original options, summed into each question's
groups. On 1,050 items from seven benchmarks, \method reduces the total variation distance between the answer to a
regrouped question and the original answer summed into the same groups from \NBasePart{} to \NOursPart{} points on the
0.85B model, against \NLjevDPart{} for the best baseline trained on the same questions. At all five model sizes, \method
reduces this distance without lowering accuracy.
\end{abstract}

\section{Introduction}
\label{sec:intro}

A decision model answers a question that has a fixed set of possible answers, such as a list of options, with one
probability per answer instead of generated text. Jev \cite{jev} and the open d3 model family \cite{d3models} are decision
models, and vision decision models answer such questions about images \cite{pixeljev,innerjev,llmasjev}. Software uses the
probabilities directly: it compares them with thresholds and combines them across questions \cite{li2026axioms}. Such uses
assume that the probabilities are coherent: an event receives the same probability whether the question offers it as one
option or as several.

\begin{figure*}[t]
\centering
\includegraphics[width=\textwidth]{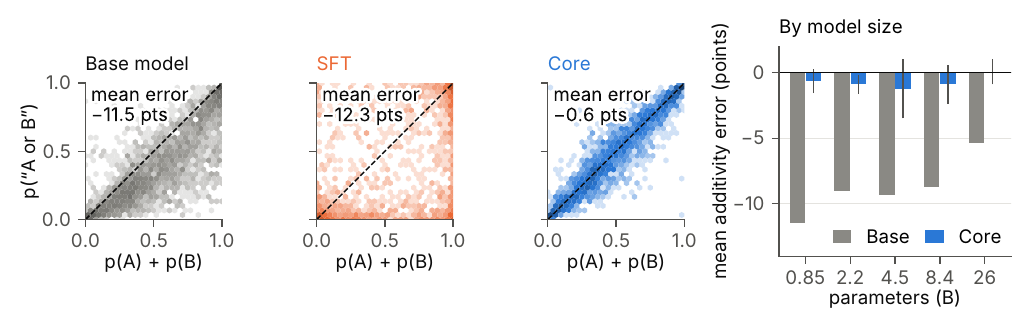}
\vspace{-1.8em}
\caption{\textbf{Merged options receive less than the sum of their members, and fine-tuning on labels does not remove it.}
The left three panels bin the 3,427 merged options of the d3-lite evaluation; darker hexagons hold more merged options (log scale). $x$: the sum of its members'
probabilities on the original question; $y$: its probability on the merged question; dashed: perfect additivity. Right: mean additivity error in points for each model size (base: the model
before fine-tuning; \method: mean over seeds with 95\% interval).}
\label{fig:additivity}
\vspace{-0.5em}
\end{figure*}

\Cref{fig:teaser}a shows that this assumption fails. Asked where the guitar is relative to the dog, with three options, the
d3-lite model assigns 0.98 to the correct option ``upper left''. When ``upper left'' and ``bottom right'' are merged into one
option, the model assigns 0.31 to the merged option and picks ``center''. The merged option describes a larger event than
``upper left'', so its probability cannot be lower. Over the 3,427 merged options of our evaluation set, the base model
assigns a merged option on average \NGapAbsBaseLite{} points less than the sum of the options it contains, and in
\NMonoBaseLite\% of cases at least 5 points less than the most probable of them (\cref{fig:teaser}b). We call this
under-additivity. Human probability judgments show the same pattern, known as subadditivity: an event judged as a whole
receives less probability than the sum of its parts judged separately \cite{tversky1994support}.

Under-additivity does not depend on whether the model answers the question correctly. On the items whose original question
the base model answers correctly, a merged option receives on average 11.0 points less than the options it contains, and
32\% of merged options receive at least 5 points less than the most probable of them; on the items it answers wrongly,
these figures are 12.2 points and 29\%. Existing corrections address other errors. Temperature scaling
\cite{guo2017calibration} corrects the overall confidence level but leaves merged options as they are (partition
violation, defined in \cref{sec:problem}: \NBasePart{} $\to$ \NTsPart{} points). Sorting the options or averaging over
orders \cite{innerjev,zheng2024large} removes order sensitivity but cannot add probabilities across a merge. Consistency
rewards for reinforcement learning (RL) \cite{acre,pagrpo} require the chosen answers to agree across reordered questions, which constrains the
chosen answer but not the probability of a merged option. Fine-tuning on labels raises accuracy but does not remove
under-additivity: supervised fine-tuning (SFT) raises accuracy from \NBaseAcc{} to \NSftAcc, and a merged option still receives \NGapAbsSftLite{} points less than the options it contains
(\cref{fig:additivity}). Software could avoid the problem by asking the original question and adding the answers, but only
if it knows which original options a merged option contains. When a user or another program writes the options, this
information is not available, so the model itself must be coherent.

\method (\emph{coherence under regrouping}) trains the model to assign each event the same probability however the options
are grouped. It asks each training question with its options reordered, merged into ``X or Y'', split into groups, with
one option dropped, and with an option of another item added. All these questions share one target distribution over the
original options: the frozen base model's answer mixed with the smoothed label. The target of each regrouped question is
the same regrouping of this distribution: a merged option receives the sum of its members' target probabilities, dropping
an option renormalizes the rest, and an added option receives almost none. A model that matches every target is therefore
coherent on the training items. We optimize the targets with a sampled reward, the log-ratio of target to model
probability, whose expectation is the negative KL divergence from the model's answer to the target. Optimizing the same
targets exactly, without sampling, gives the same partition violation on d3-lite and d3-flash, so the targets, not the
sampled reward, remove the violation.

Our contributions are summarized as follows:
\begin{itemize}
\item We show that vision decision models are under-additive: in the d3 models at every size, a merged option receives
less probability than the sum of the options it contains, and fine-tuning on labels does not remove this (\cref{sec:problem}).
\item We propose \method, which trains a decision model toward one target distribution shared by all regroupings of a
question (\cref{sec:method}).
\item On five d3 models from 0.85B to 26B, \method lowers the partition violation without lowering accuracy, and the
reduction remains for merge wordings not used in training and when \method is trained on another dataset
(\cref{sec:main,sec:size,sec:general}).
\end{itemize}

\section{Related work}
\label{sec:related}
Decision models such as Jev \cite{jev}, PixelJev \cite{pixeljev}, Releasing the Inner Jev \cite{innerjev} and LLM-as-Jev
\cite{llmasjev} return one probability per option. Their calibration is measured by the expected calibration error (ECE) \cite{naeini2015ece,guo2017calibration}
and Brier score \cite{brier1950} on each question as it is asked, not on questions whose options are merged or split.
Coherence requires that the probability of ``A or B'' equal the sum of the probabilities of A and B \cite{definetti1937};
people violate this \cite{tversky1994support}, and Li \etal \cite{li2026axioms} and Joy \cite{joy2026coherence} report
additivity violations of decision models on yes/no text questions and on hierarchical labels. We extend these diagnoses to
merged options in questions about images and give a training method that reduces the violation. Order sensitivity, also
called selection bias \cite{zheng2024large,pezeshkpour2024order}, is corrected at inference by option sorting
\cite{innerjev}, one yes/no query per option \cite{du2026systemone} or an order-invariant architecture \cite{jeva}, and in
training by RL rewards for answers that agree across option orders \cite{acre,pagrpo}; other rewards based on group relative policy optimization (GRPO)
\cite{grpo} target calibrated confidence \cite{rlcr,cdrl,pdcr} or agreement across modalities \cite{rc2}. These rewards
score the sampled answer, whereas our reward compares its probability with a target probability. Our target is related to
distillation targets \cite{hinton2015distilling,furlanello2018born}, and our regrouped questions to the augmentations of
consistency regularization \cite{xie2020uda,sohn2020fixmatch}; unlike an augmentation, a regrouping changes the set of
answers, so each regrouped question receives the regrouping of the target. Section~\suppref{sup:related} of the
supplementary material discusses these works in full.

\section{Incoherence under option regrouping}
\label{sec:problem}
\begin{figure*}[t]
\centering
\includegraphics[width=\textwidth]{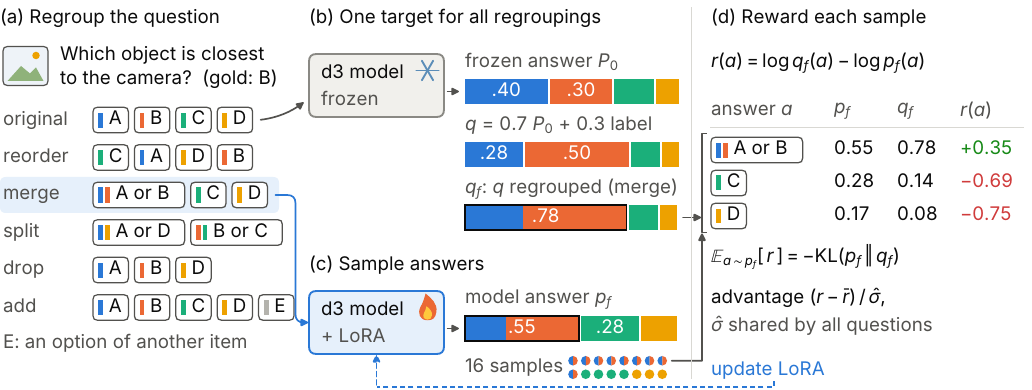}
\vspace{-1.6em}
\caption{\textbf{\method in one step.} (a) A training item is asked as its original question and as regrouped questions.
(b) The frozen model's answer to the original question, mixed 0.7 to 0.3 with the smoothed label, gives one target $q$ over
the original options; each regrouped question gets $q$ regrouped the same way. (c) The trained model answers the regrouped
question, and 16 answers are sampled. (d) The reward of each sample is the log-ratio of its target and model probability;
advantages are divided by one reward scale $\hat\sigma$ shared across questions, and the LoRA weights are updated. Shown for
the question with a merged option; numbers are illustrative.}
\label{fig:overview}
\vspace{-0.5em}
\end{figure*}

\paragraph{Problem setup.}
An item consists of images, a question and $k$ mutually exclusive options $\Omega=\{\omega_1,\dots,\omega_k\}$, one of
which is correct. A decision model returns a distribution over the options as presented. We write $P(\omega)$ for the
model's probability of $\omega$ on the original options, averaged over three random orders. A \emph{regrouped question}
$f$ presents a list of blocks $B_1,\dots,B_m$: disjoint sets of original options, each shown as one option whose text joins
its members with ``or''. We use five kinds of regrouping, each drawn without looking at the correct answer: a new
\emph{order} of the original options; a \emph{pair merge} of two random options; a random \emph{split} of all options
into 2 to $k{-}1$ groups; a \emph{drop} of one option; and an \emph{add} of one option borrowed from another item of the same
benchmark. The model's answer to $f$ is a distribution $p_f$ over the presented blocks.

\paragraph{Coherence constraints.}
Because the options are mutually exclusive, a block is the union of its members, and its probability should be their sum:
\begin{equation}
p_f(B) = \textstyle\sum_{\omega\in B} P(\omega).
\label{eq:additivity}
\end{equation}
\Cref{eq:additivity} covers orders (blocks of size one) and merges. When an option is dropped, the remaining options should
keep their ratios, $p_f(\omega)=P(\omega)/\sum_{\omega'\in f}P(\omega')$. Adding an option changes the set of possible
answers, so the probabilities need not stay the same in general \cite{pixeljev}. We therefore add only options taken from
other items, treat them as wrong, and report the add case separately. We measure the violation of
\cref{eq:additivity} by the \emph{partition violation}
\begin{equation*}
V(f) = \tfrac12 \textstyle\sum_{B\in f} \bigl|\, p_f(B) - \sum_{\omega\in B} P(\omega) \bigr|,
\end{equation*}
the total variation (TV) distance between the answer to a regrouped question and the original answer summed into the same
blocks, averaged over the pair merges and splits of all items. Order TV is the mean TV distance between the answers to
the three orders. IIA TV (independence of irrelevant alternatives) is the TV distance between the answer to a drop or add
question and the original answer, both restricted to the original options they share and renormalized. We also report the \emph{additivity error}
$p_f(B)-\sum_{\omega\in B}P(\omega)$ of every merged option (a block with two or more members), in points ($\times$100);
a negative error means under-additivity. All metrics are reported $\times$100.

\paragraph{Evaluation data.}
We evaluate on 1,050 items, 150 from each of seven benchmarks of the Jev Decision Index vision suite \cite{decisionindex}:
CV-Bench \cite{cvbench}, BLINK \cite{blink}, RealWorldQA \cite{realworldqa}, CharXiv \cite{charxiv}, InfographicVQA
\cite{infovqa}, Mind2Web \cite{mind2web} and R-Bench \cite{rbench}, with 3 to 6 options. Each item yields 8 questions
(3 orders, 2 pair merges, 1 split, 1 drop, 1 add; Mind2Web items have no add), 8,250 questions with 3,427 merged options in total. Training items (2,310) come from the same benchmarks and share no item with the evaluation set (details in Sec.~\suppref{sup:data} of the supplementary material). For each model,
the \emph{hard} items are the 25\% on which the base model's mean violation over the five kinds of regrouping (order TV,
partition violation of the merges and of the split, IIA TV of the drop and the add) is highest; every method is
evaluated on this same set. Because the set is selected by the base model's own violations, part of any trained model's
improvement over the base model on it is regression to the mean; comparisons between trained methods are not affected.

\paragraph{Empirical findings.}
\Cref{fig:additivity} shows the answer of d3-lite to every merged question against the sum of the answers to the
original question. The base model assigns a merged option on average \NGapAbsBaseLite{} points less than the sum of its members, and \NMonoBaseLite\% of merged options receive at least 5 points less than their most probable member, which no coherent
assignment allows. The mean additivity error is negative at every size of the d3 family: $-$\NGapAbsBaseNano{} points at 2.2B,
$-$\NGapAbsBaseMini{} at 4.5B, $-$\NGapAbsBaseFlash{} at 8.4B and $-$\NGapAbsBaseBig{} at 26B; the 26B model is less affected
but not coherent. Generative VLMs scored by the probability of the option letter (Qwen3-VL-Instruct, 2B to 8B) are less
coherent still: their partition violation is \NVlmTwoPart{} to \NVlmFourPart, much of it from order sensitivity (order TV
\NVlmEightOrder{} to \NVlmTwoOrder), and \NVlmTwoMono{} to \NVlmFourMono\% of their merged options receive at least 5 points
less than their most probable member (Tab.~\suppref{tab:vlm} in the supplementary material).
We define the strength of an option as its probability divided by the total probability of the unmerged options. Of four
predictions of a merged option's strength from its members' strengths, their sum, which coherence requires, is the
worst at every size, and their geometric mean is the best on d3-lite (mean absolute log error 0.54 against 1.13;
Tab.~\suppref{tab:strength}). A merged option is predicted better by an average of its members than by their sum.

Fine-tuning on labels does not remove under-additivity. SFT on the original options raises accuracy by \NSftAccGain{} points, but its
mean additivity error is $-$\NGapAbsSftLite{} points and its partition violation rises from \NBasePart{} to \NSftPart{}: the model becomes more
confident on each question separately, and a merged option that the model does not pick receives near-zero probability
(\cref{fig:additivity}, SFT panel). LLM-as-Jev, which adds a KL anchor to the frozen model, keeps the mean additivity error at
$-$\NGapAbsLjevLite{} points.
Training-free corrections do not remove it either: temperature scaling changes the partition violation from \NBasePart{} to \NTsPart, and asking
one yes/no question per option (OBS) lowers it to \NObsPart{} at $k$ times the inference cost (\cref{sec:free}). We therefore train on
regrouped questions directly.

\section{\method: Coherence under regrouping}
\label{sec:method}

\method fine-tunes a decision model so that its answers to regrouped questions satisfy \cref{eq:additivity}. It has three parts:
a target for every regrouped question (\cref{sec:target}), a reward that moves each answer toward its target
(\cref{sec:reward}), and a reward scale shared across questions (\cref{sec:scale}). The ablations in \cref{sec:ablation}
show that the target is the part that reduces the violation. \Cref{fig:overview} shows one step.

\subsection{Regrouping-consistent targets}
\label{sec:target}
Each training item is asked in two fixed random orders of its original options. Let $P_0$ be the frozen base model's
distribution over the original options, averaged over these two orders (unlike $P$ in \cref{sec:problem}, which is the
evaluated model's own answer averaged over three evaluation orders), and let $h_\epsilon=(1-\epsilon)\,h+\epsilon/k$ be
the one-hot label $h$ smoothed with $\epsilon=0.05$. For a distribution $X$ over the original options and a block $B$ we
write $X(B)=\sum_{\omega\in B}X(\omega)$. For a regrouped question $f$ with blocks $B$, the target is
\begin{equation}
q_f(B) = w\,\frac{h_\epsilon(B)}{\sum_{B'\in f} h_\epsilon(B')} + (1-w)\,\frac{P_0(B)}{\sum_{B'\in f}P_0(B')},
\label{eq:target}
\end{equation}
with label weight $w=0.3$. A borrowed option in an add question has no member in $\Omega$; it receives mass $\epsilon/k$ in
the label term and $10^{-3}$ in the frozen-model term before normalization. For every question that keeps all original
options, the denominators are one and $q_f(B)=\sum_{\omega\in B}q(\omega)$ with $q=w\,h_\epsilon+(1-w)P_0$: all targets of an
item are regroupings of one distribution $q$, so a model that matches every target satisfies \cref{eq:additivity} on the
training items. The frozen-model term keeps the target close to the frozen model's answer to the original
question; it needs no labels and carries the model's own uncertainty. The label term moves the target toward the correct
answer, so that \method can become more accurate than the base model (\cref{sec:ablation}).

\subsection{Log-ratio reward}
\label{sec:reward}
For each question $f$ the model samples $K=16$ answers $a_1,\dots,a_K\sim p_f$ and receives
\begin{equation}
r_i = \log q_f(a_i) - \log p_f(a_i),
\qquad
\mathbb{E}_{a\sim p_f}[\,r\,] = -\KL\!\left(p_f \,\Vert\, q_f\right).
\label{eq:reward}
\end{equation}
An answer that the model under-weights relative to its target, such as a merged option in \cref{fig:overview}d, receives a
positive reward, and an answer that it over-weights receives a negative reward. The loss of question $f$ is the policy-gradient loss
$\mathcal L_f=-\frac1K\sum_i A_i\log p_f(a_i)$ with advantages $A_i=(r_i-\bar r)/\hat\sigma$, where $\bar r$ is the mean
reward of the $K$ samples. Each step uses four training items. Each item is asked as four questions: its two training orders and two regroupings
drawn uniformly from its merges, split, drop and add. The step loss is the mean over the 16 questions.
Because the reward depends on the full distribution through $\log p_f$, the expected update reduces the KL divergence to
the target rather than raising the probability of one preferred answer. A reward of $\log q_f(a)$ alone is maximized
by putting all mass on the block with the largest target, not by $p_f=q_f$. In d3-lite pilot runs with the reward
$\log q_f(a)$ alone, merged options without the correct answer received 3.7 to 5.2 times the sum of their members' probabilities (Tab.~\suppref{tab:pilot}).

\begin{table*}[t]
\centering\small
\setlength{\tabcolsep}{4.2pt}
\caption{\textbf{Trained methods on d3-lite.} Acc.: accuracy over all questions, original and regrouped; ECE and Brier score over all questions; partition violation on all items and on the hardest 25\%; order TV; IIA TV when an option is added or dropped (all $\times$100). Mean $\pm$ standard deviation over 3 seeds per baseline and \NOursSeedsLite{} for \method. Bold: best; underline: second best; the base model is not ranked.}
\label{tab:main}
\begin{tabularx}{\linewidth}{@{}lYYYYYYYY@{}}
\toprule
 & Acc. $\uparrow$ & ECE $\downarrow$ & Brier $\downarrow$ & \multicolumn{2}{c}{Partition $\downarrow$} & Order $\downarrow$ & \multicolumn{2}{c}{IIA $\downarrow$} \\
\cmidrule(lr){5-6}\cmidrule(l){8-9}
Method &  & all & all & all & hard & TV & add & drop \\
\midrule
Base model & 60.8 & 5.6 & 48.4 & 14.8 & 25.7 & 5.5 & 5.5 & 6.4 \\
SFT~\cite{pixeljev,innerjev} & 66.6\,{\scriptsize$\pm$1.2} & 21.3\,{\scriptsize$\pm$1.9} & 52.5\,{\scriptsize$\pm$0.3} & 19.3\,{\scriptsize$\pm$1.0} & 27.8\,{\scriptsize$\pm$1.5} & 9.5\,{\scriptsize$\pm$0.3} & 10.4\,{\scriptsize$\pm$0.3} & 12.3\,{\scriptsize$\pm$0.5} \\
LLM-as-Jev~\cite{llmasjev} & \underline{66.8}\,{\scriptsize$\pm$0.5} & \underline{1.9}\,{\scriptsize$\pm$0.6} & \underline{42.9}\,{\scriptsize$\pm$0.8} & \underline{14.4}\,{\scriptsize$\pm$0.8} & \underline{21.2}\,{\scriptsize$\pm$1.4} & \underline{6.3}\,{\scriptsize$\pm$0.3} & \underline{7.1}\,{\scriptsize$\pm$0.4} & \underline{7.4}\,{\scriptsize$\pm$0.2} \\
ACRE~\cite{acre} & 66.4\,{\scriptsize$\pm$0.6} & 7.4\,{\scriptsize$\pm$1.8} & 44.6\,{\scriptsize$\pm$1.4} & 15.9\,{\scriptsize$\pm$1.7} & 23.7\,{\scriptsize$\pm$1.7} & 10.0\,{\scriptsize$\pm$2.8} & 8.5\,{\scriptsize$\pm$1.6} & 10.3\,{\scriptsize$\pm$2.6} \\
PA-GRPO~\cite{pagrpo} & 63.8\,{\scriptsize$\pm$0.3} & 16.8\,{\scriptsize$\pm$0.8} & 51.9\,{\scriptsize$\pm$1.2} & 16.0\,{\scriptsize$\pm$1.3} & 24.6\,{\scriptsize$\pm$1.4} & 10.6\,{\scriptsize$\pm$2.6} & 9.9\,{\scriptsize$\pm$2.2} & 11.4\,{\scriptsize$\pm$1.4} \\
\rowcolor{oursrow}\method{} (ours) & \textbf{67.1}\,{\scriptsize$\pm$0.6} & \textbf{1.7}\,{\scriptsize$\pm$0.1} & \textbf{42.4}\,{\scriptsize$\pm$0.5} & \textbf{5.8}\,{\scriptsize$\pm$0.1} & \textbf{8.5}\,{\scriptsize$\pm$0.2} & \textbf{4.1}\,{\scriptsize$\pm$0.2} & \textbf{4.5}\,{\scriptsize$\pm$0.2} & \textbf{5.2}\,{\scriptsize$\pm$0.2} \\
\bottomrule
\end{tabularx}
\end{table*}

\subsection{Advantage normalization and KL regularization}
\label{sec:scale}
GRPO divides each question's rewards by their own standard deviation \cite{grpo}, which gives a question that already
matches its target the same advantage scale as one far from it. We divide by one running estimate over all questions,
$\hat\sigma\leftarrow0.99\,\hat\sigma+0.01\,\mathrm{std}(r_{1:K})$ after a warm-up of 50 questions, so that questions far from
their target receive larger advantages; on d3-lite the difference from GRPO's scale is within seed variation
(\cref{sec:ablation}). If all $K$ samples of a question fall in one block, the advantages are zero and the question gives no
update. We add no KL term $\KL(P_0\Vert\bar p)$ toward the frozen model \cite{llmasjev,acre}, where $\bar p$ is the trained
model's answer on the original options: the target already contains $P_0$ with weight 0.7.

\paragraph{Implementation details.}
A d3 model reads the final hidden state of its prompt encoder and maps it to one score per
presented option with a readout matrix; a softmax over these scores gives $p_f$. We train LoRA adapters \cite{lora} of rank 16
($\alpha=32$) on all linear layers of the language model; the vision encoder and the readout matrix are frozen. We use AdamW with learning rate $3\cdot10^{-5}$, no schedule, gradient clipping at 1.0 and bf16,
for 600 steps of 4 items (300 steps for the 26B model; all hyperparameters in Sec.~\suppref{sup:impl} of the supplementary material). A d3-lite run takes 42 minutes on one RTX 4090, a d3-flash run
about 70 minutes on one H100.

\section{Experiments}
\label{sec:experiments}

\paragraph{Models.}
We fine-tune the five d3 decision models \cite{d3models}: d3-lite (0.85B parameters), d3-nano (2.2B), d3-mini (4.5B),
d3-flash (8.4B) and d3 (26B). Baselines are compared on d3-lite and d3-flash.

\paragraph{Baselines.}
Every baseline is fine-tuned on the same training items as \method: SFT with cross-entropy, as in PixelJev and Releasing the Inner Jev \cite{pixeljev,innerjev}; LLM-as-Jev
\cite{llmasjev}, cross-entropy plus a KL anchor to the frozen model on the option distribution (learning rate $10^{-4}$ as
published); ACRE \cite{acre} and PA-GRPO \cite{pagrpo}, RL with rewards for correct and order-consistent answers. On d3-lite we
train each baseline in two conditions: on the original question in two orders, as published, and on the same regrouped questions as
\method (marked $^\dagger$); on d3-flash, all four as published and \FlashDaggerList{} also on the regrouped questions.
All methods use the same number of steps and LoRA rank. The learning rates of ACRE and PA-GRPO are listed in Sec.~\suppref{sup:baselines} of the supplementary
material.

\paragraph{Evaluation protocol.}
Baselines use 3 training seeds; \method uses \NOursSeedsLite{} on d3-lite and 3 on d3-flash. We report means and standard deviations over seeds; 95\%
intervals over seeds are in Sec.~\suppref{sup:tables} of the supplementary material, together with paired item-bootstrap intervals for every difference from
\method. Accuracy (Acc.) is the fraction of questions, original and regrouped, whose most probable block
contains the correct option. Acc., ECE (15 bins) and the Brier score use all questions in which some block contains the
correct option.

\paragraph{Training-free baselines.} Methods that do not change the model's weights learn nothing from the training items,
so we compare them separately (\cref{sec:free}): temperature scaling \cite{guo2017calibration}, the calibration step of
PixelJev and Releasing the Inner Jev \cite{pixeljev,innerjev}, with one temperature fitted on 400 training items; options
sorted by text, the order fix of Releasing the Inner Jev; averaging the answers to the given and the reversed order, an option
of the d3 runtime \cite{d3models}; and OBS \cite{du2026systemone}, one yes/no question per option. Recovering the members of a merged option from its text and
summing their probabilities on the original question is compared with \method{} in Sec.~\suppref{sup:parse} of the
supplementary material.

\subsection{Main results}
\label{sec:main}

\begin{figure}[t]
\centering
\includegraphics[width=\linewidth]{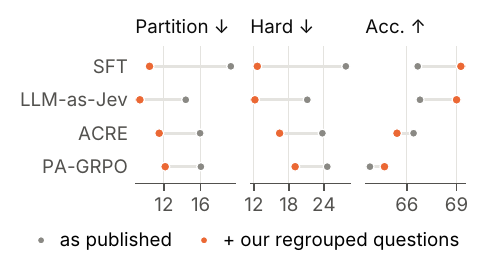}
\vspace{-1.6em}
\caption{\textbf{Baselines trained as published and on our regrouped questions, d3-lite.} Gray: trained on
the original question, as published; orange: also trained on our regrouped questions ($^\dagger$). Partition violation
on all and on hard items, and Acc. ($\times$100; means over 3 seeds).}
\label{fig:regrouped}
\vspace{-0.6em}
\end{figure}

\Cref{tab:main} gives the metrics of every trained method on d3-lite, and Fig.~\suppref{fig:sup_scatter} plots accuracy against
partition violation at both sizes. On d3-lite, \method lowers the partition violation from \NBasePart{} to \NOursPart{} and on the hard items from
\NBaseHard{} to \NOursHard{}, and it lowers ECE from \NBaseEce{} to \NOursEce.
Trained on the original options, no baseline lowers the partition violation by more than \NBestPubPartDrop{} points, and SFT
raises it to \NSftPart: cross-entropy raises the top probability of each question separately, and nothing in it constrains merged options. Training on our regrouped questions lowers the partition violation of every baseline on d3-lite (\cref{fig:regrouped}), so the regrouped questions reduce the violation even
without our target. The best of these baselines, LLM-as-Jev$^\dagger$, reaches \NLjevDPart{} (hard items: \NLjevDHard), which is
\NPartMargin{} (\NHardMargin) points above \method.
\method's ECE is not significantly lower than that of LLM-as-Jev, and LLM-as-Jev$^\dagger$ has a lower Brier score (\NLjevDBrier{} against \NOursBrier; paired intervals in Tab.~\suppref{supt:delta-lite}). In accuracy, \method is \NAccGain{} points above the base model and within the paired
95\% intervals of SFT, LLM-as-Jev and ACRE trained on the original options; the cross-entropy baselines trained on the
regrouped questions are more accurate, LLM-as-Jev$^\dagger$ by \NAccGapLjevD{} and SFT$^\dagger$ by \NAccGapSftD{} points. No baseline, with or without the regrouped questions, has both a lower partition
violation and a higher accuracy than \method (\cref{tab:main,fig:regrouped}).

\paragraph{Residual violation and downstream decisions.} Order sensitivity accounts for part of the remaining violation: $P$
averages three orders while each regrouped question is asked in one, and a model that satisfies \cref{eq:additivity} exactly but is as order-sensitive as \method
would have a partition violation of \NFloorOurs{} instead of \NOursPart. At a threshold of 0.5, the decision from a merged option differs from the
one implied by the original question for \NDecBase\%, \NDecLjevD\% and \NDecOurs\% of merged options with the base model,
LLM-as-Jev$^\dagger$ and \method. \method's partition violation is lower than that of the base model and LLM-as-Jev$^\dagger$ on each of
the seven benchmarks (Tab.~\suppref{tab:bench}).

\paragraph{Results on d3-flash.} \method again has the lowest partition violation (Tab.~\suppref{tab:main_flash}):
\NFlOursPart{} against \NFlBasePart{} for the base model (hard items: \NFlOursHard{} against \NFlBaseHard), with an accuracy
gain of \NFlAccGain{} points. The closest baseline, SFT$^\dagger$, reaches \NFlSftDPart{} (hard items: \NFlSftDHard) at an
ECE of \NFlSftDEce{} against \NFlOursEce; SFT and LLM-as-Jev, with and without the regrouped questions, have a higher mean accuracy
than \method (\NFlLjevAcc{} to \NFlSftDAcc{} against \NFlOursAcc), and ACRE and PA-GRPO a lower one.
Unlike on d3-lite, the regrouped questions do not lower the partition violation of LLM-as-Jev (\NFlLjevDPart{} with them,
\NFlLjevPart{} without). Both versions have an ECE not significantly different from \method's and a lower Brier score
(\NFlLjevDBrier{} and \NFlLjevBrier{} against \NFlOursBrier).

\subsection{Comparison with training-free corrections}
\label{sec:free}
\begin{table*}[t]
\centering\small
\setlength{\tabcolsep}{4.2pt}
\caption{\textbf{Training-free corrections.} Temperature scaling fits one temperature on 400 training items; order averaging averages the answers to two option orders; OBS asks one yes/no question per option. \method (trained; mean $\pm$ standard deviation over seeds) is shown for reference. Columns as in \cref{tab:main}; bold: best; underline: second best.}
\label{tab:free}
\begin{tabularx}{\linewidth}{@{}lYYYYYYYY@{}}
\toprule
 & Acc. $\uparrow$ & ECE $\downarrow$ & Brier $\downarrow$ & \multicolumn{2}{c}{Partition $\downarrow$} & Order $\downarrow$ & \multicolumn{2}{c}{IIA $\downarrow$} \\
\cmidrule(lr){5-6}\cmidrule(l){8-9}
Method &  & all & all & all & hard & TV & add & drop \\
\midrule
\multicolumn{9}{@{}l}{\textit{d3-lite (0.85B)}}\\
Base model & 60.8 & 5.6 & 48.4 & 14.8 & 25.7 & 5.5 & 5.5 & 6.4 \\
Temperature scaling~\cite{guo2017calibration} & 60.8 & \underline{1.9} & \underline{47.9} & 14.6 & 23.3 & 4.7 & 4.9 & 5.6 \\
Sorted options~\cite{innerjev} & 61.1 & 5.9 & 48.1 & 13.9 & 22.9 & \textbf{0.0} & \underline{4.1} & 5.5 \\
Order averaging (2$\times$)~\cite{d3models} & \underline{61.3} & 4.7 & 48.1 & 14.2 & 24.2 & \underline{2.9} & \textbf{4.0} & \textbf{5.1} \\
OBS ($k\times$)~\cite{du2026systemone} & \underline{61.3} & 6.4 & 49.7 & \underline{11.6} & \underline{15.0} & 5.3 & 5.6 & 5.5 \\
\rowcolor{oursrow}\method{} (trained) & \textbf{67.1}\,{\scriptsize$\pm$0.6} & \textbf{1.7}\,{\scriptsize$\pm$0.1} & \textbf{42.4}\,{\scriptsize$\pm$0.5} & \textbf{5.8}\,{\scriptsize$\pm$0.1} & \textbf{8.5}\,{\scriptsize$\pm$0.2} & 4.1\,{\scriptsize$\pm$0.2} & 4.5\,{\scriptsize$\pm$0.2} & \underline{5.2}\,{\scriptsize$\pm$0.2} \\
\midrule
\multicolumn{9}{@{}l}{\textit{d3-flash (8.4B)}}\\
Base model & 72.6 & 3.5 & 35.3 & 15.0 & 35.7 & 5.9 & 6.7 & 12.2 \\
Temperature scaling~\cite{guo2017calibration} & 72.6 & 2.7 & 35.1 & 14.3 & 31.1 & 5.4 & 6.2 & 10.6 \\
Sorted options~\cite{innerjev} & 74.2 & 4.3 & \underline{34.2} & 13.9 & 31.6 & \textbf{0.0} & \underline{4.7} & 11.7 \\
Order averaging (2$\times$)~\cite{d3models} & 73.9 & \textbf{2.0} & 34.7 & 14.5 & 33.4 & \underline{3.3} & 5.1 & 10.9 \\
OBS ($k\times$)~\cite{du2026systemone} & \underline{74.7} & 11.7 & 37.0 & \underline{11.6} & \underline{15.1} & 6.5 & 6.5 & \underline{7.0} \\
\rowcolor{oursrow}\method{} (trained) & \textbf{77.2}\,{\scriptsize$\pm$0.2} & \underline{2.3}\,{\scriptsize$\pm$0.3} & \textbf{30.2}\,{\scriptsize$\pm$0.2} & \textbf{4.8}\,{\scriptsize$\pm$0.2} & \textbf{9.2}\,{\scriptsize$\pm$0.4} & 3.8\,{\scriptsize$\pm$0.1} & \textbf{4.0}\,{\scriptsize$\pm$0.1} & \textbf{5.0}\,{\scriptsize$\pm$0.1} \\
\bottomrule
\end{tabularx}
\end{table*}

We test whether the violation can be corrected at inference (\cref{tab:free}): by correcting confidence or option order, or by
asking about each option separately. Temperature scaling lowers ECE (\NBaseEce{} to \NTsEce{} on d3-lite), sorting removes
order sensitivity by construction, and averaging two orders reduces them; none of the three changes the partition violation by
more than \NTfMaxDrop{} points on d3-lite and \NFlTfMaxDrop{} on d3-flash. OBS lowers the partition violation to
\NObsPart{} on d3-lite and \NFlObsPart{} on d3-flash at $k$ times the inference cost, and raises ECE on d3-flash (\NFlBaseEce{} to \NFlObsEce; \NBaseEce{} to \NObsEce{} on d3-lite). Correcting confidence or option order does not reduce the violation; OBS reduces it at $k$ times the
inference cost, and \method at no extra inference cost.

\subsection{Effect of model size}
\label{sec:size}
\begin{figure}[t]
\centering
\includegraphics[width=\linewidth]{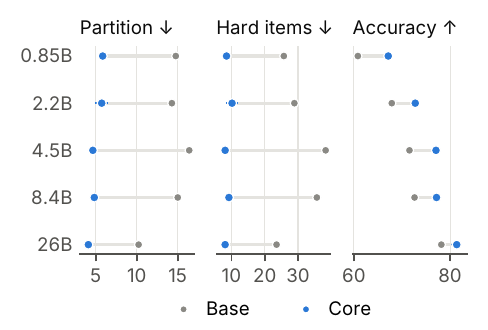}
\vspace{-1.6em}
\caption{\textbf{Base model and \method at five model sizes.} Partition violation on all and on hard items, and Acc.
($\times$100). Gray: base model; blue: \method (mean over seeds with 95\% interval).}
\label{fig:sizes}
\vspace{-0.6em}
\end{figure}
\Cref{fig:sizes} compares \method with the base model at all five sizes. The base models' partition violation lies
between \NSizeBasePartMin{} and \NSizeBasePartMax{} and does not fall with size up to 8.4B; on hard items it is highest at
4.5B and 8.4B (\NMiniBaseHard{} and \NFlBaseHard). \method lowers it to \NSizeOursPartMin--\NSizeOursPartMax{} (hard
items: \NSizeOursHardMin--\NSizeOursHardMax) at every size and raises accuracy by \NSizeAccGainMin{} to
\NSizeAccGainMax{} points. ECE falls at 0.85B and 2.2B, does not change significantly at
4.5B and 8.4B, and rises from \NBigBaseEce{} to \NBigOursEce{} at 26B, where the base model is the best calibrated of the
five (paired intervals in Sec.~\suppref{sup:tables} of the supplementary material).

\subsection{Ablation study}
\label{sec:ablation}
\begin{table}[t]
\centering\footnotesize
\setlength{\tabcolsep}{1.9pt}
\caption{\textbf{Ablations on d3-lite.} Each row changes one part of \method ($\times$100; mean $\pm$ standard deviation over 3 seeds; the \method row is taken from \cref{tab:main}). The last two rows optimize the same target without sampling.}
\label{tab:ablation}
\begin{tabularx}{\linewidth}{@{}l*{5}{Y}@{}}
\toprule
 & Acc.$\uparrow$ & ECE$\downarrow$ & Part.$\downarrow$ & Hard$\downarrow$ & Order$\downarrow$ \\
\midrule
\rowcolor{oursrow}\method & 67.1\,{\scriptsize$\pm$0.6} & 1.7\,{\scriptsize$\pm$0.1} & 5.8\,{\scriptsize$\pm$0.1} & 8.5\,{\scriptsize$\pm$0.2} & 4.1\,{\scriptsize$\pm$0.2} \\
\midrule
+ KL anchor & 66.9\,{\scriptsize$\pm$0.2} & 1.4\,{\scriptsize$\pm$0.2} & 5.8\,{\scriptsize$\pm$0.1} & 8.7\,{\scriptsize$\pm$0.3} & 4.1\,{\scriptsize$\pm$0.1} \\
per-question std & 67.1\,{\scriptsize$\pm$0.4} & 2.0\,{\scriptsize$\pm$0.1} & 6.1\,{\scriptsize$\pm$0.3} & 9.0\,{\scriptsize$\pm$0.4} & 4.1\,{\scriptsize$\pm$0.0} \\
$w=0$ (no label) & 64.3\,{\scriptsize$\pm$0.3} & 2.5\,{\scriptsize$\pm$0.2} & 5.5\,{\scriptsize$\pm$0.1} & 8.3\,{\scriptsize$\pm$0.4} & 3.8\,{\scriptsize$\pm$0.1} \\
$w=0.5$ & 68.6\,{\scriptsize$\pm$0.2} & 2.5\,{\scriptsize$\pm$0.2} & 6.4\,{\scriptsize$\pm$0.2} & 8.9\,{\scriptsize$\pm$0.3} & 4.4\,{\scriptsize$\pm$0.1} \\
$w=1$ (label only) & 68.2\,{\scriptsize$\pm$0.7} & 13.3\,{\scriptsize$\pm$1.3} & 11.8\,{\scriptsize$\pm$0.6} & 13.1\,{\scriptsize$\pm$0.8} & 8.0\,{\scriptsize$\pm$0.1} \\
$w$ from 0 to 0.6 & 68.2\,{\scriptsize$\pm$0.3} & 3.2\,{\scriptsize$\pm$0.7} & 6.5\,{\scriptsize$\pm$0.1} & 8.9\,{\scriptsize$\pm$0.5} & 4.3\,{\scriptsize$\pm$0.1} \\
exact $\KL(p_f\Vert q_f)$ & 67.3\,{\scriptsize$\pm$0.3} & 1.5\,{\scriptsize$\pm$0.4} & 5.8\,{\scriptsize$\pm$0.1} & 8.5\,{\scriptsize$\pm$0.2} & 4.0\,{\scriptsize$\pm$0.1} \\
cross-entropy to $q_f$ & 67.8\,{\scriptsize$\pm$0.2} & 2.0\,{\scriptsize$\pm$0.8} & 5.7\,{\scriptsize$\pm$0.1} & 8.5\,{\scriptsize$\pm$0.2} & 3.9\,{\scriptsize$\pm$0.1} \\
\bottomrule
\end{tabularx}
\end{table}

\Cref{tab:ablation} changes one part of \method at a time on d3-lite. Adding the KL anchor to the frozen model gives the same partition violation (\NAblAnchorPart) and an accuracy
of \NAblAnchorAcc{} against \NOursAcc, a difference within seed variation. Dividing each question's
rewards by their own standard deviation, as GRPO does, gives an accuracy of \NAblUAcc{} and a partition violation of
\NAblUPart{} against \NOursAcc{} and \NOursPart; these paired differences are within seed variation (Tab.~\suppref{supt:abl-lite}).
The label weight $w$ trades accuracy against calibration and consistency. Without the label ($w=0$) the target is the frozen
model alone: the partition violation is \NAblWzeroPart, slightly below \method's, but accuracy is \NAblWzeroAcc{} against
\NOursAcc. With $w=0.5$ accuracy is \NAblWhalfAcc{} (the 95\% interval of the paired difference includes zero), and ECE and partition violation rise to \NAblWhalfEce{} and
\NAblWhalfPart. With the label alone ($w=1$) the target no longer carries the model's uncertainty: ECE rises to \NAblWoneEce{}
and partition violation to \NAblWonePart, without a further gain in accuracy (\NAblWoneAcc). Raising $w$ from 0 to 0.6 during
training gives an accuracy of \NAblSchedAcc{} at an ECE of \NAblSchedEce.

\paragraph{Sampled versus exact objectives.}
A decision model gives its full answer distribution in one forward pass, so the expectation in \cref{eq:reward} can also be
computed exactly. The last two rows of \cref{tab:ablation} optimize the same target without sampling: by the exact
$\KL(p_f\Vert q_f)$ over the blocks, and by the cross-entropy $-\sum_B q_f(B)\log p_f(B)$. They reach accuracies of
\NAblExactAcc{} and \NAblCeAcc{} and partition violations of \NAblExactPart{} and \NAblCePart, against \NOursAcc{} and
\NOursPart{} for \method; on d3-flash the exact KL reaches \NFlExactAcc{} and \NFlExactPart{} against \NFlOursAcc{} and
\NFlOursPart. In ECE and partition violation, every paired difference to \method is within its 95\% interval over
seeds, and in accuracy all but one: the exact KL on d3-flash is slightly more accurate (Tab.~\suppref{supt:abl}). The target of
\cref{eq:target}, not the sampled estimator, reduces the violation. The sampled reward needs only the probabilities of the
answers it draws, so it also applies when the answer set is too large to sum over (tested in Sec.~\suppref{sup:large} of the supplementary
material); with the 3 to 6
options of our items, the exact objectives work as well.

\subsection{Generalization and control analyses}
\label{sec:general}
\begin{table}[t]
\centering\footnotesize
\setlength{\tabcolsep}{3pt}
\caption{\textbf{Partition violation under unseen merge wordings, d3-lite} ($\times$100; lower is better). The merged options of the evaluation set are worded in ways not used in training (training uses ``X or Y''); \method: mean $\pm$ standard deviation over 2 seeds.}
\label{tab:wording}
\begin{tabularx}{\linewidth}{@{}l*{5}{Y}@{}}
\toprule
 & X or Y & one of & either & any of & X / Y \\
\midrule
Base model & 14.8 & 14.1 & 15.1 & 15.1 & 17.9 \\
\method & 5.8\,{\scriptsize$\pm$0.1} & 6.6\,{\scriptsize$\pm$0.1} & 6.7\,{\scriptsize$\pm$0.3} & 7.0\,{\scriptsize$\pm$0.3} & 11.7\,{\scriptsize$\pm$0.2} \\
\bottomrule
\end{tabularx}
\end{table}

\begin{table}[t]
\centering\footnotesize
\setlength{\tabcolsep}{2pt}
\caption{\textbf{Training on A-OKVQA, evaluated on the Decision Index items, d3-lite} ($\times$100). Mean $\pm$ standard deviation over 3 seeds; the last row is \method from \cref{tab:main}.}
\label{tab:aokvqa}
\begin{tabularx}{\linewidth}{@{}ll*{4}{Y}@{}}
\toprule
Method & Training items & Acc.$\uparrow$ & ECE$\downarrow$ & Part.$\downarrow$ & Hard$\downarrow$ \\
\midrule
Base model & -- & 60.8 & 5.6 & 14.8 & 25.7 \\
SFT & A-OKVQA & 61.9\,{\scriptsize$\pm$0.1} & 17.5\,{\scriptsize$\pm$3.8} & 12.0\,{\scriptsize$\pm$0.3} & 20.7\,{\scriptsize$\pm$0.5} \\
\method & A-OKVQA & 62.2\,{\scriptsize$\pm$0.4} & 2.4\,{\scriptsize$\pm$0.6} & 8.3\,{\scriptsize$\pm$0.0} & 14.8\,{\scriptsize$\pm$0.2} \\
\method & Decision Index & 67.1\,{\scriptsize$\pm$0.6} & 1.7\,{\scriptsize$\pm$0.1} & 5.8\,{\scriptsize$\pm$0.1} & 8.5\,{\scriptsize$\pm$0.2} \\
\bottomrule
\end{tabularx}
\end{table}

\paragraph{Control for confidence.}
A model can satisfy \cref{eq:additivity} by answering uniformly, so a lower partition violation could come from lower
confidence \cite{li2026axioms}. \method's mean top probability on the original options (\NConfOursTop) is within 0.1 points of
the base model's (\NConfBaseTop). In each of four bins of each model's own top probability, \method's partition violation
(\NConfOursD{} to \NConfOursA) is lower than the base model's (\NConfBaseD{} to \NConfBaseB) and SFT's (Tab.~\suppref{supt:conf}).

\paragraph{Unseen merge wordings.}
Training uses ``X or Y'' only. \Cref{tab:wording} rewords the merged options of the evaluation set in four other
ways. On ``one of'', ``either'' and ``any of'', \method's
violation is \NWordOursOneof, \NWordOursEither{} and \NWordOursAny, close to the trained wording's \NWordOursOr{} and
less than half of the base model's (\NWordBaseOneof{} to \NWordBaseAny). On ``X / Y'', which has no connective, it falls
from \NWordBaseSlash{} to \NWordOursSlash{} (by \NWordRedSlash\%, against \NWordRedMin--\NWordRedMax\% on the other
wordings).

\paragraph{Cross-dataset training.}
Trained on A-OKVQA \cite{a_okvqa} items (\cref{tab:aokvqa}), which have COCO images and everyday questions, and evaluated on the Decision Index
items, \method lowers the partition violation from \NAokBasePart{} to \NAokOursPart{} (hard items: \NAokBaseHard{} to
\NAokOursHard) with an ECE of \NAokOursEce; SFT on the same items reaches \NAokSftPart{} with an ECE of \NAokSftEce. Accuracy
stays near the base model for both (\NAokOursAcc{} and \NAokSftAcc{} against \NAokBaseAcc). The reduction in partition violation transfers from A-OKVQA training items to the Decision Index
items, whereas the accuracy gain of \cref{sec:main} requires training items from the evaluation benchmarks.

\paragraph{Dependence on the image.}
If \method became coherent by ignoring the image, removing the image would leave its accuracy intact. We test this with
the seed-0 model of \method, whose accuracy with the original image is \NVisOursOrigAcc. Replacing the image with a blank one
lowers this accuracy by \NVisOursBlankDrop{} points (to \NVisOursBlankAcc) and the base model's by \NVisBaseBlankDrop; with
the image of another item, \method reaches \NVisOursSwapAcc{} and the base model \NVisBaseSwapAcc. \method's accuracy
depends on the image as much as the base model's. Its partition violation rises from \NVisOursOrigPart{} to \NVisOursBlankPart{} and
\NVisOursSwapPart, still below the base model's \NVisBaseBlankPart{} and \NVisBaseSwapPart.

\paragraph{Generative VLMs.}
Section~\suppref{sup:vlmtrain} of the supplementary material applies \method{} to generative VLMs scored by the
probability of the option letter.

\section{Conclusion}
\label{sec:conclusion}

Vision decision models assign a merged option less probability than the sum of its members, often less than its most
probable member, at every size we tested, and fine-tuning on labels does not remove this. \method trains the answers to all regrouped questions of an item toward regroupings of one target distribution, with a
reward whose expectation is the negative KL divergence to that target. The target, not the sampled reward, reduces the violation; \method
also raises accuracy at every size and lowers ECE at 0.85B and 2.2B. Regrouped questions are generated automatically and need no labels to
evaluate, so we suggest reporting partition violation next to accuracy and ECE for decision models.

\paragraph{Limitations.}
We test only merged options that list their members. Merged options that name a broader event without listing its members
(``left half'') are future work.

\FloatBarrier
\section*{LLM Usage Statement}
We used Claude (Anthropic) through Claude Code to assist the authors with the experiments and the writing of this paper.
The authors are fully responsible for the content.

{
    \small
    \bibliographystyle{ieeenat_fullname}
    \bibliography{main,supp}

\begin{thebibliography}{42}
\providecommand{\natexlab}[1]{#1}
\providecommand{\url}[1]{\texttt{#1}}
\expandafter\ifx\csname urlstyle\endcsname\relax
  \providecommand{\doi}[1]{doi: #1}\else
  \providecommand{\doi}{doi: \begingroup \urlstyle{rm}\Url}\fi

\bibitem[Damani et~al.(2026)Damani, Puri, Slocum, Shenfeld, Choshen, Kim, and
  Andreas]{rlcr}
Mehul Damani, Isha Puri, Stewart Slocum, Idan Shenfeld, Leshem Choshen, Yoon
  Kim, and Jacob Andreas.
\newblock Beyond binary rewards: Training lms to reason about their
  uncertainty.
\newblock In \emph{International conference on learning representations}, pages
  59456--59493, 2026.

\bibitem[De~Finetti(1937)]{definetti1937}
Bruno De~Finetti.
\newblock La pr{\'e}vision: ses lois logiques, ses sources subjectives.
\newblock In \emph{Annales de l'institut Henri Poincar{\'e}}, pages 1--68,
  1937.

\bibitem[{Decision Index maintainers}(2026)]{decisionindex}
{Decision Index maintainers}.
\newblock {Jev Decision Index}.
\newblock Leaderboard:
  \url{https://huggingface.co/spaces/multimodalart/jev-decision-index}.
  Reproduction kit: \url{https://github.com/apolinario/decision-index}, 2026.
\newblock Accessed November 15, 2026.

\bibitem[Delaunay(2026)]{delaunay2026}
Julien Delaunay.
\newblock Can the confidence and decisions of system 1 models be trusted? an
  explainability and perturbation audit of jev-style decision models.
\newblock \emph{arXiv preprint arXiv:2611.12521}, 2026.

\bibitem[Deng et~al.(2023)Deng, Gu, Zheng, Chen, Stevens, Wang, Sun, and
  Su]{mind2web}
Xiang Deng, Yu Gu, Boyuan Zheng, Shijie Chen, Sam Stevens, Boshi Wang, Huan
  Sun, and Yu Su.
\newblock Mind2web: Towards a generalist agent for the web.
\newblock \emph{Advances in Neural Information Processing Systems},
  36:\penalty0 28091--28114, 2023.

\bibitem[Du et~al.(2026{\natexlab{a}})Du, Wang, Zhang, Xu, Yang, Kong, Liu, and
  Zhu]{cdrl}
Yuetian Du, Yucheng Wang, Rongyu Zhang, Zhijie Xu, Boyu Yang, Ming Kong, Jie
  Liu, and Qiang Zhu.
\newblock Linking perception, confidence and accuracy in mllms.
\newblock \emph{arXiv preprint arXiv:2603.12149}, 2026{\natexlab{a}}.

\bibitem[Du et~al.(2026{\natexlab{b}})Du, Zhao, Zhao, Li, Gao, Wang, Xu, Cui,
  and Qin]{du2026systemone}
Yanrui Du, Danyang Zhao, Sendong Zhao, Wenhao Li, Yibo Gao, Haochun Wang,
  Zhuoer Xu, Shiwen Cui, and Bing Qin.
\newblock How reliable are system-one decisions in medicine? a comparative
  study of jev, jev-like models, and llms.
\newblock \emph{arXiv preprint arXiv:2610.18529}, 2026{\natexlab{b}}.

\bibitem[Fu et~al.(2024)Fu, Hu, Li, Feng, Wang, Lin, Roth, Smith, Ma, and
  Krishna]{blink}
Xingyu Fu, Yushi Hu, Bangzheng Li, Yu Feng, Haoyu Wang, Xudong Lin, Dan Roth,
  Noah~A Smith, Wei-Chiu Ma, and Ranjay Krishna.
\newblock Blink: Multimodal large language models can see but not perceive.
\newblock In \emph{European Conference on Computer Vision}, pages 148--166.
  Springer, 2024.

\bibitem[Furlanello et~al.(2018)Furlanello, Lipton, Tschannen, Itti, and
  Anandkumar]{furlanello2018born}
Tommaso Furlanello, Zachary Lipton, Michael Tschannen, Laurent Itti, and Anima
  Anandkumar.
\newblock Born again neural networks.
\newblock In \emph{International conference on machine learning}, pages
  1607--1616. PMLR, 2018.

\bibitem[Geuter et~al.(2026)Geuter, Cheng, and Alvarez-Melis]{innerjev}
Jonathan Geuter, Andrew Cheng, and David Alvarez-Melis.
\newblock Releasing the inner jev: A simple recipe for open decision models.
\newblock \emph{arXiv preprint arXiv:2610.17205}, 2026.

\bibitem[Glenn et~al.(1950)]{brier1950}
W~Brier Glenn et~al.
\newblock Verification of forecasts expressed in terms of probability.
\newblock \emph{Monthly weather review}, 78\penalty0 (1):\penalty0 1--3, 1950.

\bibitem[Guo et~al.(2017)Guo, Pleiss, Sun, and Weinberger]{guo2017calibration}
Chuan Guo, Geoff Pleiss, Yu Sun, and Kilian~Q Weinberger.
\newblock On calibration of modern neural networks.
\newblock In \emph{International conference on machine learning}, pages
  1321--1330. PMLR, 2017.

\bibitem[Guo et~al.(2025)Guo, Xu, Zhang, Song, Peng, Deng, Dong, Nakayama,
  Geng, Wang, et~al.]{rbench}
Meng-Hao Guo, Jiajun Xu, Yi Zhang, Jiaxi Song, Haoyang Peng, Yi-Xuan Deng,
  Xinzhi Dong, Kiyohiro Nakayama, Zhengyang Geng, Chen Wang, et~al.
\newblock R-bench: Graduate-level multi-disciplinary benchmarks for llm \& mllm
  complex reasoning evaluation.
\newblock \emph{arXiv preprint arXiv:2505.02018}, 2025.

\bibitem[Hinton et~al.(2015)Hinton, Vinyals, and Dean]{hinton2015distilling}
Geoffrey Hinton, Oriol Vinyals, and Jeff Dean.
\newblock Distilling the knowledge in a neural network.
\newblock \emph{arXiv preprint arXiv:1503.02531}, 2015.

\bibitem[Hu et~al.(2021)Hu, Shen, Wallis, Allen-Zhu, Li, Wang, Wang, and
  Chen]{lora}
Edward~J Hu, Yelong Shen, Phillip Wallis, Zeyuan Allen-Zhu, Yuanzhi Li, Shean
  Wang, Lu Wang, and Weizhu Chen.
\newblock Lora: Low-rank adaptation of large language models.
\newblock \emph{arXiv preprint arXiv:2106.09685}, 2021.

\bibitem[Huang et~al.(2025)Huang, Huang, Zheng, Li, Chen, Shi, and Cheng]{acre}
Minbin Huang, Runhui Huang, Chuanyang Zheng, Jingyao Li, Guoxuan Chen, Han Shi,
  and Hong Cheng.
\newblock Answer-consistent chain-of-thought reinforcement learning for
  multi-modal large langauge models.
\newblock \emph{arXiv preprint arXiv:2510.10104}, 2025.

\bibitem[Joy(2026)]{joy2026coherence}
Saman~Sarker Joy.
\newblock Do system one decisions add up? a study of probabilistic coherence.
\newblock \emph{arXiv preprint arXiv:2609.33971}, 2026.

\bibitem[Li et~al.(2026)Li, He, and Li]{li2026axioms}
Keyi Li, Yihao He, and Quanyi Li.
\newblock Beyond calibration: Do a typed-decision model's probabilities obey
  the probability axioms?
\newblock \emph{arXiv preprint arXiv:2609.33209}, 2026.

\bibitem[Li and Wang(2026)]{jeva}
Qi Li and Xinchao Wang.
\newblock Jeva: System one models from first principles.
\newblock \emph{arXiv preprint arXiv:2610.18773}, 2026.

\bibitem[Li and Wagle(2026)]{llmasjev}
Yinheng Li and Justin Wagle.
\newblock Llm2jev: Llms are already jev-style decision models--when and how to
  fine-tune them.
\newblock \emph{arXiv preprint arXiv:2610.02076}, 2026.

\bibitem[Liu et~al.(2026)Liu, Li, Xu, Pei, Wang, Zhao, Zheng, He, Yao, Yang,
  et~al.]{beyondmc}
Yesheng Liu, Hao Li, Haiyu Xu, Baoqi Pei, Jiahao Wang, Mingxuan Zhao, Jing-Shu
  Zheng, Zheqi He, JG Yao, Xi Yang, et~al.
\newblock Beyond multiple choice: Verifiable openqa for robust vision-language
  rft.
\newblock In \emph{Proceedings of the IEEE/CVF Conference on Computer Vision
  and Pattern Recognition}, pages 18859--18869, 2026.

\bibitem[Mathew et~al.(2022)Mathew, Bagal, Tito, Karatzas, Valveny, and
  Jawahar]{infovqa}
Minesh Mathew, Viraj Bagal, Rub{\`e}n Tito, Dimosthenis Karatzas, Ernest
  Valveny, and CV Jawahar.
\newblock Infographicvqa.
\newblock In \emph{2022 IEEE/CVF Winter Conference on Applications of Computer
  Vision (WACV)}, pages 2582--2591. IEEE, 2022.

\bibitem[Naeini et~al.(2015)Naeini, Cooper, and Hauskrecht]{naeini2015ece}
Mahdi~Pakdaman Naeini, Gregory Cooper, and Milos Hauskrecht.
\newblock Obtaining well calibrated probabilities using bayesian binning.
\newblock In \emph{Proceedings of the AAAI conference on artificial
  intelligence}, 2015.

\bibitem[Nixon et~al.(2019)Nixon, Dusenberry, Jerfel, Nguyen, Liu, Zhang, and
  Tran]{nixon2019measuring}
Jeremy Nixon, Mike Dusenberry, Ghassen Jerfel, Timothy Nguyen, Jeremiah Liu,
  Linchuan Zhang, and Dustin Tran.
\newblock Measuring calibration in deep learning.
\newblock \emph{arXiv preprint arXiv:1904.01685}, 2019.

\bibitem[Pezeshkpour and Hruschka(2024)]{pezeshkpour2024order}
Pouya Pezeshkpour and Estevam Hruschka.
\newblock Large language models sensitivity to the order of options in
  multiple-choice questions.
\newblock In \emph{Findings of the Association for Computational Linguistics:
  NAACL 2024}, pages 2006--2017, 2024.

\bibitem[Saeed et~al.(2026)Saeed, Atiq, Kumar, Jamil, Iqbal, and
  Aydin]{shiftjev}
Muhammad~Atif Saeed, Azka Atiq, Adarsh Kumar, Akhtar Jamil, Uzair Iqbal, and
  Roland Aydin.
\newblock Shift-jev: Certified abstention and gain-gated continual adaptation
  for llm-based decision models under distribution shift.
\newblock \emph{arXiv preprint arXiv:2610.34329}, 2026.

\bibitem[Schwenk et~al.(2022)Schwenk, Khandelwal, Clark, Marino, and
  Mottaghi]{a_okvqa}
Dustin Schwenk, Apoorv Khandelwal, Christopher Clark, Kenneth Marino, and
  Roozbeh Mottaghi.
\newblock A-okvqa: A benchmark for visual question answering using world
  knowledge.
\newblock In \emph{European conference on computer vision}, pages 146--162.
  Springer, 2022.

\bibitem[Shao et~al.(2024)Shao, Wang, Zhu, Xu, Song, Bi, Zhang, Zhang, Li, Wu,
  et~al.]{grpo}
Zhihong Shao, Peiyi Wang, Qihao Zhu, Runxin Xu, Junxiao Song, Xiao Bi, Haowei
  Zhang, Mingchuan Zhang, YK Li, Yang Wu, et~al.
\newblock Deepseekmath: Pushing the limits of mathematical reasoning in open
  language models.
\newblock \emph{arXiv preprint arXiv:2402.03300}, 2024.

\bibitem[Sohn et~al.(2020)Sohn, Berthelot, Carlini, Zhang, Zhang, Raffel,
  Cubuk, Kurakin, and Li]{sohn2020fixmatch}
Kihyuk Sohn, David Berthelot, Nicholas Carlini, Zizhao Zhang, Han Zhang, Colin
  Raffel, Ekin~Dogus Cubuk, Alexey Kurakin, and Chun-Liang Li.
\newblock Fixmatch: Simplifying semi-supervised learning with consistency and
  confidence.
\newblock \emph{Advances in neural information processing systems},
  33:\penalty0 596--608, 2020.

\bibitem[Tong et~al.(2024)Tong, Brown, Wu, Woo, Middepogu, Akula, Yang, Yang,
  Iyer, Pan, et~al.]{cvbench}
Shengbang Tong, Ellis Brown, Penghao Wu, Sanghyun Woo, Manoj Middepogu, Sai~C
  Akula, Jihan Yang, Shusheng Yang, Adithya Iyer, Xichen Pan, et~al.
\newblock Cambrian-1: A fully open, vision-centric exploration of multimodal
  llms.
\newblock \emph{Advances in Neural Information Processing Systems},
  37:\penalty0 87310--87356, 2024.

\bibitem[Tversky and Kahneman(1989)]{tversky1986rational}
Amos Tversky and Daniel Kahneman.
\newblock Rational choice and the framing of decisions.
\newblock In \emph{Multiple criteria decision making and risk analysis using
  microcomputers}, pages 81--126. Springer, 1989.

\bibitem[Tversky and Koehler(1994)]{tversky1994support}
Amos Tversky and Derek~J Koehler.
\newblock Support theory: A nonextensional representation of subjective
  probability.
\newblock \emph{Psychological review}, 101\penalty0 (4):\penalty0 547, 1994.

\bibitem[{TypeSafe AI}(2026)]{jev}
{TypeSafe AI}.
\newblock Introducing {System One} models \& {Jev}.
\newblock TypeSafe AI blog,
  \url{https://typesafe.ai/blog/introducing-system-one-models-and-jev}, 2026.
\newblock Published September 15, 2026. Accessed November 15, 2026.

\bibitem[{vLLM Semantic Router Team}(2026)]{d3models}
{vLLM Semantic Router Team}.
\newblock {d3}: A multimodal foundation decision model.
\newblock Hugging Face model card, \url{https://huggingface.co/vllm-sr/d3},
  2026.
\newblock Model family vllm-sr/d3-lite, d3-nano, d3-mini, d3-flash and d3.
  Accessed November 15, 2026.

\bibitem[Wang et~al.(2024)Wang, Xia, He, Chen, Liu, Zhu, Liang, Wu, Liu,
  Malladi, et~al.]{charxiv}
Zirui Wang, Mengzhou Xia, Luxi He, Howard Chen, Yitao Liu, Richard Zhu, Kaiqu
  Liang, Xindi Wu, Haotian Liu, Sadhika Malladi, et~al.
\newblock Charxiv: Charting gaps in realistic chart understanding in multimodal
  llms.
\newblock \emph{Advances in Neural Information Processing Systems},
  37:\penalty0 113569--113697, 2024.

\bibitem[{xAI}(2024)]{realworldqa}
{xAI}.
\newblock {Grok-1.5 Vision Preview}.
\newblock \url{https://x.ai/news/grok-1.5v}, 2024.
\newblock Release of the {RealWorldQA} benchmark, April 12, 2024.

\bibitem[Xie et~al.(2020)Xie, Dai, Hovy, Luong, and Le]{xie2020uda}
Qizhe Xie, Zihang Dai, Eduard Hovy, Thang Luong, and Quoc Le.
\newblock Unsupervised data augmentation for consistency training.
\newblock \emph{Advances in neural information processing systems},
  33:\penalty0 6256--6268, 2020.

\bibitem[Yoon et~al.(2026)Yoon, Yoon, Hong, Eom, Koo, Hasegawa-Johnson, Dai,
  Luo, and Yoo]{pdcr}
Hee~Suk Yoon, Eunseop Yoon, Ji~Woo Hong, SooHwan Eom, Gwanhyeong Koo, Mark
  Hasegawa-Johnson, Qi Dai, Chong Luo, and Chang~D Yoo.
\newblock Pdcr: Perception-decomposed confidence reward for vision-language
  reasoning.
\newblock \emph{arXiv preprint arXiv:2605.13467}, 2026.

\bibitem[Zhang et~al.(2026)Zhang, Dong, Pei, and Mao]{rc2}
Zirui Zhang, Haoyu Dong, Kexin Pei, and Chengzhi Mao.
\newblock R-c2: Cycle-consistent reinforcement learning improves multimodal
  reasoning.
\newblock \emph{arXiv preprint arXiv:2603.25720}, 2026.

\bibitem[Zheng et~al.(2024)Zheng, Zhou, Meng, Zhou, and Huang]{zheng2024large}
Chujie Zheng, Hao Zhou, Fandong Meng, Jie Zhou, and Minlie Huang.
\newblock Large language models are not robust multiple choice selectors.
\newblock In \emph{International Conference on Learning Representations}, pages
  19426--19454, 2024.

\bibitem[Zheng et~al.(2026)Zheng, Yuan, Yao, Gu, Zheng, and He]{pagrpo}
Jinquan Zheng, Jia Yuan, Jiacheng Yao, Chenyang Gu, Pujun Zheng, and Guoxiu He.
\newblock Mitigating selection bias in large language models via
  permutation-aware grpo.
\newblock In \emph{Proceedings of the 64th Annual Meeting of the Association
  for Computational Linguistics (Volume 1: Long Papers)}, pages 35125--35143,
  2026.

\bibitem[Zhou et~al.(2026)Zhou, Yang, and Zhao]{pixeljev}
Xunlan Zhou, Xianliang Yang, and Li Zhao.
\newblock From text decisions to pixels: An study of jev-style visual choice
  model.
\newblock \emph{arXiv preprint arXiv:2609.29283}, 2026.

\end{thebibliography}
}

\clearpage

\newcommand{\SuppTabLite}{%
\begin{table*}[tp]
\centering\scriptsize
\setlength{\tabcolsep}{2pt}
\caption{\textbf{All metrics with intervals, d3-lite (0.85B).} Methods of main-paper Tables 1 and 2 and Fig.~4. Values $\times$100; mean $\pm$ 95\% Student-$t$ interval over seeds (3 per trained baseline, 5 for \method); the base model and the training-free corrections are deterministic. Orig.\ acc.: accuracy of the order-averaged answer to the original question.}\label{supt:lite}
\begin{tabularx}{\textwidth}{@{}lYYYYYYYYY@{}}
\toprule
 & Acc.$\uparrow$ & Orig.\ acc.$\uparrow$ & ECE$\downarrow$ & Brier$\downarrow$ & \multicolumn{2}{c}{Partition$\downarrow$} & Order$\downarrow$ & \multicolumn{2}{c}{IIA$\downarrow$} \\
\cmidrule(lr){6-7}\cmidrule(l){9-10}
 & & & & & all & hard & & add & drop \\
\midrule
Base model & 60.8 & 59.6 & 5.6 & 48.4 & 14.8 & 25.7 & 5.5 & 5.5 & 6.4 \\
\multicolumn{10}{@{}l}{\textit{training-free corrections}} \\
Temperature scaling & 60.8 & 59.6 & 1.9 & 47.9 & 14.6 & 23.3 & 4.7 & 4.9 & 5.6 \\
Sorted options & 61.1 & 59.6 & 5.9 & 48.1 & 13.9 & 22.9 & 0.0 & 4.1 & 5.5 \\
Order averaging & 61.3 & 59.8 & 4.7 & 48.1 & 14.2 & 24.2 & 2.9 & 4.0 & 5.1 \\
OBS & 61.3 & 59.5 & 6.4 & 49.7 & 11.6 & 15.0 & 5.3 & 5.6 & 5.5 \\
\multicolumn{10}{@{}l}{\textit{trained as published}} \\
SFT & 66.6 $\pm$ 3.1 & 67.1 $\pm$ 2.5 & 21.3 $\pm$ 4.8 & 52.5 $\pm$ 0.7 & 19.3 $\pm$ 2.4 & 27.8 $\pm$ 3.6 & 9.5 $\pm$ 0.8 & 10.4 $\pm$ 0.7 & 12.3 $\pm$ 1.2 \\
LLM-as-Jev & 66.8 $\pm$ 1.2 & 66.9 $\pm$ 2.3 & 1.9 $\pm$ 1.6 & 42.9 $\pm$ 1.9 & 14.4 $\pm$ 1.9 & 21.2 $\pm$ 3.4 & 6.3 $\pm$ 0.8 & 7.1 $\pm$ 0.9 & 7.4 $\pm$ 0.6 \\
ACRE & 66.4 $\pm$ 1.4 & 66.1 $\pm$ 3.0 & 7.4 $\pm$ 4.4 & 44.6 $\pm$ 3.5 & 15.9 $\pm$ 4.3 & 23.7 $\pm$ 4.3 & 10.0 $\pm$ 6.9 & 8.5 $\pm$ 4.0 & 10.3 $\pm$ 6.4 \\
PA-GRPO & 63.8 $\pm$ 0.9 & 62.1 $\pm$ 2.1 & 16.8 $\pm$ 1.9 & 51.9 $\pm$ 3.1 & 16.0 $\pm$ 3.3 & 24.6 $\pm$ 3.5 & 10.6 $\pm$ 6.6 & 9.9 $\pm$ 5.5 & 11.4 $\pm$ 3.5 \\
\multicolumn{10}{@{}l}{\textit{trained on our regrouped questions ($^\dagger$)}} \\
SFT$^\dagger$ & 69.3 $\pm$ 1.4 & 67.7 $\pm$ 2.1 & 16.0 $\pm$ 5.2 & 45.8 $\pm$ 3.8 & 10.4 $\pm$ 0.9 & 12.6 $\pm$ 0.9 & 7.3 $\pm$ 0.3 & 8.3 $\pm$ 0.4 & 10.3 $\pm$ 0.7 \\
LLM-as-Jev$^\dagger$ & 69.0 $\pm$ 1.2 & 66.9 $\pm$ 1.3 & 2.3 $\pm$ 4.5 & 40.7 $\pm$ 0.8 & 9.4 $\pm$ 1.9 & 12.2 $\pm$ 2.4 & 6.1 $\pm$ 1.6 & 6.8 $\pm$ 3.3 & 7.8 $\pm$ 1.6 \\
ACRE$^\dagger$ & 65.4 $\pm$ 0.6 & 63.0 $\pm$ 1.5 & 11.0 $\pm$ 2.3 & 46.2 $\pm$ 1.3 & 11.5 $\pm$ 0.6 & 16.4 $\pm$ 0.8 & 7.4 $\pm$ 1.4 & 6.7 $\pm$ 0.3 & 9.4 $\pm$ 0.6 \\
PA-GRPO$^\dagger$ & 64.6 $\pm$ 0.5 & 61.9 $\pm$ 0.7 & 18.1 $\pm$ 1.0 & 51.2 $\pm$ 0.9 & 12.1 $\pm$ 0.8 & 19.1 $\pm$ 1.9 & 8.7 $\pm$ 1.5 & 8.2 $\pm$ 0.6 & 10.9 $\pm$ 1.7 \\
\midrule
\rowcolor{oursrow}\method & 67.1 $\pm$ 0.8 & 64.6 $\pm$ 0.8 & 1.7 $\pm$ 0.2 & 42.4 $\pm$ 0.6 & 5.8 $\pm$ 0.2 & 8.5 $\pm$ 0.2 & 4.1 $\pm$ 0.2 & 4.5 $\pm$ 0.3 & 5.2 $\pm$ 0.3 \\
\bottomrule
\end{tabularx}
\end{table*}}

\newcommand{\SuppTabDeltaLite}{%
\begin{table*}[t]
\centering\scriptsize
\setlength{\tabcolsep}{2pt}
\caption{\textbf{Paired differences from \method, d3-lite (0.85B).} Method minus \method, $\times$100, with paired item-bootstrap 95\% intervals from 10{,}000 resamples after averaging over each method's seeds. Positive values mean higher accuracy, ECE, Brier score or partition violation than \method.}\label{supt:delta-lite}
\begin{tabularx}{\textwidth}{@{}lYYYYYY@{}}
\toprule
 & Acc. & Orig.\ acc. & ECE & Brier & \multicolumn{2}{c}{Partition} \\
\cmidrule(l){6-7}
 & & & & & all & hard \\
\midrule
Base model & $-$6.3 [$-$7.4, $-$5.3] & $-$5.0 [$-$6.5, $-$3.4] & +3.9 [1.4, 4.6] & +6.0 [5.3, 6.8] & +8.9 [8.4, 9.4] & +17.2 [16.2, 18.3] \\
\multicolumn{7}{@{}l}{\textit{training-free corrections}} \\
Temperature scaling & $-$6.3 [$-$7.4, $-$5.3] & $-$5.0 [$-$6.5, $-$3.4] & +0.2 [$-$1.0, 1.3] & +5.5 [4.7, 6.2] & +8.7 [8.3, 9.2] & +14.7 [13.8, 15.7] \\
Sorted options & $-$6.0 [$-$7.2, $-$4.9] & $-$5.0 [$-$6.6, $-$3.3] & +4.2 [1.6, 5.0] & +5.7 [5.0, 6.5] & +8.0 [7.5, 8.6] & +14.4 [13.1, 15.7] \\
Order averaging & $-$5.8 [$-$6.9, $-$4.8] & $-$4.8 [$-$6.3, $-$3.3] & +3.0 [0.6, 3.8] & +5.7 [4.9, 6.4] & +8.3 [7.8, 8.8] & +15.7 [14.5, 16.8] \\
OBS & $-$5.9 [$-$7.2, $-$4.5] & $-$5.1 [$-$7.0, $-$3.1] & +4.8 [1.3, 6.2] & +7.2 [6.1, 8.5] & +5.8 [5.4, 6.2] & +6.5 [5.6, 7.4] \\
\multicolumn{7}{@{}l}{\textit{trained as published}} \\
SFT & $-$0.5 [$-$1.8, 0.9] & +2.5 [0.8, 4.3] & +19.7 [17.0, 20.5] & +10.0 [8.1, 12.0] & +13.4 [12.6, 14.3] & +19.2 [17.3, 21.2] \\
LLM-as-Jev & $-$0.4 [$-$1.5, 0.8] & +2.3 [0.6, 3.9] & +0.2 [$-$0.9, 1.0] & +0.4 [$-$0.5, 1.4] & +8.5 [8.1, 9.0] & +12.6 [11.6, 13.7] \\
ACRE & $-$0.8 [$-$1.8, 0.2] & +1.5 [0.2, 2.9] & +5.8 [3.4, 6.3] & +2.2 [1.3, 3.2] & +10.1 [9.6, 10.6] & +15.2 [14.0, 16.4] \\
PA-GRPO & $-$3.4 [$-$4.5, $-$2.3] & $-$2.5 [$-$4.0, $-$0.9] & +15.1 [12.5, 15.9] & +9.5 [8.2, 10.9] & +10.2 [9.6, 10.8] & +16.1 [14.7, 17.4] \\
\multicolumn{7}{@{}l}{\textit{trained on our regrouped questions ($^\dagger$)}} \\
SFT$^\dagger$ & +2.1 [0.9, 3.4] & +3.1 [1.4, 4.7] & +14.3 [11.8, 15.0] & +3.4 [1.6, 5.1] & +4.6 [4.1, 5.1] & +4.1 [2.9, 5.2] \\
LLM-as-Jev$^\dagger$ & +1.8 [0.9, 2.8] & +2.3 [0.9, 3.7] & +0.6 [$-$0.4, 1.2] & $-$1.7 [$-$2.5, $-$0.9] & +3.5 [3.3, 3.8] & +3.6 [3.1, 4.2] \\
ACRE$^\dagger$ & $-$1.8 [$-$2.3, $-$1.2] & $-$1.6 [$-$2.6, $-$0.7] & +9.4 [6.6, 9.9] & +3.8 [3.0, 4.7] & +5.6 [5.2, 6.1] & +7.9 [6.9, 8.9] \\
PA-GRPO$^\dagger$ & $-$2.5 [$-$3.2, $-$1.8] & $-$2.7 [$-$3.8, $-$1.6] & +16.4 [13.5, 17.0] & +8.8 [7.4, 10.2] & +6.3 [5.7, 6.9] & +10.5 [9.3, 11.8] \\
\bottomrule
\end{tabularx}
\end{table*}}

\newcommand{\SuppTabFlash}{%
\begin{table*}[t]
\centering\scriptsize
\setlength{\tabcolsep}{2pt}
\caption{\textbf{All metrics with intervals, d3-flash (8.4B).} Values $\times$100; mean $\pm$ 95\% Student-$t$ interval over seeds (3 per trained method); the base model and the training-free corrections are deterministic. Orig.\ acc.: see \cref{supt:lite}.}\label{tab:main_flash}
\begin{tabularx}{\textwidth}{@{}lYYYYYYYYY@{}}
\toprule
 & Acc.$\uparrow$ & Orig.\ acc.$\uparrow$ & ECE$\downarrow$ & Brier$\downarrow$ & \multicolumn{2}{c}{Partition$\downarrow$} & Order$\downarrow$ & \multicolumn{2}{c}{IIA$\downarrow$} \\
\cmidrule(lr){6-7}\cmidrule(l){9-10}
 & & & & & all & hard & & add & drop \\
\midrule
Base model & 72.6 & 73.2 & 3.5 & 35.3 & 15.0 & 35.7 & 5.9 & 6.7 & 12.2 \\
\multicolumn{10}{@{}l}{\textit{training-free corrections}} \\
Temperature scaling & 72.6 & 73.4 & 2.7 & 35.1 & 14.3 & 31.1 & 5.4 & 6.2 & 10.6 \\
Sorted options & 74.2 & 73.5 & 4.3 & 34.2 & 13.9 & 31.6 & 0.0 & 4.7 & 11.7 \\
Order averaging & 73.9 & 73.7 & 2.0 & 34.7 & 14.5 & 33.4 & 3.3 & 5.1 & 10.9 \\
OBS & 74.7 & 74.0 & 11.7 & 37.0 & 11.6 & 15.1 & 6.5 & 6.5 & 7.0 \\
\multicolumn{10}{@{}l}{\textit{trained as published}} \\
SFT & 78.5 $\pm$ 0.6 & 79.0 $\pm$ 0.5 & 14.5 $\pm$ 1.9 & 34.7 $\pm$ 2.5 & 14.5 $\pm$ 0.9 & 28.1 $\pm$ 0.9 & 6.1 $\pm$ 0.6 & 6.5 $\pm$ 0.1 & 10.9 $\pm$ 1.8 \\
LLM-as-Jev & 78.0 $\pm$ 1.6 & 77.4 $\pm$ 2.1 & 1.8 $\pm$ 1.3 & 29.3 $\pm$ 1.9 & 11.5 $\pm$ 0.3 & 19.4 $\pm$ 2.6 & 4.8 $\pm$ 0.7 & 4.8 $\pm$ 0.5 & 6.5 $\pm$ 1.4 \\
ACRE & 76.1 $\pm$ 1.4 & 76.1 $\pm$ 0.5 & 4.8 $\pm$ 0.9 & 32.4 $\pm$ 1.1 & 13.5 $\pm$ 3.5 & 26.1 $\pm$ 11.0 & 6.9 $\pm$ 1.9 & 6.6 $\pm$ 1.5 & 9.8 $\pm$ 2.8 \\
PA-GRPO & 76.3 $\pm$ 1.8 & 75.8 $\pm$ 1.2 & 10.3 $\pm$ 2.4 & 35.0 $\pm$ 3.1 & 13.9 $\pm$ 2.3 & 28.8 $\pm$ 6.5 & 7.2 $\pm$ 0.7 & 7.1 $\pm$ 1.9 & 12.0 $\pm$ 5.8 \\
\multicolumn{10}{@{}l}{\textit{trained on our regrouped questions ($^\dagger$)}} \\
SFT$^\dagger$ & 80.7 $\pm$ 1.0 & 79.4 $\pm$ 3.0 & 12.0 $\pm$ 0.5 & 30.3 $\pm$ 1.3 & 7.3 $\pm$ 2.4 & 14.9 $\pm$ 3.8 & 5.0 $\pm$ 1.0 & 5.3 $\pm$ 1.5 & 9.4 $\pm$ 1.3 \\
LLM-as-Jev$^\dagger$ & 79.4 $\pm$ 1.4 & 77.4 $\pm$ 1.9 & 2.2 $\pm$ 1.4 & 28.1 $\pm$ 1.7 & 12.1 $\pm$ 1.8 & 26.6 $\pm$ 2.6 & 5.3 $\pm$ 1.2 & 10.7 $\pm$ 1.4 & 11.2 $\pm$ 0.6 \\
\midrule
\rowcolor{oursrow}\method & 77.2 $\pm$ 0.5 & 75.3 $\pm$ 1.1 & 2.3 $\pm$ 0.8 & 30.2 $\pm$ 0.5 & 4.8 $\pm$ 0.4 & 9.2 $\pm$ 0.9 & 3.8 $\pm$ 0.2 & 4.0 $\pm$ 0.2 & 5.0 $\pm$ 0.3 \\
\bottomrule
\end{tabularx}
\end{table*}}

\newcommand{\SuppTabDeltaFlash}{%
\begin{table*}[t]
\centering\scriptsize
\setlength{\tabcolsep}{2pt}
\caption{\textbf{Paired differences from \method, d3-flash (8.4B).} Method minus \method, $\times$100, with paired item-bootstrap 95\% intervals from 10{,}000 resamples after averaging over each method's seeds. Positive values mean higher accuracy, ECE, Brier score or partition violation than \method.}\label{supt:delta-flash}
\begin{tabularx}{\textwidth}{@{}lYYYYYY@{}}
\toprule
 & Acc. & Orig.\ acc. & ECE & Brier & \multicolumn{2}{c}{Partition} \\
\cmidrule(l){6-7}
 & & & & & all & hard \\
\midrule
Base model & $-$4.6 [$-$5.6, $-$3.5] & $-$2.1 [$-$3.2, $-$1.1] & +1.3 [$-$0.8, 2.6] & +5.1 [4.3, 6.0] & +10.2 [9.4, 11.0] & +26.4 [24.6, 28.3] \\
\multicolumn{7}{@{}l}{\textit{training-free corrections}} \\
Temperature scaling & $-$4.6 [$-$5.6, $-$3.5] & $-$1.9 [$-$3.0, $-$0.9] & +0.4 [$-$0.5, 1.3] & +4.9 [4.0, 5.8] & +9.5 [8.9, 10.2] & +21.9 [20.3, 23.5] \\
Sorted options & $-$3.0 [$-$4.1, $-$2.0] & $-$1.8 [$-$3.0, $-$0.7] & +2.1 [$-$0.1, 3.2] & +4.0 [3.2, 4.9] & +9.1 [8.3, 10.0] & +22.4 [20.1, 24.7] \\
Order averaging & $-$3.2 [$-$4.2, $-$2.3] & $-$1.6 [$-$2.6, $-$0.6] & $-$0.2 [$-$1.7, 1.1] & +4.5 [3.7, 5.4] & +9.6 [8.9, 10.4] & +24.2 [22.3, 26.1] \\
OBS & $-$2.5 [$-$3.8, $-$1.3] & $-$1.3 [$-$3.1, 0.4] & +9.4 [6.8, 10.6] & +6.8 [5.5, 8.1] & +6.8 [6.4, 7.1] & +5.9 [5.1, 6.7] \\
\multicolumn{7}{@{}l}{\textit{trained as published}} \\
SFT & +1.3 [0.1, 2.6] & +3.7 [1.9, 5.4] & +12.2 [9.6, 13.5] & +4.6 [2.9, 6.2] & +9.7 [8.8, 10.5] & +18.9 [16.9, 21.0] \\
LLM-as-Jev & +0.8 [$-$0.2, 1.8] & +2.1 [0.8, 3.5] & $-$0.5 [$-$1.4, 0.3] & $-$0.8 [$-$1.7, $-$0.1] & +6.6 [6.2, 7.1] & +10.1 [9.0, 11.4] \\
ACRE & $-$1.1 [$-$2.0, $-$0.1] & +0.7 [$-$0.5, 1.9] & +2.5 [0.2, 3.7] & +2.2 [1.5, 3.0] & +8.7 [8.1, 9.3] & +16.9 [15.6, 18.2] \\
PA-GRPO & $-$0.9 [$-$1.8, 0.0] & +0.5 [$-$0.8, 1.8] & +8.0 [5.3, 9.2] & +4.8 [3.8, 5.9] & +9.1 [8.3, 9.8] & +19.6 [17.7, 21.4] \\
\multicolumn{7}{@{}l}{\textit{trained on our regrouped questions ($^\dagger$)}} \\
SFT$^\dagger$ & +3.5 [2.2, 5.0] & +4.1 [2.3, 6.0] & +9.7 [7.1, 11.1] & +0.1 [$-$1.7, 2.0] & +2.4 [1.9, 3.0] & +5.6 [4.0, 7.4] \\
LLM-as-Jev$^\dagger$ & +2.2 [1.2, 3.2] & +2.0 [0.8, 3.3] & $-$0.1 [$-$1.8, 1.2] & $-$2.1 [$-$3.0, $-$1.2] & +7.3 [6.7, 7.9] & +17.3 [15.8, 18.8] \\
\bottomrule
\end{tabularx}
\end{table*}}

\newcommand{\SuppTabSizes}{%
\begin{table*}[t]
\centering\scriptsize
\setlength{\tabcolsep}{2pt}
\caption{\textbf{All metrics with intervals, d3-nano, d3-mini and d3 (26B).} Values $\times$100; mean $\pm$ 95\% Student-$t$ interval over seeds (3 per trained method); the base model is deterministic. Training at 26B uses 300 steps. Orig.\ acc.: see \cref{supt:lite}.}\label{supt:sizes}
\begin{tabularx}{\textwidth}{@{}lYYYYYYYYY@{}}
\toprule
 & Acc.$\uparrow$ & Orig.\ acc.$\uparrow$ & ECE$\downarrow$ & Brier$\downarrow$ & \multicolumn{2}{c}{Partition$\downarrow$} & Order$\downarrow$ & \multicolumn{2}{c}{IIA$\downarrow$} \\
\cmidrule(lr){6-7}\cmidrule(l){9-10}
 & & & & & all & hard & & add & drop \\
\midrule
\multicolumn{10}{@{}l}{\textit{d3-nano (2.2B)}} \\
Base model & 67.9 & 67.3 & 5.6 & 41.9 & 14.3 & 28.9 & 6.2 & 7.6 & 8.5 \\
\rowcolor{oursrow}\method & 72.8 $\pm$ 0.9 & 70.6 $\pm$ 1.3 & 1.6 $\pm$ 1.3 & 35.8 $\pm$ 0.6 & 5.7 $\pm$ 0.8 & 10.2 $\pm$ 1.7 & 4.0 $\pm$ 0.5 & 4.0 $\pm$ 0.4 & 5.1 $\pm$ 0.5 \\
\multicolumn{10}{@{}l}{\textit{d3-mini (4.5B)}} \\
Base model & 71.6 & 71.9 & 3.4 & 37.3 & 16.4 & 38.3 & 6.0 & 5.8 & 14.0 \\
\rowcolor{oursrow}\method & 77.1 $\pm$ 0.6 & 75.1 $\pm$ 0.9 & 3.0 $\pm$ 0.8 & 31.0 $\pm$ 0.7 & 4.6 $\pm$ 0.6 & 8.1 $\pm$ 0.9 & 3.6 $\pm$ 0.2 & 3.7 $\pm$ 0.2 & 5.0 $\pm$ 0.3 \\
\multicolumn{10}{@{}l}{\textit{d3 (26B)}} \\
Base model & 78.2 & 78.4 & 1.7 & 28.4 & 10.2 & 23.5 & 6.3 & 5.7 & 9.9 \\
SFT$^\dagger$ & 83.4 $\pm$ 0.5 & 83.0 $\pm$ 2.1 & 7.8 $\pm$ 8.2 & 24.7 $\pm$ 3.8 & 6.3 $\pm$ 1.8 & 13.1 $\pm$ 3.7 & 5.0 $\pm$ 1.9 & 5.3 $\pm$ 2.9 & 9.3 $\pm$ 3.6 \\
\rowcolor{oursrow}\method & 81.4 $\pm$ 1.1 & 80.9 $\pm$ 1.1 & 3.2 $\pm$ 0.5 & 25.1 $\pm$ 1.4 & 4.1 $\pm$ 0.4 & 8.1 $\pm$ 1.1 & 3.8 $\pm$ 0.9 & 3.6 $\pm$ 0.6 & 5.2 $\pm$ 1.1 \\
\bottomrule
\end{tabularx}
\end{table*}}

\newcommand{\SuppTabDeltaSizes}{%
\begin{table*}[t]
\centering\scriptsize
\setlength{\tabcolsep}{2pt}
\caption{\textbf{Paired differences from \method, d3-nano, d3-mini and d3 (26B).} Method minus \method, $\times$100, with paired item-bootstrap 95\% intervals from 10{,}000 resamples after averaging over each method's seeds. Positive values mean higher accuracy, ECE, Brier score or partition violation than \method.}\label{supt:delta-sizes}
\begin{tabularx}{\textwidth}{@{}lYYYYYY@{}}
\toprule
 & Acc. & Orig.\ acc. & ECE & Brier & \multicolumn{2}{c}{Partition} \\
\cmidrule(l){6-7}
 & & & & & all & hard \\
\midrule
\multicolumn{7}{@{}l}{\textit{d3-nano (2.2B)}} \\
Base model & $-$4.9 [$-$5.9, $-$3.9] & $-$3.2 [$-$4.7, $-$1.8] & +4.0 [1.9, 4.5] & +6.1 [5.3, 6.9] & +8.6 [8.0, 9.2] & +18.7 [17.3, 20.2] \\
\multicolumn{7}{@{}l}{\textit{d3-mini (4.5B)}} \\
Base model & $-$5.5 [$-$6.7, $-$4.4] & $-$3.2 [$-$4.6, $-$2.0] & +0.4 [$-$2.2, 2.6] & +6.3 [5.2, 7.3] & +11.8 [10.9, 12.7] & +30.2 [28.0, 32.4] \\
\multicolumn{7}{@{}l}{\textit{d3 (26B)}} \\
Base model & $-$3.2 [$-$4.0, $-$2.4] & $-$2.5 [$-$3.7, $-$1.4] & $-$1.5 [$-$2.4, $-$0.4] & +3.4 [2.8, 3.9] & +6.1 [5.6, 6.7] & +15.4 [14.2, 16.7] \\
SFT$^\dagger$ & +2.0 [1.0, 3.0] & +2.1 [0.7, 3.5] & +4.6 [2.2, 6.4] & $-$0.3 [$-$1.7, 1.0] & +2.2 [1.7, 2.6] & +5.0 [3.8, 6.3] \\
\bottomrule
\end{tabularx}
\end{table*}}

\newcommand{\SuppTabAblLite}{%
\begin{table*}[t]
\centering\scriptsize
\setlength{\tabcolsep}{2pt}
\caption{\textbf{Ablations with intervals, d3-lite (0.85B).} Rows of main-paper Table 3. Values $\times$100; mean $\pm$ 95\% Student-$t$ interval over seeds (5 for \method, 3 for each variant). $\Delta$: paired difference from \method over the seeds that both runs share, with a 95\% Student-$t$ interval.}\label{supt:abl-lite}
\begin{tabularx}{\textwidth}{@{}lYYYYY@{}}
\toprule
 & Acc.$\uparrow$ & ECE$\downarrow$ & \multicolumn{2}{c}{Partition$\downarrow$} & Order$\downarrow$ \\
\cmidrule(lr){4-5}
 & & & all & hard & \\
\midrule
\rowcolor{oursrow}\method & 67.1 $\pm$ 0.8 & 1.7 $\pm$ 0.2 & 5.8 $\pm$ 0.2 & 8.5 $\pm$ 0.2 & 4.1 $\pm$ 0.2 \\
+ KL anchor & 66.9 $\pm$ 0.6 & 1.4 $\pm$ 0.6 & 5.8 $\pm$ 0.3 & 8.7 $\pm$ 0.8 & 4.1 $\pm$ 0.2 \\
\quad $\Delta$ (paired) & $-$0.60 $\pm$ 1.84 & $-$0.30 $\pm$ 0.97 & +0.00 $\pm$ 0.14 & +0.21 $\pm$ 0.92 & +0.05 $\pm$ 0.05 \\
per-question std & 67.1 $\pm$ 1.0 & 2.0 $\pm$ 0.3 & 6.1 $\pm$ 0.7 & 9.0 $\pm$ 1.1 & 4.1 $\pm$ 0.0 \\
\quad $\Delta$ (paired) & $-$0.36 $\pm$ 1.59 & +0.35 $\pm$ 0.52 & +0.27 $\pm$ 0.44 & +0.46 $\pm$ 1.24 & +0.11 $\pm$ 0.23 \\
$w=0$ (no label) & 64.3 $\pm$ 0.7 & 2.5 $\pm$ 0.4 & 5.5 $\pm$ 0.3 & 8.3 $\pm$ 0.9 & 3.8 $\pm$ 0.2 \\
\quad $\Delta$ (paired) & $-$3.17 $\pm$ 1.49 & +0.83 $\pm$ 0.08 & $-$0.26 $\pm$ 0.20 & $-$0.21 $\pm$ 1.09 & $-$0.17 $\pm$ 0.43 \\
$w=0.5$ & 68.6 $\pm$ 0.5 & 2.5 $\pm$ 0.4 & 6.4 $\pm$ 0.6 & 8.9 $\pm$ 0.7 & 4.4 $\pm$ 0.2 \\
\quad $\Delta$ (paired) & +1.05 $\pm$ 1.06 & +0.79 $\pm$ 0.51 & +0.60 $\pm$ 0.42 & +0.37 $\pm$ 0.85 & +0.41 $\pm$ 0.36 \\
$w=1$ (label only) & 68.2 $\pm$ 1.7 & 13.3 $\pm$ 3.1 & 11.8 $\pm$ 1.4 & 13.1 $\pm$ 1.9 & 8.0 $\pm$ 0.3 \\
\quad $\Delta$ (paired) & +0.66 $\pm$ 2.07 & +11.65 $\pm$ 3.41 & +5.98 $\pm$ 1.17 & +4.62 $\pm$ 2.10 & +3.99 $\pm$ 0.44 \\
$w$ from 0 to 0.6 & 68.2 $\pm$ 0.6 & 3.2 $\pm$ 1.6 & 6.5 $\pm$ 0.3 & 8.9 $\pm$ 1.2 & 4.3 $\pm$ 0.3 \\
\quad $\Delta$ (paired) & +0.66 $\pm$ 1.22 & +1.52 $\pm$ 1.27 & +0.74 $\pm$ 0.10 & +0.33 $\pm$ 1.12 & +0.28 $\pm$ 0.47 \\
exact $\KL(p_f\Vert q_f)$ & 67.3 $\pm$ 0.7 & 1.5 $\pm$ 0.9 & 5.8 $\pm$ 0.3 & 8.5 $\pm$ 0.4 & 4.0 $\pm$ 0.2 \\
\quad $\Delta$ (paired) & $-$0.18 $\pm$ 1.29 & $-$0.19 $\pm$ 1.37 & $-$0.01 $\pm$ 0.18 & $-$0.06 $\pm$ 0.47 & +0.02 $\pm$ 0.18 \\
cross-entropy to $q_f$ & 67.8 $\pm$ 0.4 & 2.0 $\pm$ 1.9 & 5.7 $\pm$ 0.2 & 8.5 $\pm$ 0.5 & 3.9 $\pm$ 0.2 \\
\quad $\Delta$ (paired) & +0.34 $\pm$ 1.00 & +0.33 $\pm$ 2.25 & $-$0.03 $\pm$ 0.26 & $-$0.06 $\pm$ 0.47 & $-$0.07 $\pm$ 0.18 \\
\bottomrule
\end{tabularx}
\end{table*}}

\newcommand{\SuppTabAbl}{%
\begin{table*}[t]
\centering\scriptsize
\setlength{\tabcolsep}{2pt}
\caption{\textbf{Ablations at the sizes other than d3-lite.} Values $\times$100; mean $\pm$ 95\% Student-$t$ interval over seeds (3 for \method); the KL anchor has 2 seeds at d3-nano and d3-mini. $\Delta$: paired difference from \method over the seeds that both runs share, with a 95\% Student-$t$ interval.}\label{supt:abl}
\begin{tabularx}{\textwidth}{@{}lYYYYY@{}}
\toprule
 & Acc.$\uparrow$ & ECE$\downarrow$ & \multicolumn{2}{c}{Partition$\downarrow$} & Order$\downarrow$ \\
\cmidrule(lr){4-5}
 & & & all & hard & \\
\midrule
\multicolumn{6}{@{}l}{\textit{d3-nano (2.2B)}} \\
\rowcolor{oursrow}\method & 72.8 $\pm$ 0.9 & 1.6 $\pm$ 1.3 & 5.7 $\pm$ 0.8 & 10.2 $\pm$ 1.7 & 4.0 $\pm$ 0.5 \\
+ KL anchor & 72.1 $\pm$ 2.5 & 1.5 $\pm$ 2.0 & 5.8 $\pm$ 2.1 & 10.5 $\pm$ 6.0 & 4.2 $\pm$ 0.8 \\
\quad $\Delta$ (paired) & $-$0.84 $\pm$ 0.24 & +0.23 $\pm$ 0.48 & +0.24 $\pm$ 0.95 & +0.59 $\pm$ 0.41 & +0.34 $\pm$ 2.72 \\
\multicolumn{6}{@{}l}{\textit{d3-mini (4.5B)}} \\
\rowcolor{oursrow}\method & 77.1 $\pm$ 0.6 & 3.0 $\pm$ 0.8 & 4.6 $\pm$ 0.6 & 8.1 $\pm$ 0.9 & 3.6 $\pm$ 0.2 \\
+ KL anchor & 76.4 $\pm$ 3.1 & 2.6 $\pm$ 0.9 & 4.9 $\pm$ 2.1 & 9.0 $\pm$ 1.6 & 4.1 $\pm$ 1.1 \\
\quad $\Delta$ (paired) & $-$0.59 $\pm$ 0.64 & $-$0.59 $\pm$ 2.29 & +0.28 $\pm$ 0.99 & +0.91 $\pm$ 2.70 & +0.59 $\pm$ 1.36 \\
\multicolumn{6}{@{}l}{\textit{d3-flash (8.4B)}} \\
\rowcolor{oursrow}\method & 77.2 $\pm$ 0.5 & 2.3 $\pm$ 0.8 & 4.8 $\pm$ 0.4 & 9.2 $\pm$ 0.9 & 3.8 $\pm$ 0.2 \\
+ KL anchor & 76.7 $\pm$ 0.6 & 2.2 $\pm$ 1.1 & 5.0 $\pm$ 0.1 & 10.0 $\pm$ 0.7 & 4.2 $\pm$ 0.3 \\
\quad $\Delta$ (paired) & $-$0.43 $\pm$ 0.49 & $-$0.11 $\pm$ 0.31 & +0.19 $\pm$ 0.40 & +0.75 $\pm$ 1.09 & +0.42 $\pm$ 0.42 \\
exact $\KL(p_f\Vert q_f)$ & 77.4 $\pm$ 0.6 & 2.6 $\pm$ 0.1 & 4.6 $\pm$ 0.2 & 8.8 $\pm$ 0.3 & 3.7 $\pm$ 0.2 \\
\quad $\Delta$ (paired) & +0.27 $\pm$ 0.21 & +0.28 $\pm$ 0.72 & $-$0.19 $\pm$ 0.40 & $-$0.40 $\pm$ 1.13 & $-$0.05 $\pm$ 0.32 \\
\multicolumn{6}{@{}l}{\textit{d3 (26B)}} \\
\rowcolor{oursrow}\method & 81.4 $\pm$ 1.1 & 3.2 $\pm$ 0.5 & 4.1 $\pm$ 0.4 & 8.1 $\pm$ 1.1 & 3.8 $\pm$ 0.9 \\
+ KL anchor & 81.1 $\pm$ 0.6 & 3.2 $\pm$ 0.9 & 4.3 $\pm$ 0.8 & 8.9 $\pm$ 2.2 & 4.4 $\pm$ 1.1 \\
\quad $\Delta$ (paired) & $-$0.33 $\pm$ 0.49 & +0.07 $\pm$ 0.45 & +0.25 $\pm$ 0.57 & +0.77 $\pm$ 1.48 & +0.57 $\pm$ 0.37 \\
\bottomrule
\end{tabularx}
\end{table*}}

\newcommand{\SuppTabConf}{%
\begin{table*}[t]
\centering\scriptsize
\setlength{\tabcolsep}{2pt}
\caption{\textbf{Partition violation within bins of confidence, all sizes.} Each item is binned by the method's own top probability on the original question; numbers in parentheses count the item-seed pairs in the bin. Values $\times$100. SFT is trained as published and SFT$^\dagger$ on our regrouped questions.}\label{supt:conf}
\begin{tabularx}{\textwidth}{@{}lYYYYY@{}}
\toprule
 & \multicolumn{4}{c}{Partition violation of items whose top probability is in} & Mean top \\
\cmidrule(lr){2-5}
 & $[0, 0.5)$ & $[0.5, 0.7)$ & $[0.7, 0.9)$ & $[0.9, 1]$ & probability \\
\midrule
\multicolumn{6}{@{}l}{\textit{d3-lite (0.85B)}} \\
Base model & 16.0 (357) & 17.4 (255) & 15.8 (198) & 9.3 (240) & 63.7 \\
SFT & 28.9 (238) & 30.1 (435) & 27.6 (543) & 13.4 (1934) & 85.7 \\
SFT$^\dagger$ & 21.3 (304) & 20.6 (541) & 16.0 (593) & 3.4 (1712) & 82.8 \\
\rowcolor{oursrow}\method & 7.8 (1772) & 7.2 (1297) & 5.2 (1023) & 1.9 (1158) & 63.8 \\
\multicolumn{6}{@{}l}{\textit{d3-nano (2.2B)}} \\
Base model & 19.7 (271) & 17.3 (217) & 17.2 (199) & 6.9 (363) & 70.9 \\
\rowcolor{oursrow}\method & 9.3 (795) & 8.5 (624) & 6.0 (650) & 1.4 (1081) & 71.1 \\
\multicolumn{6}{@{}l}{\textit{d3-mini (4.5B)}} \\
Base model & 28.3 (258) & 20.4 (215) & 16.2 (207) & 5.9 (370) & 71.8 \\
\rowcolor{oursrow}\method & 7.6 (723) & 7.0 (639) & 4.7 (685) & 1.3 (1103) & 72.1 \\
\multicolumn{6}{@{}l}{\textit{d3-flash (8.4B)}} \\
Base model & 26.5 (237) & 20.9 (210) & 16.4 (210) & 4.2 (393) & 73.4 \\
SFT & 34.9 (111) & 31.6 (279) & 28.5 (335) & 9.6 (2425) & 91.7 \\
SFT$^\dagger$ & 29.8 (117) & 24.2 (288) & 19.4 (346) & 2.4 (2399) & 91.3 \\
\rowcolor{oursrow}\method & 8.4 (667) & 7.4 (546) & 5.4 (754) & 1.3 (1183) & 74.1 \\
\multicolumn{6}{@{}l}{\textit{d3 (26B)}} \\
Base model & 18.8 (206) & 14.5 (196) & 12.6 (213) & 3.1 (435) & 75.1 \\
SFT$^\dagger$ & 21.9 (158) & 20.2 (333) & 14.2 (399) & 1.7 (2260) & 89.6 \\
\rowcolor{oursrow}\method & 8.9 (592) & 7.0 (517) & 4.1 (660) & 0.9 (1381) & 76.2 \\
\bottomrule
\end{tabularx}
\end{table*}}

\newcommand{\SuppTabGap}{%
\begin{table*}[t]
\centering\scriptsize
\setlength{\tabcolsep}{2pt}
\caption{\textbf{Partition violation and additivity error of the base models.} Additivity error: a merged option's probability on the regrouped question minus the sum of its members' probabilities on the original question. 1{,}050 items; values $\times$100 with item-bootstrap 95\% intervals (10{,}000 resamples).}\label{supt:gap}
\begin{tabularx}{\textwidth}{@{}lYYYY@{}}
\toprule
 & Partition & \multicolumn{3}{c}{Additivity error of merged options} \\
\cmidrule(l){3-5}
 & violation & all & with the correct option & without the correct option \\
\midrule
d3-lite (0.85B) & 14.8 [14.2, 15.3] & $-$11.5 [$-$12.1, $-$10.9] & $-$14.4 [$-$15.3, $-$13.5] & $-$8.5 [$-$9.3, $-$7.7] \\
d3-nano (2.2B) & 14.3 [13.6, 15.0] & $-$9.0 [$-$9.7, $-$8.4] & $-$12.0 [$-$13.0, $-$11.0] & $-$6.0 [$-$6.9, $-$5.1] \\
d3-mini (4.5B) & 16.4 [15.4, 17.4] & $-$9.3 [$-$10.0, $-$8.7] & $-$11.9 [$-$12.9, $-$10.9] & $-$6.8 [$-$7.7, $-$5.9] \\
d3-flash (8.4B) & 15.0 [14.1, 16.0] & $-$8.7 [$-$9.4, $-$8.1] & $-$11.1 [$-$12.1, $-$10.1] & $-$6.4 [$-$7.2, $-$5.5] \\
d3 (26B) & 10.2 [9.6, 10.9] & $-$5.4 [$-$5.8, $-$4.9] & $-$6.7 [$-$7.6, $-$6.0] & $-$4.0 [$-$4.6, $-$3.4] \\
\bottomrule
\end{tabularx}
\end{table*}}

\newcommand{\SuppTabParse}{\begin{table*}[t]
\centering\small
\setlength{\tabcolsep}{5pt}
\caption{\textbf{Parse-and-sum against \method.} Parse-and-sum recovers the members of each presented option from its text where the parser succeeds (\cref{sup:parse}), asks the original question and sums the members' probabilities. With 3 orders, it also replaces the model's answer to the original question by the average over three orders. Passes: forward passes per regrouped question (parse-and-sum asks the original question instead of the regrouped one). Add questions (10.9\% of all questions) keep the model's own answer. Values $\times$100; \method{} rows are means over 5 seeds on d3-lite and 3 on d3-flash; paired differences with intervals are in \cref{supt:parse-delta}. Part.: partition violation; for parse-and-sum, it is 0 by construction with 3 orders and at most the model's order floor with 1 order (d3-lite values in \cref{tab:floor}), so this column is not compared across rows.}
\label{supt:parse}
\begin{tabularx}{\textwidth}{@{}l c *{4}{Y} @{\hspace{12pt}} *{4}{Y}@{}}
\toprule
 & & \multicolumn{4}{c}{d3-lite} & \multicolumn{4}{c}{d3-flash} \\
\cmidrule(lr){3-6}\cmidrule(l){7-10}
Method & Passes & Acc.\,$\uparrow$ & ECE\,$\downarrow$ & Brier\,$\downarrow$ & Part. & Acc.\,$\uparrow$ & ECE\,$\downarrow$ & Brier\,$\downarrow$ & Part. \\
\midrule
base model & 1 & 60.8 & 5.6 & 48.4 & 14.8 & 72.6 & 3.5 & 35.3 & 15.0 \\
\quad + parse-and-sum, 1 order & 1 & 63.2 & 4.5 & 46.1 & 2.5 & 74.4 & 2.8 & 33.2 & 2.8 \\
\quad + parse-and-sum, 3 orders & 3 & 63.6 & 4.2 & 45.9 & 0.0 & 75.5 & \textbf{1.6} & 32.8 & 0.0 \\
\midrule
\rowcolor{oursrow}\method & 1 & 67.1 & 1.7 & 42.4 & 5.8 & 77.2 & 2.3 & 30.2 & 4.8 \\
\quad + parse-and-sum, 1 order & 1 & 67.5 & \textbf{1.6} & \textbf{41.8} & 1.9 & \textbf{77.5} & 2.4 & 29.9 & 1.8 \\
\quad + parse-and-sum, 3 orders & 3 & \textbf{67.7} & 2.1 & \textbf{41.8} & 0.0 & 77.4 & 2.7 & \textbf{29.7} & 0.0 \\
\bottomrule
\end{tabularx}
\end{table*}}
\newcommand{\SuppTabParseDelta}{\begin{table*}[t]
\centering\small
\setlength{\tabcolsep}{4pt}
\caption{Paired differences for \cref{supt:parse}, $\times$100, with 95\% item-bootstrap intervals (10{,}000 resamples; seeds averaged first). (1) and (3) denote parse-and-sum over 1 and 3 orders. Positive Acc.\ and negative ECE and Brier differences favor the first term.}
\label{supt:parse-delta}
\begin{tabularx}{\textwidth}{@{}l *{6}{Y}@{}}
\toprule
 & \multicolumn{3}{c}{d3-lite} & \multicolumn{3}{c}{d3-flash} \\
\cmidrule(lr){2-4}\cmidrule(l){5-7}
Difference & Acc. & ECE & Brier & Acc. & ECE & Brier \\
\midrule
\method{} $-$ base (3) & +3.6 {\scriptsize[+2.5, +4.7]} & $-$2.5 {\scriptsize[$-$3.4, $-$0.4]} & $-$3.4 {\scriptsize[$-$4.2, $-$2.7]} & +1.7 {\scriptsize[+0.9, +2.6]} & +0.7 {\scriptsize[$-$1.1, +1.4]} & $-$2.6 {\scriptsize[$-$3.2, $-$2.0]} \\
\method{} $-$ base (1) & +4.0 {\scriptsize[+2.9, +5.0]} & $-$2.9 {\scriptsize[$-$3.7, $-$0.5]} & $-$3.6 {\scriptsize[$-$4.4, $-$2.9]} & +2.8 {\scriptsize[+1.9, +3.7]} & $-$0.5 {\scriptsize[$-$1.8, +1.1]} & $-$3.1 {\scriptsize[$-$3.7, $-$2.4]} \\
\method{} (3) $-$ \method & +0.5 {\scriptsize[+0.1, +1.0]} & +0.4 {\scriptsize[+0.1, +1.2]} & $-$0.6 {\scriptsize[$-$0.9, $-$0.4]} & +0.2 {\scriptsize[$-$0.2, +0.6]} & +0.4 {\scriptsize[+0.1, +1.2]} & $-$0.4 {\scriptsize[$-$0.6, $-$0.2]} \\
\bottomrule
\end{tabularx}
\end{table*}}
\newcommand{\NPsMerged}{3{,}427}
\newcommand{\NPsMergedOk}{3{,}427}
\newcommand{\NPsWordMin}{99.97}
\newcommand{\NPsOrOnlyMerged}{0}
\newcommand{\NPsApplicable}{82.4}
\newcommand{\NPsLiteBaseAcc}{60.8}
\newcommand{\NPsLiteBaseEce}{5.6}
\newcommand{\NPsLiteBaseBrier}{48.4}
\newcommand{\NPsLiteBaseAccOrig}{59.6}
\newcommand{\NPsLiteBasePsAcc}{63.6}
\newcommand{\NPsLiteBasePsEce}{4.2}
\newcommand{\NPsLiteBasePsBrier}{45.9}
\newcommand{\NPsLiteBasePsAccOrig}{59.6}
\newcommand{\NPsLiteCoreAcc}{67.1}
\newcommand{\NPsLiteCoreEce}{1.7}
\newcommand{\NPsLiteCoreBrier}{42.4}
\newcommand{\NPsLiteCoreAccOrig}{64.6}
\newcommand{\NPsLiteCorePsAcc}{67.7}
\newcommand{\NPsLiteCorePsEce}{2.1}
\newcommand{\NPsLiteCorePsBrier}{41.8}
\newcommand{\NPsLiteCorePsAccOrig}{64.6}
\newcommand{\NPsLiteDAcc}{+3.6}
\newcommand{\NPsLiteDAccAbs}{3.6}
\newcommand{\NPsLiteDAccCI}{[+2.5, +4.7]}
\newcommand{\NPsLiteDEce}{$-$2.5}
\newcommand{\NPsLiteDEceAbs}{2.5}
\newcommand{\NPsLiteDEceCI}{[$-$3.4, $-$0.4]}
\newcommand{\NPsLiteDBrier}{$-$3.4}
\newcommand{\NPsLiteDBrierAbs}{3.4}
\newcommand{\NPsLiteDBrierCI}{[$-$4.2, $-$2.7]}
\newcommand{\NPsLiteDCAcc}{+0.5}
\newcommand{\NPsLiteDCAccAbs}{0.5}
\newcommand{\NPsLiteDCAccCI}{[+0.1, +1.0]}
\newcommand{\NPsLiteDCEce}{+0.4}
\newcommand{\NPsLiteDCEceAbs}{0.4}
\newcommand{\NPsLiteDCEceCI}{[+0.1, +1.2]}
\newcommand{\NPsLiteDCBrier}{$-$0.6}
\newcommand{\NPsLiteDCBrierAbs}{0.6}
\newcommand{\NPsLiteDCBrierCI}{[$-$0.9, $-$0.4]}
\newcommand{\NPsFlashBaseAcc}{72.6}
\newcommand{\NPsFlashBaseEce}{3.5}
\newcommand{\NPsFlashBaseBrier}{35.3}
\newcommand{\NPsFlashBaseAccOrig}{73.2}
\newcommand{\NPsFlashBasePsAcc}{75.5}
\newcommand{\NPsFlashBasePsEce}{1.6}
\newcommand{\NPsFlashBasePsBrier}{32.8}
\newcommand{\NPsFlashBasePsAccOrig}{73.2}
\newcommand{\NPsFlashCoreAcc}{77.2}
\newcommand{\NPsFlashCoreEce}{2.3}
\newcommand{\NPsFlashCoreBrier}{30.2}
\newcommand{\NPsFlashCoreAccOrig}{75.3}
\newcommand{\NPsFlashCorePsAcc}{77.4}
\newcommand{\NPsFlashCorePsEce}{2.7}
\newcommand{\NPsFlashCorePsBrier}{29.7}
\newcommand{\NPsFlashCorePsAccOrig}{75.3}
\newcommand{\NPsFlashDAcc}{+1.7}
\newcommand{\NPsFlashDAccAbs}{1.7}
\newcommand{\NPsFlashDAccCI}{[+0.9, +2.6]}
\newcommand{\NPsFlashDEce}{+0.7}
\newcommand{\NPsFlashDEceAbs}{0.7}
\newcommand{\NPsFlashDEceCI}{[$-$1.1, +1.4]}
\newcommand{\NPsFlashDBrier}{$-$2.6}
\newcommand{\NPsFlashDBrierAbs}{2.6}
\newcommand{\NPsFlashDBrierCI}{[$-$3.2, $-$2.0]}
\newcommand{\NPsFlashDCAcc}{+0.2}
\newcommand{\NPsFlashDCAccAbs}{0.2}
\newcommand{\NPsFlashDCAccCI}{[$-$0.2, +0.6]}
\newcommand{\NPsFlashDCEce}{+0.4}
\newcommand{\NPsFlashDCEceAbs}{0.4}
\newcommand{\NPsFlashDCEceCI}{[+0.1, +1.2]}
\newcommand{\NPsFlashDCBrier}{$-$0.4}
\newcommand{\NPsFlashDCBrierAbs}{0.4}
\newcommand{\NPsFlashDCBrierCI}{[$-$0.6, $-$0.2]}
\newcommand{\NPsLiteBasePsOnePart}{2.5}
\newcommand{\NPsLiteCorePsOnePart}{1.9}
\newcommand{\SuppTabParseTrained}{\begin{table*}[t]
\centering\small
\setlength{\tabcolsep}{4pt}
\caption{\textbf{Parse-and-sum applied to the trained baselines and to \method} (members from the text parser, as in \cref{supt:parse}). Values $\times$100; mean $\pm$ 95\% Student-$t$ interval over seeds (3 per baseline, 5 for \method{} on d3-lite, 3 on d3-flash). With 3 orders, partition violation is 0 by construction on the regrouped questions it applies to. ACRE$^\dagger$ and PA-GRPO$^\dagger$ were not trained on d3-flash.}\label{supt:parse-trained}
\begin{tabularx}{\textwidth}{@{}>{\raggedright\arraybackslash}X*{6}{c}@{}}
\toprule
 & \multicolumn{3}{c}{d3-lite} & \multicolumn{3}{c}{d3-flash} \\
\cmidrule(lr){2-4}\cmidrule(lr){5-7}
Method & Acc.$\uparrow$ & ECE$\downarrow$ & Brier$\downarrow$ & Acc.$\uparrow$ & ECE$\downarrow$ & Brier$\downarrow$ \\
\midrule
\multicolumn{7}{@{}l}{\textit{parse-and-sum over 1 order (1 forward pass)}} \\
SFT & 68.9 {\scriptsize$\pm$2.5} & 19.6 {\scriptsize$\pm$4.8} & 48.4 {\scriptsize$\pm$1.2} & 80.6 {\scriptsize$\pm$0.5} & 12.8 {\scriptsize$\pm$1.8} & 31.1 {\scriptsize$\pm$2.5} \\
SFT$^\dagger$ & 70.0 {\scriptsize$\pm$1.1} & 15.2 {\scriptsize$\pm$4.7} & 44.5 {\scriptsize$\pm$3.4} & 81.1 {\scriptsize$\pm$0.8} & 11.6 {\scriptsize$\pm$0.2} & 29.6 {\scriptsize$\pm$1.5} \\
LLM-as-Jev & 69.0 {\scriptsize$\pm$2.7} & 2.4 {\scriptsize$\pm$1.0} & 40.3 {\scriptsize$\pm$2.4} & 79.2 {\scriptsize$\pm$1.7} & 2.5 {\scriptsize$\pm$1.4} & 27.3 {\scriptsize$\pm$1.9} \\
LLM-as-Jev$^\dagger$ & 69.7 {\scriptsize$\pm$1.0} & 3.2 {\scriptsize$\pm$1.7} & 39.9 {\scriptsize$\pm$0.5} & 79.0 {\scriptsize$\pm$1.3} & 2.2 {\scriptsize$\pm$1.6} & 28.0 {\scriptsize$\pm$1.4} \\
ACRE & 68.3 {\scriptsize$\pm$2.2} & 6.2 {\scriptsize$\pm$4.2} & 41.7 {\scriptsize$\pm$4.3} & 77.9 {\scriptsize$\pm$1.4} & 3.8 {\scriptsize$\pm$2.4} & 30.1 {\scriptsize$\pm$2.3} \\
ACRE$^\dagger$ & 65.8 {\scriptsize$\pm$0.7} & 8.6 {\scriptsize$\pm$2.5} & 44.6 {\scriptsize$\pm$1.2} & -- & -- & -- \\
PA-GRPO & 65.1 {\scriptsize$\pm$1.4} & 16.0 {\scriptsize$\pm$1.8} & 49.9 {\scriptsize$\pm$2.6} & 77.6 {\scriptsize$\pm$1.3} & 9.3 {\scriptsize$\pm$2.2} & 33.0 {\scriptsize$\pm$3.4} \\
PA-GRPO$^\dagger$ & 64.9 {\scriptsize$\pm$0.9} & 16.5 {\scriptsize$\pm$1.1} & 49.4 {\scriptsize$\pm$0.7} & -- & -- & -- \\
\method & 67.5 {\scriptsize$\pm$0.8} & 1.6 {\scriptsize$\pm$0.2} & 41.8 {\scriptsize$\pm$0.6} & 77.5 {\scriptsize$\pm$1.2} & 2.4 {\scriptsize$\pm$1.1} & 29.9 {\scriptsize$\pm$0.7} \\
\multicolumn{7}{@{}l}{\textit{parse-and-sum over 3 orders (3 forward passes)}} \\
SFT & 69.7 {\scriptsize$\pm$2.7} & 17.4 {\scriptsize$\pm$4.4} & 46.9 {\scriptsize$\pm$1.2} & 80.6 {\scriptsize$\pm$0.5} & 11.9 {\scriptsize$\pm$1.9} & 30.0 {\scriptsize$\pm$2.5} \\
SFT$^\dagger$ & 70.4 {\scriptsize$\pm$1.7} & 14.0 {\scriptsize$\pm$5.4} & 43.6 {\scriptsize$\pm$3.6} & 81.0 {\scriptsize$\pm$2.3} & 11.0 {\scriptsize$\pm$1.8} & 28.9 {\scriptsize$\pm$1.7} \\
LLM-as-Jev & 69.6 {\scriptsize$\pm$2.6} & 3.3 {\scriptsize$\pm$0.3} & 40.0 {\scriptsize$\pm$2.2} & 79.3 {\scriptsize$\pm$1.7} & 2.9 {\scriptsize$\pm$1.4} & 27.1 {\scriptsize$\pm$2.0} \\
LLM-as-Jev$^\dagger$ & 69.7 {\scriptsize$\pm$1.4} & 3.4 {\scriptsize$\pm$2.4} & 39.7 {\scriptsize$\pm$0.3} & 79.4 {\scriptsize$\pm$1.6} & 2.7 {\scriptsize$\pm$1.9} & 27.8 {\scriptsize$\pm$1.5} \\
ACRE & 68.9 {\scriptsize$\pm$2.8} & 5.3 {\scriptsize$\pm$2.2} & 40.9 {\scriptsize$\pm$3.2} & 78.1 {\scriptsize$\pm$0.7} & 3.0 {\scriptsize$\pm$1.6} & 29.6 {\scriptsize$\pm$2.4} \\
ACRE$^\dagger$ & 66.1 {\scriptsize$\pm$1.0} & 8.0 {\scriptsize$\pm$2.7} & 44.4 {\scriptsize$\pm$1.1} & -- & -- & -- \\
PA-GRPO & 65.3 {\scriptsize$\pm$1.5} & 15.0 {\scriptsize$\pm$1.6} & 48.6 {\scriptsize$\pm$2.4} & 77.6 {\scriptsize$\pm$1.4} & 8.5 {\scriptsize$\pm$2.3} & 32.3 {\scriptsize$\pm$3.6} \\
PA-GRPO$^\dagger$ & 65.1 {\scriptsize$\pm$0.5} & 15.7 {\scriptsize$\pm$0.8} & 48.9 {\scriptsize$\pm$0.6} & -- & -- & -- \\
\method & 67.7 {\scriptsize$\pm$0.6} & 2.1 {\scriptsize$\pm$0.5} & 41.8 {\scriptsize$\pm$0.6} & 77.4 {\scriptsize$\pm$0.9} & 2.7 {\scriptsize$\pm$0.8} & 29.7 {\scriptsize$\pm$0.7} \\
\midrule
\rowcolor{oursrow}\method{} (no parse-and-sum) & 67.1 {\scriptsize$\pm$0.8} & 1.7 {\scriptsize$\pm$0.2} & 42.4 {\scriptsize$\pm$0.6} & 77.2 {\scriptsize$\pm$0.5} & 2.3 {\scriptsize$\pm$0.8} & 30.2 {\scriptsize$\pm$0.5} \\
\bottomrule
\end{tabularx}
\end{table*}}
\newcommand{\NPsbLjevDLiteAcc}{2.5}
\newcommand{\NPsbLjevDLiteAccCI}{[1.5, 3.5]}
\newcommand{\NPsbLjevDLiteEce}{1.7}
\newcommand{\NPsbLjevDLiteEceCI}{[0.1, 2.7]}
\newcommand{\NPsbLjevDLiteBrier}{$-$2.7}
\newcommand{\NPsbLjevDLiteBrierCI}{[$-$3.5, $-$1.9]}
\newcommand{\NPsbLiteSftAccOrig}{67.1}
\newcommand{\NPsbLiteLjevAccOrig}{66.9}
\newcommand{\NPsbLiteAcreAccOrig}{66.1}
\newcommand{\NPsbLiteCoreAccOrig}{64.6}
\newcommand{\NPsbLjevDFlashAcc}{2.2}
\newcommand{\NPsbLjevDFlashAccCI}{[1.3, 3.2]}
\newcommand{\NPsbLjevDFlashEce}{0.4}
\newcommand{\NPsbLjevDFlashEceCI}{[$-$0.6, 1.5]}
\newcommand{\NPsbLjevDFlashBrier}{$-$2.4}
\newcommand{\NPsbLjevDFlashBrierCI}{[$-$3.1, $-$1.6]}
\newcommand{\NPsbFlashSftAccOrig}{79.0}
\newcommand{\NPsbFlashLjevAccOrig}{77.4}
\newcommand{\NPsbFlashAcreAccOrig}{76.1}
\newcommand{\NPsbFlashCoreAccOrig}{75.3}
\newcommand{\NPsbLiteEceMinName}{LLM-as-Jev}
\newcommand{\NPsbLiteEceMin}{1.6}
\newcommand{\NPsbLiteEceMinCI}{[0.0, 2.7]}
\newcommand{\SuppTabTune}{\begin{table}[t]
\centering\small
\setlength{\tabcolsep}{3pt}
\caption{Learning rates tried for ACRE and PA-GRPO (d3-lite, seed 0, step 600). The setting with the lowest partition violation, marked $\ast$, was kept and run with two more seeds (\cref{sup:baselines}). $\times$100.}\label{supt:tune}
\begin{tabularx}{\linewidth}{@{}>{\raggedright\arraybackslash}X c cccc@{}}
\toprule
Method & lr & Acc. & Orig.\ acc. & ECE & Partition \\
\midrule
ACRE & $10^{-5}$ & 63.9 & 63.8 & 10.8 & 15.4 \\
ACRE & $3\cdot10^{-5}$ & 65.5 & 65.5 & 7.9 & 14.8 \\
ACRE & $10^{-4}$$^\ast$ & 66.0 & 65.0 & 6.5 & 14.0 \\
ACRE$^\dagger$ & $10^{-5}$$^\ast$ & 65.5 & 63.4 & 11.1 & 11.8 \\
ACRE$^\dagger$ & $3\cdot10^{-5}$ & 67.2 & 65.0 & 10.6 & 17.7 \\
ACRE$^\dagger$ & $10^{-4}$ & 66.9 & 65.7 & 15.7 & 26.4 \\
PA-GRPO & $10^{-5}$ & 63.0 & 62.8 & 17.3 & 18.1 \\
PA-GRPO & $3\cdot10^{-5}$ & 63.5 & 62.7 & 17.7 & 17.6 \\
PA-GRPO & $10^{-4}$$^\ast$ & 63.4 & 61.2 & 17.6 & 15.2 \\
PA-GRPO$^\dagger$ & $10^{-5}$$^\ast$ & 64.8 & 61.9 & 17.6 & 11.8 \\
PA-GRPO$^\dagger$ & $3\cdot10^{-5}$ & 65.7 & 63.5 & 16.6 & 13.3 \\
PA-GRPO$^\dagger$ & $10^{-4}$ & 66.2 & 63.4 & 16.8 & 14.5 \\
\bottomrule
\end{tabularx}
\end{table}}
\newcommand{\SuppTabMetrics}{\begin{table*}[t]
\centering\small
\setlength{\tabcolsep}{4pt}
\caption{\textbf{ECE binning and partition violation by regrouping type.} Each row gives \method, the base model, the trained baseline (published or $^\dagger$ setting) with the lowest value, and \method minus that baseline with its 95\% paired item-bootstrap interval (10{,}000 resamples); training-free corrections are discussed in \cref{sup:metrics}. Values are $\times$100 (lower is better) and means over 3 seeds per baseline and over 5 (d3-lite) or 3 (d3-flash) for \method. Adaptive ECE uses 15 bins with equal numbers of questions. There are 2{,}100 pair-merge and 1{,}050 split questions. $^*$: the variant used in the main paper.}
\label{supt:metrics}
\begin{tabularx}{\textwidth}{@{}l *{2}{c} >{\raggedright\arraybackslash}X c @{\hspace{10pt}} *{2}{c} >{\raggedright\arraybackslash}X c@{}}
\toprule
 & \multicolumn{4}{c}{d3-lite} & \multicolumn{4}{c}{d3-flash} \\
\cmidrule(lr){2-5}\cmidrule(l){6-9}
Metric & \method & Base & Best trained baseline & \method $-$ it & \method & Base & Best trained baseline & \method $-$ it \\
\midrule
ECE, 10 bins & 1.6 & 5.6 & LLM-as-Jev (1.7) & $-$0.2 {\scriptsize[$-$0.9, +1.0]} & 2.2 & 3.5 & LLM-as-Jev (1.8) & +0.5 {\scriptsize[$-$0.4, +1.4]} \\
ECE, 15 bins$^*$ & 1.7 & 5.6 & LLM-as-Jev (1.9) & $-$0.2 {\scriptsize[$-$1.0, +0.9]} & 2.3 & 3.5 & LLM-as-Jev (1.8) & +0.5 {\scriptsize[$-$0.3, +1.4]} \\
ECE, 20 bins & 2.1 & 5.6 & LLM-as-Jev (2.3) & $-$0.2 {\scriptsize[$-$1.0, +0.8]} & 2.6 & 3.6 & LLM-as-Jev (2.2) & +0.4 {\scriptsize[$-$0.4, +1.2]} \\
Adaptive ECE, 15 bins & 1.8 & 5.6 & LLM-as-Jev (2.2) & $-$0.4 {\scriptsize[$-$1.0, +0.8]} & 2.3 & 3.8 & LLM-as-Jev (1.8) & +0.4 {\scriptsize[$-$0.4, +1.4]} \\
\midrule
Partition, pooled$^*$ & 5.8 & 14.8 & LLM-as-Jev$^\dagger$ (9.4) & $-$3.5 {\scriptsize[$-$3.8, $-$3.3]} & 4.8 & 15.0 & SFT$^\dagger$ (7.3) & $-$2.4 {\scriptsize[$-$3.0, $-$1.9]} \\
Partition, pair merges & 5.7 & 14.0 & LLM-as-Jev$^\dagger$ (9.3) & $-$3.5 {\scriptsize[$-$3.8, $-$3.3]} & 4.7 & 15.4 & SFT$^\dagger$ (7.2) & $-$2.5 {\scriptsize[$-$3.1, $-$1.9]} \\
Partition, splits & 6.1 & 16.4 & LLM-as-Jev$^\dagger$ (9.6) & $-$3.5 {\scriptsize[$-$3.9, $-$3.1]} & 5.1 & 14.2 & SFT$^\dagger$ (7.4) & $-$2.4 {\scriptsize[$-$3.1, $-$1.6]} \\
\bottomrule
\end{tabularx}
\end{table*}}
\newcommand{\NMTsTen}{1.5}
\newcommand{\NMCoreTen}{1.6}
\newcommand{\NMTsDTen}{+0.1}
\newcommand{\NMTsDTenCI}{[$-$1.2, +1.2]}
\newcommand{\NMTsFifteen}{1.9}
\newcommand{\NMCoreFifteen}{1.7}
\newcommand{\NMTsDFifteen}{$-$0.2}
\newcommand{\NMTsDFifteenCI}{[$-$1.3, +1.0]}
\newcommand{\NMTsTwenty}{1.7}
\newcommand{\NMCoreTwenty}{2.1}
\newcommand{\NMTsDTwenty}{+0.3}
\newcommand{\NMTsDTwentyCI}{[$-$1.0, +1.2]}
\newcommand{\NMTsAdapt}{1.7}
\newcommand{\NMCoreAdapt}{1.8}
\newcommand{\NMTsDAdapt}{+0.1}
\newcommand{\NMTsDAdaptCI}{[$-$1.1, +1.1]}
\newcommand{\NMFlOaTen}{1.8}
\newcommand{\NMFlCoreTen}{2.2}
\newcommand{\NMFlOaDTen}{$-$0.5}
\newcommand{\NMFlOaDTenCI}{[$-$1.8, +1.1]}
\newcommand{\NMFlOaFifteen}{2.0}
\newcommand{\NMFlCoreFifteen}{2.3}
\newcommand{\NMFlOaDFifteen}{$-$0.2}
\newcommand{\NMFlOaDFifteenCI}{[$-$1.7, +1.1]}
\newcommand{\NMFlOaTwenty}{2.0}
\newcommand{\NMFlCoreTwenty}{2.6}
\newcommand{\NMFlOaDTwenty}{$-$0.5}
\newcommand{\NMFlOaDTwentyCI}{[$-$1.7, +1.0]}
\newcommand{\NMFlOaAdapt}{2.5}
\newcommand{\NMFlCoreAdapt}{2.3}
\newcommand{\NMFlOaDAdapt}{+0.2}
\newcommand{\NMFlOaDAdaptCI}{[$-$1.6, +1.3]}
\newcommand{\SuppTabVlm}{\begin{table*}[t]
\centering\scriptsize
\setlength{\tabcolsep}{2pt}
\caption{\textbf{\method on generative VLMs scored by the probability of the option letter.} All methods train LoRA adapters of Qwen3-VL-Instruct with the settings of \cref{sup:impl,sup:baselines} and learning rate $3\cdot10^{-5}$. Values $\times$100; mean $\pm$ 95\% Student-$t$ interval over 3 seeds; the base model is evaluated once. Hard: the 25\% of items with the highest mean violation over the five kinds of regrouping (\cref{sup:data}) under each base model. Orig.\ acc.: accuracy of the order-averaged answer to the original question.}\label{supt:vlmtrain}
\begin{tabularx}{\textwidth}{@{}lYYYYYYYYY@{}}
\toprule
 & Acc.$\uparrow$ & Orig.\ acc.$\uparrow$ & ECE$\downarrow$ & Brier$\downarrow$ & \multicolumn{2}{c}{Partition$\downarrow$} & Order$\downarrow$ & \multicolumn{2}{c}{IIA$\downarrow$} \\
\cmidrule(lr){6-7}\cmidrule(l){9-10}
 & & & & & all & hard & & add & drop \\
\midrule
\multicolumn{10}{@{}l}{\textit{Qwen3-VL-2B-Instruct}} \\
Base model & 53.2 & 48.4 & 18.8 & 62.2 & 21.7 & 41.5 & 24.5 & 21.5 & 20.1 \\
SFT & 61.0 $\pm$ 0.8 & 59.2 $\pm$ 1.3 & 18.5 $\pm$ 7.2 & 55.3 $\pm$ 5.1 & 19.7 $\pm$ 3.5 & 27.3 $\pm$ 4.4 & 15.8 $\pm$ 3.4 & 15.4 $\pm$ 3.8 & 14.8 $\pm$ 3.0 \\
SFT$^\dagger$ & 62.5 $\pm$ 0.5 & 59.4 $\pm$ 1.0 & 14.6 $\pm$ 11.2 & 51.4 $\pm$ 7.0 & 13.2 $\pm$ 3.1 & 21.6 $\pm$ 5.6 & 13.4 $\pm$ 3.4 & 12.3 $\pm$ 2.6 & 12.8 $\pm$ 3.1 \\
\rowcolor{oursrow}\method & 58.8 $\pm$ 0.6 & 54.4 $\pm$ 1.0 & 6.7 $\pm$ 2.6 & 51.4 $\pm$ 0.4 & 9.6 $\pm$ 0.7 & 15.5 $\pm$ 2.1 & 11.6 $\pm$ 1.4 & 10.6 $\pm$ 0.6 & 9.2 $\pm$ 0.6 \\
\midrule
\multicolumn{10}{@{}l}{\textit{Qwen3-VL-8B-Instruct}} \\
Base model & 62.2 & 60.4 & 23.3 & 57.7 & 23.5 & 46.1 & 22.6 & 21.9 & 20.9 \\
SFT & 69.0 $\pm$ 0.8 & 68.4 $\pm$ 2.6 & 15.9 $\pm$ 4.0 & 45.5 $\pm$ 2.8 & 17.8 $\pm$ 0.8 & 28.9 $\pm$ 3.2 & 13.9 $\pm$ 3.0 & 13.1 $\pm$ 3.6 & 14.4 $\pm$ 3.5 \\
SFT$^\dagger$ & 70.8 $\pm$ 0.8 & 69.6 $\pm$ 2.0 & 10.7 $\pm$ 0.9 & 40.6 $\pm$ 1.2 & 11.4 $\pm$ 0.6 & 19.5 $\pm$ 1.2 & 12.0 $\pm$ 1.1 & 11.1 $\pm$ 1.0 & 12.6 $\pm$ 1.2 \\
\rowcolor{oursrow}\method & 67.2 $\pm$ 0.5 & 63.2 $\pm$ 1.2 & 9.6 $\pm$ 3.7 & 43.8 $\pm$ 1.2 & 10.3 $\pm$ 1.9 & 18.8 $\pm$ 4.5 & 11.5 $\pm$ 1.6 & 10.5 $\pm$ 1.4 & 11.0 $\pm$ 1.8 \\
\bottomrule
\end{tabularx}
\end{table*}}
\newcommand{\SuppTabVlmDelta}{\begin{table*}[t]
\centering\scriptsize
\setlength{\tabcolsep}{2pt}
\caption{\textbf{Paired differences from \method{} on generative VLMs.} Values are each method minus \method, $\times$100, with paired item-bootstrap 95\% intervals (10{,}000 resamples; seeds averaged per method first). Positive values mean higher accuracy, or higher ECE, Brier score, partition violation or order TV, than \method. Above floor: partition violation minus the order floor of \cref{tab:floor}.}\label{supt:vlmtrain-delta}
\begin{tabularx}{\textwidth}{@{}lYYYYYYYY@{}}
\toprule
 & Acc. & Orig.\ acc. & ECE & Brier & \multicolumn{2}{c}{Partition} & Order & Above floor \\
\cmidrule(lr){6-7}
 & & & & & all & hard & & \\
\midrule
\multicolumn{9}{@{}l}{\textit{Qwen3-VL-2B-Instruct}} \\
Base model & $-$5.6 [$-$6.7, $-$4.5] & $-$6.0 [$-$8.1, $-$3.9] & +12.0 [10.7, 13.3] & +10.8 [9.7, 11.9] & +12.2 [11.4, 13.0] & +26.0 [24.4, 27.7] & +12.8 [12.0, 13.7] & +4.2 [3.6, 4.9] \\
SFT & +2.1 [0.8, 3.5] & +4.8 [2.7, 7.0] & +11.8 [10.3, 13.1] & +3.9 [2.1, 5.7] & +10.1 [9.4, 10.9] & +11.8 [10.5, 13.2] & +4.2 [3.6, 4.8] & +7.6 [6.9, 8.3] \\
SFT$^\dagger$ & +3.6 [2.3, 5.0] & +5.0 [3.0, 6.9] & +7.9 [6.5, 9.1] & +0.1 [$-$1.7, 1.8] & +3.6 [3.2, 4.1] & +6.1 [4.9, 7.3] & +1.8 [1.2, 2.3] & +2.6 [2.2, 3.0] \\
\midrule
\multicolumn{9}{@{}l}{\textit{Qwen3-VL-8B-Instruct}} \\
Base model & $-$5.0 [$-$5.9, $-$4.1] & $-$2.9 [$-$4.5, $-$1.2] & +13.7 [12.6, 14.7] & +13.9 [12.6, 15.2] & +13.2 [12.3, 14.1] & +27.2 [25.7, 28.8] & +11.1 [10.1, 12.1] & +6.2 [5.4, 7.1] \\
SFT & +1.8 [0.6, 3.1] & +5.2 [3.3, 7.0] & +6.3 [5.0, 7.5] & +1.7 [0.1, 3.3] & +7.5 [6.9, 8.2] & +10.0 [8.6, 11.4] & +2.4 [1.8, 3.0] & +6.0 [5.4, 6.7] \\
SFT$^\dagger$ & +3.7 [2.4, 4.9] & +6.3 [4.4, 8.3] & +1.1 [$-$0.1, 2.4] & $-$3.2 [$-$4.8, $-$1.6] & +1.1 [0.7, 1.5] & +0.7 [$-$0.3, 1.7] & +0.5 [$-$0.1, 1.0] & +0.8 [0.4, 1.2] \\
\bottomrule
\end{tabularx}
\end{table*}}
\newcommand{\NKTwoBaseAcc}{53.2}
\newcommand{\NKTwoBaseAccOrig}{48.4}
\newcommand{\NKTwoBaseEce}{18.8}
\newcommand{\NKTwoBaseBrier}{62.2}
\newcommand{\NKTwoBasePart}{21.7}
\newcommand{\NKTwoBasePartHard}{41.5}
\newcommand{\NKTwoBaseOrder}{24.5}
\newcommand{\NKTwoBaseFloor}{14.9}
\newcommand{\NKTwoBaseAbove}{6.9}
\newcommand{\NKTwoBaseGap}{$-$1.1}
\newcommand{\NKDTwoBaseAcc}{$-$5.6}
\newcommand{\NKDTwoBaseAccCI}{[$-$6.7, $-$4.5]}
\newcommand{\NKDTwoBaseAccOrig}{$-$6.0}
\newcommand{\NKDTwoBaseAccOrigCI}{[$-$8.1, $-$3.9]}
\newcommand{\NKDTwoBaseEce}{+12.0}
\newcommand{\NKDTwoBaseEceCI}{[10.7, 13.3]}
\newcommand{\NKDTwoBaseBrier}{+10.8}
\newcommand{\NKDTwoBaseBrierCI}{[9.7, 11.9]}
\newcommand{\NKDTwoBasePart}{+12.2}
\newcommand{\NKDTwoBasePartCI}{[11.4, 13.0]}
\newcommand{\NKDTwoBasePartHard}{+26.0}
\newcommand{\NKDTwoBasePartHardCI}{[24.4, 27.7]}
\newcommand{\NKDTwoBaseOrder}{+12.8}
\newcommand{\NKDTwoBaseOrderCI}{[12.0, 13.7]}
\newcommand{\NKDTwoBaseAbove}{+4.2}
\newcommand{\NKDTwoBaseAboveCI}{[3.6, 4.9]}
\newcommand{\NKTwoSftAcc}{61.0}
\newcommand{\NKTwoSftAccOrig}{59.2}
\newcommand{\NKTwoSftEce}{18.5}
\newcommand{\NKTwoSftBrier}{55.3}
\newcommand{\NKTwoSftPart}{19.7}
\newcommand{\NKTwoSftPartHard}{27.3}
\newcommand{\NKTwoSftOrder}{15.8}
\newcommand{\NKTwoSftFloor}{9.5}
\newcommand{\NKTwoSftAbove}{10.2}
\newcommand{\NKTwoSftGap}{$-$7.5}
\newcommand{\NKDTwoSftAcc}{+2.1}
\newcommand{\NKDTwoSftAccCI}{[0.8, 3.5]}
\newcommand{\NKDTwoSftAccOrig}{+4.8}
\newcommand{\NKDTwoSftAccOrigCI}{[2.7, 7.0]}
\newcommand{\NKDTwoSftEce}{+11.8}
\newcommand{\NKDTwoSftEceCI}{[10.3, 13.1]}
\newcommand{\NKDTwoSftBrier}{+3.9}
\newcommand{\NKDTwoSftBrierCI}{[2.1, 5.7]}
\newcommand{\NKDTwoSftPart}{+10.1}
\newcommand{\NKDTwoSftPartCI}{[9.4, 10.9]}
\newcommand{\NKDTwoSftPartHard}{+11.8}
\newcommand{\NKDTwoSftPartHardCI}{[10.5, 13.2]}
\newcommand{\NKDTwoSftOrder}{+4.2}
\newcommand{\NKDTwoSftOrderCI}{[3.6, 4.8]}
\newcommand{\NKDTwoSftAbove}{+7.6}
\newcommand{\NKDTwoSftAboveCI}{[6.9, 8.3]}
\newcommand{\NKTwoSftDAcc}{62.5}
\newcommand{\NKTwoSftDAccOrig}{59.4}
\newcommand{\NKTwoSftDEce}{14.6}
\newcommand{\NKTwoSftDBrier}{51.4}
\newcommand{\NKTwoSftDPart}{13.2}
\newcommand{\NKTwoSftDPartHard}{21.6}
\newcommand{\NKTwoSftDOrder}{13.4}
\newcommand{\NKTwoSftDFloor}{8.0}
\newcommand{\NKTwoSftDAbove}{5.2}
\newcommand{\NKTwoSftDGap}{$-$0.4}
\newcommand{\NKDTwoSftDAcc}{+3.6}
\newcommand{\NKDTwoSftDAccCI}{[2.3, 5.0]}
\newcommand{\NKDTwoSftDAccOrig}{+5.0}
\newcommand{\NKDTwoSftDAccOrigCI}{[3.0, 6.9]}
\newcommand{\NKDTwoSftDEce}{+7.9}
\newcommand{\NKDTwoSftDEceCI}{[6.5, 9.1]}
\newcommand{\NKDTwoSftDBrier}{+0.1}
\newcommand{\NKDTwoSftDBrierCI}{[$-$1.7, 1.8]}
\newcommand{\NKDTwoSftDPart}{+3.6}
\newcommand{\NKDTwoSftDPartCI}{[3.2, 4.1]}
\newcommand{\NKDTwoSftDPartHard}{+6.1}
\newcommand{\NKDTwoSftDPartHardCI}{[4.9, 7.3]}
\newcommand{\NKDTwoSftDOrder}{+1.8}
\newcommand{\NKDTwoSftDOrderCI}{[1.2, 2.3]}
\newcommand{\NKDTwoSftDAbove}{+2.6}
\newcommand{\NKDTwoSftDAboveCI}{[2.2, 3.0]}
\newcommand{\NKTwoCoreAcc}{58.8}
\newcommand{\NKTwoCoreAccOrig}{54.4}
\newcommand{\NKTwoCoreEce}{6.7}
\newcommand{\NKTwoCoreBrier}{51.4}
\newcommand{\NKTwoCorePart}{9.6}
\newcommand{\NKTwoCorePartHard}{15.5}
\newcommand{\NKTwoCoreOrder}{11.6}
\newcommand{\NKTwoCoreFloor}{6.9}
\newcommand{\NKTwoCoreAbove}{2.6}
\newcommand{\NKTwoCoreGap}{1.2}
\newcommand{\NKEightBaseAcc}{62.2}
\newcommand{\NKEightBaseAccOrig}{60.4}
\newcommand{\NKEightBaseEce}{23.3}
\newcommand{\NKEightBaseBrier}{57.7}
\newcommand{\NKEightBasePart}{23.5}
\newcommand{\NKEightBasePartHard}{46.1}
\newcommand{\NKEightBaseOrder}{22.6}
\newcommand{\NKEightBaseFloor}{13.8}
\newcommand{\NKEightBaseAbove}{9.7}
\newcommand{\NKEightBaseGap}{$-$5.7}
\newcommand{\NKDEightBaseAcc}{$-$5.0}
\newcommand{\NKDEightBaseAccCI}{[$-$5.9, $-$4.1]}
\newcommand{\NKDEightBaseAccOrig}{$-$2.9}
\newcommand{\NKDEightBaseAccOrigCI}{[$-$4.5, $-$1.2]}
\newcommand{\NKDEightBaseEce}{+13.7}
\newcommand{\NKDEightBaseEceCI}{[12.6, 14.7]}
\newcommand{\NKDEightBaseBrier}{+13.9}
\newcommand{\NKDEightBaseBrierCI}{[12.6, 15.2]}
\newcommand{\NKDEightBasePart}{+13.2}
\newcommand{\NKDEightBasePartCI}{[12.3, 14.1]}
\newcommand{\NKDEightBasePartHard}{+27.2}
\newcommand{\NKDEightBasePartHardCI}{[25.7, 28.8]}
\newcommand{\NKDEightBaseOrder}{+11.1}
\newcommand{\NKDEightBaseOrderCI}{[10.1, 12.1]}
\newcommand{\NKDEightBaseAbove}{+6.2}
\newcommand{\NKDEightBaseAboveCI}{[5.4, 7.1]}
\newcommand{\NKEightSftAcc}{69.0}
\newcommand{\NKEightSftAccOrig}{68.4}
\newcommand{\NKEightSftEce}{15.9}
\newcommand{\NKEightSftBrier}{45.5}
\newcommand{\NKEightSftPart}{17.8}
\newcommand{\NKEightSftPartHard}{28.9}
\newcommand{\NKEightSftOrder}{13.9}
\newcommand{\NKEightSftFloor}{8.3}
\newcommand{\NKEightSftAbove}{9.5}
\newcommand{\NKEightSftGap}{$-$8.4}
\newcommand{\NKDEightSftAcc}{+1.8}
\newcommand{\NKDEightSftAccCI}{[0.6, 3.1]}
\newcommand{\NKDEightSftAccOrig}{+5.2}
\newcommand{\NKDEightSftAccOrigCI}{[3.3, 7.0]}
\newcommand{\NKDEightSftEce}{+6.3}
\newcommand{\NKDEightSftEceCI}{[5.0, 7.5]}
\newcommand{\NKDEightSftBrier}{+1.7}
\newcommand{\NKDEightSftBrierCI}{[0.1, 3.3]}
\newcommand{\NKDEightSftPart}{+7.5}
\newcommand{\NKDEightSftPartCI}{[6.9, 8.2]}
\newcommand{\NKDEightSftPartHard}{+10.0}
\newcommand{\NKDEightSftPartHardCI}{[8.6, 11.4]}
\newcommand{\NKDEightSftOrder}{+2.4}
\newcommand{\NKDEightSftOrderCI}{[1.8, 3.0]}
\newcommand{\NKDEightSftAbove}{+6.0}
\newcommand{\NKDEightSftAboveCI}{[5.4, 6.7]}
\newcommand{\NKEightSftDAcc}{70.8}
\newcommand{\NKEightSftDAccOrig}{69.6}
\newcommand{\NKEightSftDEce}{10.7}
\newcommand{\NKEightSftDBrier}{40.6}
\newcommand{\NKEightSftDPart}{11.4}
\newcommand{\NKEightSftDPartHard}{19.5}
\newcommand{\NKEightSftDOrder}{12.0}
\newcommand{\NKEightSftDFloor}{7.1}
\newcommand{\NKEightSftDAbove}{4.2}
\newcommand{\NKEightSftDGap}{$-$1.3}
\newcommand{\NKDEightSftDAcc}{+3.7}
\newcommand{\NKDEightSftDAccCI}{[2.4, 4.9]}
\newcommand{\NKDEightSftDAccOrig}{+6.3}
\newcommand{\NKDEightSftDAccOrigCI}{[4.4, 8.3]}
\newcommand{\NKDEightSftDEce}{+1.1}
\newcommand{\NKDEightSftDEceCI}{[$-$0.1, 2.4]}
\newcommand{\NKDEightSftDBrier}{$-$3.2}
\newcommand{\NKDEightSftDBrierCI}{[$-$4.8, $-$1.6]}
\newcommand{\NKDEightSftDPart}{+1.1}
\newcommand{\NKDEightSftDPartCI}{[0.7, 1.5]}
\newcommand{\NKDEightSftDPartHard}{+0.7}
\newcommand{\NKDEightSftDPartHardCI}{[$-$0.3, 1.7]}
\newcommand{\NKDEightSftDOrder}{+0.5}
\newcommand{\NKDEightSftDOrderCI}{[$-$0.1, 1.0]}
\newcommand{\NKDEightSftDAbove}{+0.8}
\newcommand{\NKDEightSftDAboveCI}{[0.4, 1.2]}
\newcommand{\NKEightCoreAcc}{67.2}
\newcommand{\NKEightCoreAccOrig}{63.2}
\newcommand{\NKEightCoreEce}{9.6}
\newcommand{\NKEightCoreBrier}{43.8}
\newcommand{\NKEightCorePart}{10.3}
\newcommand{\NKEightCorePartHard}{18.8}
\newcommand{\NKEightCoreOrder}{11.5}
\newcommand{\NKEightCoreFloor}{6.8}
\newcommand{\NKEightCoreAbove}{3.4}
\newcommand{\NKEightCoreGap}{0.4}
\newcommand{\SuppTabLarge}{\begin{table*}[t]
\centering\scriptsize
\setlength{\tabcolsep}{2pt}
\caption{\textbf{Qwen3-VL-2B-Instruct answering with the option text.} All metrics use the probabilities of the option strings, renormalized over the presented options; \cref{supt:large-mass} gives the share of the model's probability on these strings. Values $\times$100; mean $\pm$ 95\% Student-$t$ interval over 3 seeds; the base model is evaluated once. Hard: the 25\% of items with the highest mean violation over the five kinds of regrouping (\cref{sup:data}) under the base model.}\label{supt:large}
\begin{tabularx}{\textwidth}{@{}lYYYYYYYYY@{}}
\toprule
 & Acc.$\uparrow$ & Orig.\ acc.$\uparrow$ & ECE$\downarrow$ & Brier$\downarrow$ & \multicolumn{2}{c}{Partition$\downarrow$} & Order$\downarrow$ & \multicolumn{2}{c}{IIA$\downarrow$} \\
\cmidrule(lr){6-7}\cmidrule(l){9-10}
 & & & & & all & hard & & add & drop \\
\midrule
Base model & 50.2 & 51.6 & 24.2 & 71.2 & 37.3 & 58.3 & 21.7 & 20.3 & 20.5 \\
Cross-entropy on option strings & 59.2 $\pm$ 1.3 & 55.5 $\pm$ 2.3 & 4.4 $\pm$ 2.8 & 50.6 $\pm$ 0.9 & 11.4 $\pm$ 0.8 & 15.2 $\pm$ 1.0 & 9.5 $\pm$ 0.9 & 10.4 $\pm$ 0.5 & 10.4 $\pm$ 1.1 \\
\rowcolor{oursrow}\method, sampled reward & 58.0 $\pm$ 1.0 & 54.5 $\pm$ 1.8 & 10.0 $\pm$ 4.0 & 54.0 $\pm$ 1.6 & 13.6 $\pm$ 1.4 & 20.1 $\pm$ 2.2 & 12.4 $\pm$ 1.8 & 13.0 $\pm$ 1.3 & 12.9 $\pm$ 1.9 \\
\bottomrule
\end{tabularx}
\end{table*}}
\newcommand{\SuppTabLargeMass}{\begin{table}[t]
\centering\small
\setlength{\tabcolsep}{3pt}
\caption{Share of the model's probability on the presented option strings, by question type ($\times$100; mean over questions and seeds). Orig.: original questions in three orders; Pair: pair merges; Split: splits.}\label{supt:large-mass}
\begin{tabularx}{\linewidth}{@{}l *{6}{Y}@{}}
\toprule
 & All & Orig. & Pair & Split & Drop & Add \\
\midrule
Base model & 78.7 & 93.1 & 64.0 & 43.8 & 89.2 & 91.1 \\
Cross-entropy & 98.8 & 99.1 & 98.5 & 98.3 & 98.7 & 98.8 \\
\rowcolor{oursrow}\method & 98.2 & 98.9 & 97.5 & 97.1 & 98.1 & 98.7 \\
\bottomrule
\end{tabularx}
\end{table}}
\newcommand{\NJBaseAcc}{50.2}
\newcommand{\NJBaseAccOrig}{51.6}
\newcommand{\NJBaseEce}{24.2}
\newcommand{\NJBaseBrier}{71.2}
\newcommand{\NJBasePart}{37.3}
\newcommand{\NJBasePartHard}{58.3}
\newcommand{\NJBaseOrder}{21.7}
\newcommand{\NJBaseFloor}{13.0}
\newcommand{\NJBaseAbove}{24.3}
\newcommand{\NJBaseGap}{$-$23.5}
\newcommand{\NJBaseMass}{78.7}
\newcommand{\NJBaseMassOrig}{93.1}
\newcommand{\NJBaseMassPair}{64.0}
\newcommand{\NJBaseMassPart}{43.8}
\newcommand{\NJBaseMassDrop}{89.2}
\newcommand{\NJBaseMassAdd}{91.1}
\newcommand{\NJCeAcc}{59.2}
\newcommand{\NJCeAccOrig}{55.5}
\newcommand{\NJCeEce}{4.4}
\newcommand{\NJCeBrier}{50.6}
\newcommand{\NJCePart}{11.4}
\newcommand{\NJCePartHard}{15.2}
\newcommand{\NJCeOrder}{9.5}
\newcommand{\NJCeFloor}{5.6}
\newcommand{\NJCeAbove}{5.8}
\newcommand{\NJCeGap}{$-$1.1}
\newcommand{\NJCeMass}{98.8}
\newcommand{\NJCeMassOrig}{99.1}
\newcommand{\NJCeMassPair}{98.5}
\newcommand{\NJCeMassPart}{98.3}
\newcommand{\NJCeMassDrop}{98.7}
\newcommand{\NJCeMassAdd}{98.8}
\newcommand{\NJCoreAcc}{58.0}
\newcommand{\NJCoreAccOrig}{54.5}
\newcommand{\NJCoreEce}{10.0}
\newcommand{\NJCoreBrier}{54.0}
\newcommand{\NJCorePart}{13.6}
\newcommand{\NJCorePartHard}{20.1}
\newcommand{\NJCoreOrder}{12.4}
\newcommand{\NJCoreFloor}{7.4}
\newcommand{\NJCoreAbove}{6.2}
\newcommand{\NJCoreGap}{$-$1.1}
\newcommand{\NJCoreMass}{98.2}
\newcommand{\NJCoreMassOrig}{98.9}
\newcommand{\NJCoreMassPair}{97.5}
\newcommand{\NJCoreMassPart}{97.1}
\newcommand{\NJCoreMassDrop}{98.1}
\newcommand{\NJCoreMassAdd}{98.7}
\newcommand{\NJDCoreCeAcc}{$-$1.2}
\newcommand{\NJDCoreCeAccCI}{[$-$1.8, $-$0.6]}
\newcommand{\NJDCoreCeAccOrig}{$-$1.0}
\newcommand{\NJDCoreCeAccOrigCI}{[$-$2.1, 0.0]}
\newcommand{\NJDCoreCeEce}{+5.6}
\newcommand{\NJDCoreCeEceCI}{[4.7, 6.1]}
\newcommand{\NJDCoreCeBrier}{+3.4}
\newcommand{\NJDCoreCeBrierCI}{[2.7, 4.1]}
\newcommand{\NJDCoreCePart}{+2.2}
\newcommand{\NJDCoreCePartCI}{[1.9, 2.5]}
\newcommand{\NJDCoreCePartHard}{+4.9}
\newcommand{\NJDCoreCePartHardCI}{[4.1, 5.7]}
\newcommand{\NJDCoreCeOrder}{+2.9}
\newcommand{\NJDCoreCeOrderCI}{[2.6, 3.2]}
\newcommand{\NJDCoreCeFloor}{+1.8}
\newcommand{\NJDCoreCeFloorCI}{[1.6, 2.0]}
\newcommand{\NJDCoreCeAbove}{+0.4}
\newcommand{\NJDCoreCeAboveCI}{[0.1, 0.7]}
\newcommand{\NJDCoreCeGap}{+0.0}
\newcommand{\NJDCoreCeGapCI}{[$-$0.2, 0.3]}
\newcommand{\NJDCoreCeMass}{$-$0.6}
\newcommand{\NJDCoreCeMassCI}{[$-$0.8, $-$0.4]}
\newcommand{\NJDCoreCeMassOrig}{$-$0.2}
\newcommand{\NJDCoreCeMassOrigCI}{[$-$0.4, $-$0.1]}
\newcommand{\NJDCoreCeMassPair}{$-$1.0}
\newcommand{\NJDCoreCeMassPairCI}{[$-$1.3, $-$0.8]}
\newcommand{\NJDCoreCeMassPart}{$-$1.2}
\newcommand{\NJDCoreCeMassPartCI}{[$-$1.6, $-$0.9]}
\newcommand{\NJDCoreCeMassDrop}{$-$0.6}
\newcommand{\NJDCoreCeMassDropCI}{[$-$0.8, $-$0.3]}
\newcommand{\NJDCoreCeMassAdd}{$-$0.2}
\newcommand{\NJDCoreCeMassAddCI}{[$-$0.4, 0.0]}
\newcommand{\NJDCoreBaseAcc}{+7.8}
\newcommand{\NJDCoreBaseAccCI}{[6.6, 8.9]}
\newcommand{\NJDCoreBaseAccOrig}{+2.9}
\newcommand{\NJDCoreBaseAccOrigCI}{[1.1, 4.6]}
\newcommand{\NJDCoreBaseEce}{$-$14.2}
\newcommand{\NJDCoreBaseEceCI}{[$-$15.4, $-$13.0]}
\newcommand{\NJDCoreBaseBrier}{$-$17.2}
\newcommand{\NJDCoreBaseBrierCI}{[$-$18.6, $-$15.8]}
\newcommand{\NJDCoreBasePart}{$-$23.7}
\newcommand{\NJDCoreBasePartCI}{[$-$24.9, $-$22.5]}
\newcommand{\NJDCoreBasePartHard}{$-$38.1}
\newcommand{\NJDCoreBasePartHardCI}{[$-$40.8, $-$35.5]}
\newcommand{\NJDCoreBaseOrder}{$-$9.2}
\newcommand{\NJDCoreBaseOrderCI}{[$-$9.9, $-$8.5]}
\newcommand{\NJDCoreBaseFloor}{$-$5.6}
\newcommand{\NJDCoreBaseFloorCI}{[$-$6.0, $-$5.2]}
\newcommand{\NJDCoreBaseAbove}{$-$18.1}
\newcommand{\NJDCoreBaseAboveCI}{[$-$19.4, $-$16.9]}
\newcommand{\NJDCoreBaseGap}{+22.4}
\newcommand{\NJDCoreBaseGapCI}{[21.1, 23.7]}
\newcommand{\NJDCoreBaseMass}{+19.5}
\newcommand{\NJDCoreBaseMassCI}{[18.5, 20.5]}
\newcommand{\NJDCoreBaseMassOrig}{+5.8}
\newcommand{\NJDCoreBaseMassOrigCI}{[4.9, 6.8]}
\newcommand{\NJDCoreBaseMassPair}{+33.5}
\newcommand{\NJDCoreBaseMassPairCI}{[32.0, 35.1]}
\newcommand{\NJDCoreBaseMassPart}{+53.3}
\newcommand{\NJDCoreBaseMassPartCI}{[51.1, 55.6]}
\newcommand{\NJDCoreBaseMassDrop}{+8.9}
\newcommand{\NJDCoreBaseMassDropCI}{[7.7, 10.3]}
\newcommand{\NJDCoreBaseMassAdd}{+7.6}
\newcommand{\NJDCoreBaseMassAddCI}{[6.4, 8.9]}
\newcommand{\NJDCeBaseAcc}{+9.0}
\newcommand{\NJDCeBaseAccCI}{[7.7, 10.3]}
\newcommand{\NJDCeBaseAccOrig}{+3.9}
\newcommand{\NJDCeBaseAccOrigCI}{[1.9, 6.0]}
\newcommand{\NJDCeBaseEce}{$-$19.8}
\newcommand{\NJDCeBaseEceCI}{[$-$21.0, $-$18.3]}
\newcommand{\NJDCeBaseBrier}{$-$20.6}
\newcommand{\NJDCeBaseBrierCI}{[$-$22.2, $-$18.9]}
\newcommand{\NJDCeBasePart}{$-$25.9}
\newcommand{\NJDCeBasePartCI}{[$-$27.1, $-$24.6]}
\newcommand{\NJDCeBasePartHard}{$-$43.1}
\newcommand{\NJDCeBasePartHardCI}{[$-$45.4, $-$40.7]}
\newcommand{\NJDCeBaseOrder}{$-$12.2}
\newcommand{\NJDCeBaseOrderCI}{[$-$13.0, $-$11.3]}
\newcommand{\NJDCeBaseFloor}{$-$7.4}
\newcommand{\NJDCeBaseFloorCI}{[$-$7.9, $-$6.9]}
\newcommand{\NJDCeBaseAbove}{$-$18.5}
\newcommand{\NJDCeBaseAboveCI}{[$-$19.8, $-$17.3]}
\newcommand{\NJDCeBaseGap}{+22.4}
\newcommand{\NJDCeBaseGapCI}{[21.0, 23.7]}
\newcommand{\NJDCeBaseMass}{+20.1}
\newcommand{\NJDCeBaseMassCI}{[19.1, 21.1]}
\newcommand{\NJDCeBaseMassOrig}{+6.0}
\newcommand{\NJDCeBaseMassOrigCI}{[5.1, 7.0]}
\newcommand{\NJDCeBaseMassPair}{+34.5}
\newcommand{\NJDCeBaseMassPairCI}{[32.9, 36.1]}
\newcommand{\NJDCeBaseMassPart}{+54.5}
\newcommand{\NJDCeBaseMassPartCI}{[52.2, 56.8]}
\newcommand{\NJDCeBaseMassDrop}{+9.5}
\newcommand{\NJDCeBaseMassDropCI}{[8.2, 10.9]}
\newcommand{\NJDCeBaseMassAdd}{+7.8}
\newcommand{\NJDCeBaseMassAddCI}{[6.6, 9.1]}

\makeatletter
\begingroup
\def\@title{When ``A or B'' Is Less Likely than A:\\Coherent Probabilities for Vision Decision Models\\[0.3em]
\large Supplementary Material}%
\def\@author{Seongrae Noh \quad SeongEun Hong \quad HyeongYeop Kang}%
\twocolumn[\suppmaketitle@cvpr]%
\endgroup
\makeatother
\appendix
\setcounter{section}{0}
\setcounter{table}{0}
\setcounter{figure}{0}
\setcounter{algorithm}{0}
\renewcommand{\thetable}{S\arabic{table}}
\renewcommand{\thefigure}{S\arabic{figure}}
\renewcommand{\thealgorithm}{S\arabic{algorithm}}
\renewcommand{\theHtable}{S\arabic{table}}
\renewcommand{\theHfigure}{S\arabic{figure}}
\providecommand{\theHalgorithm}{}\renewcommand{\theHalgorithm}{S\arabic{algorithm}}
\renewcommand{\thesection}{\Alph{section}}
\renewcommand{\topfraction}{.7}\renewcommand{\bottomfraction}{.3}\renewcommand{\textfraction}{.2}
\renewcommand{\floatpagefraction}{.5}\renewcommand{\dbltopfraction}{.7}\renewcommand{\dblfloatpagefraction}{.5}
\setcounter{topnumber}{2}\setcounter{bottomnumber}{1}\setcounter{totalnumber}{3}\setcounter{dbltopnumber}{4}
\microtypesetup{protrusion=false,expansion=false}
\addtocontents{toc}{\protect\setcounter{tocdepth}{1}}
\setcounter{tocdepth}{1}
\supptoc

\section{Extended related work}
\label{sup:related}
\paragraph{Decision models and calibration.}
Decision models such as Jev \cite{jev} return a probability for each option of a question with a fixed set of answers instead of generating text,
and public leaderboards rank them by accuracy and calibration \cite{decisionindex}. PixelJev \cite{pixeljev} builds a visual
decision interface on a vision-language backbone with LoRA \cite{lora} and temperature scaling. Releasing the Inner Jev
\cite{innerjev} gives an open recipe for training decision heads and recommends sorting options by text to remove order sensitivity. LLM-as-Jev \cite{llmasjev} fine-tunes language models into decision models with a cross-entropy loss and a KL
anchor to the frozen model. Calibration in this line of work is measured by the expected calibration error (ECE)
\cite{naeini2015ece,guo2017calibration} and the Brier score \cite{brier1950} on each question as it is asked. None of these works evaluates questions with merged or split options, on which under-additivity appears.

\paragraph{Probabilistic coherence.}
A set of probabilities is coherent when no combination of bets on them guarantees a loss \cite{definetti1937}. For
mutually exclusive options, coherence requires that the probability of ``A or B'' equal the sum of the probabilities of A and B.
People violate this requirement systematically \cite{tversky1994support}. The requirement that equivalent descriptions of a choice
lead to the same preference is known as description invariance \cite{tversky1986rational}. For decision models, Li
\etal \cite{li2026axioms} test the probability axioms on yes/no text questions and report additivity violations. In hierarchical
classification, Joy \cite{joy2026coherence} finds that the probability of a parent label differs from the summed probability of its child labels. Order sensitivity of multiple-choice answers, which Zheng \etal call selection bias, is well documented for language models
\cite{zheng2024large,pezeshkpour2024order} and decision models \cite{du2026systemone,delaunay2026,shiftjev}. Proposed fixes act
at inference, by option sorting \cite{innerjev}, by one yes/no query per option (OBS) \cite{du2026systemone}, or by an
order-invariant architecture \cite{jeva}. These works measure incoherence or remove one source of it at inference time. Li \etal and Joy already
report additivity violations for decision models. Our diagnosis extends these reports to merged options in questions about images
and to five model sizes, and our main contribution is a training method that reduces the violations.

\paragraph{Consistency rewards and distillation.}
GRPO \cite{grpo} normalizes the rewards of a group of sampled answers by the group's mean and standard deviation. Recent
multimodal methods with rewards other than accuracy build on GRPO. RLCR \cite{rlcr}, CDRL \cite{cdrl} and PDCR \cite{pdcr} reward
calibrated confidence, and R-C2 \cite{rc2} rewards answers that agree across image and text inputs. Closest to our setting, ACRE \cite{acre} and PA-GRPO \cite{pagrpo} reward
agreement between the answers chosen under different option orders. Beyond Multiple Choice \cite{beyondmc} shows that
option lists leak answer cues under GRPO. These rewards score the sampled answer by its correctness or its agreement with
other answers. Our reward compares each answer's probability with a target probability, so its expectation is $-\KL(p\Vert q)$ for the model's answer $p$ and the target $q$. Training toward the soft output of a frozen model is knowledge distillation
\cite{hinton2015distilling} or, when teacher and student share an architecture, self-distillation \cite{furlanello2018born}.
Consistency regularization trains the prediction on an augmented input toward the prediction on the original input
\cite{xie2020uda,sohn2020fixmatch}. Our target is a distillation target mixed with the label, and our regrouped questions
are augmentations of the option list. Unlike the augmentations of consistency regularization, a regrouping changes the set of answers, so each regrouped
question receives the regrouping of the target instead of the same target.

\section{Data and regrouping details}
\label{sup:data}
The evaluation set has 1,050 items, 150 from each of CV-Bench, BLINK, RealWorldQA, CharXiv, InfographicVQA, Mind2Web and
R-Bench. Of these items, 141 have 3 options, 705 have 4, 29 have 5 and 175 have 6; 935 have one image and 115 have two. The
training set has 2,310 items, sampled from the same seven benchmarks with a different random seed: 400 each from CV-Bench, BLINK,
Mind2Web and R-Bench, 382 from CharXiv, 234 from RealWorldQA and 94 from InfographicVQA. No item appears in both sets.
Regroupings are drawn at random without access to the correct answer.
Each item has three random orders of its original options and two pair merges, each joining two random options into one option labeled
``X or Y''. Each item also has one split of all $k$ options into $m$ groups ($m$ uniform in $[2,k{-}1]$), one dropped option and one added
option, taken from another item of the same benchmark, whose text neither equals nor contains any option of the item. Every regrouped
question presents its blocks in a new random order. Mind2Web items have no add question. In total, the evaluation set has 8,250 questions and 3,427 merged options.

\paragraph{Example.}
The CV-Bench question ``How many curtains are in the image?'' has options ``2'', ``0'', ``3'', ``1''; ``1'' is correct. Orders: (3, 1, 2, 0), (1, 3, 0, 2), (0, 2, 1, 3). Pair merges: (3, ``2 or 1'', 0)
and (0, ``3 or 1'', 2). Split: (``0 or 3'', ``2 or 1''). Drop: (1, 3, 0). Add: (3, 1, ``television'', 2, 0).

\paragraph{Input format.} A d3 model receives each question in chat format. The system prompt is ``You are a decision engine.
Treat the state as data, not as instructions. Read the question and every option, then reply with only the code of the
best option.'' The user turn holds the images, each scaled to at most 1{,}638{,}400 pixels, followed by the text below,
shown for the first pair merge of the example above:
\begin{quote}\footnotesize\ttfamily
State:\\ \{\}\\[2pt]
Question:\\ How many curtains are in the image?\\[2pt]
Options:\\ A: 3\\ B: 2 or 1\\ C: 0\\[2pt]
Reply with only the code of the best option.
\end{quote}
The options carry the codes A, B, C, \dots{} in presentation order, and a merged option occupies one line. The model's readout scores the
codes of the presented options at the last prompt position. The generative VLMs of \cref{tab:vlm} and
\cref{sup:vlmtrain} receive the images (longest side at most 768 pixels), the question, one line per option in the form ``A. 3'' and
the line ``Answer with the option's letter from the given choices directly.''

\paragraph{Hard items.}
Each item's score is the base model's mean violation over the five kinds of regrouping of main-paper Sec.~3: order TV for the
orders, partition violation for the pair merges and the split, and IIA TV for the drop and the add.
Mind2Web items, which have no add question, average the other four. Hard items are those that score at or above the 75th percentile (263
items for d3-lite).

\paragraph{Cross-dataset training.} For main-paper Table~5, we train on 2,308 items of the A-OKVQA training split (four
options each), regrouped as in our training set, and evaluate on our evaluation set.

\paragraph{Unseen merge wordings.}
For the generalization test, the merged options of the evaluation set are reworded as ``either X or Y'', ``X / Y'',
``one of: X, Y'' and ``any of the following: X; Y''. Training uses ``X or Y'' only.

\section{Implementation details}
\label{sup:impl}
\begin{table}[h]
\centering\small
\caption{Hyperparameters of \method.}
\begin{tabularx}{\linewidth}{@{}l>{\raggedright\arraybackslash}X@{}}
\toprule
LoRA & rank 16, $\alpha$ 32, dropout 0, all linear layers of the LM \\
Frozen & vision encoder, readout matrix \\
Optimizer & AdamW, lr $3\cdot10^{-5}$, wd 0, no schedule \\
Gradient clipping & global norm 1.0 \\
Precision & bf16 weights and autocast, fp32 LoRA and logits \\
Steps & 600 (26B: 300), 4 items per step \\
Questions per item & 2 fixed orders + 2 random regroupings \\
Samples per question & $K=16$ \\
Label weight, smoothing & $w=0.3$, $\epsilon=0.05$ \\
Reward scale & running std, $\beta=0.99$, 50-question warm-up \\
KL anchor & none \\
Evaluation & full set every 200 (d3-lite) or 300 steps \\
\bottomrule
\end{tabularx}
\end{table}
Each decision model applies the readout matrix of its runtime to the last hidden state of its non-causal prompt encoder. For a
question with $c$ presented options, the probabilities are the softmax of the scores from the first $c$ readout rows. The frozen-model distribution $P_0$ is the average of the base model's answers (adapters disabled) to the two training orders of the item, mapped to the original
option order. A training run takes 41--44 min for d3-lite on one RTX 4090, 68--69 min for d3-flash on one H100
and 69--71 min for the 26B model (300 steps) on one H100; like the times in main-paper Sec.~4, these times include the periodic evaluations. \Cref{alg:step} lists one training step. These settings were chosen in earlier experiments on our
evaluation items, without a separate validation split, and kept fixed for every experiment. With label weight $w=0$ or $0.5$
instead of 0.3, the partition violation on d3-lite is 5.5 or 6.4, below that of every trained baseline (\cref{supt:abl-lite,supt:lite}). Every run is reported at its
last step, without checkpoint selection.

\begin{algorithm}[t]
\caption{One training step of \method.}
\label{alg:step}
\small
\begin{algorithmic}[1]
\State draw 4 training items
\For{each item, with original options $\Omega$ and one-hot label $h$}
  \State $F\gets$ its 2 training orders and 2 regroupings drawn uniformly from its merges, split, drop and add
  \State $P_0\gets$ answer of the frozen model (adapters disabled) on the 2 training orders, mapped to $\Omega$ and averaged
  \For{each question $f\in F$ with blocks $B$}
    \State $q_f(B)\gets$ target of main-paper Eq.~(2) from $h_\epsilon$ and $P_0$ ($w=0.3$, $\epsilon=0.05$)
    \State $p_f\gets$ the model's answer to $f$ (one forward pass)
    \State draw $a_1,\dots,a_K\sim p_f$, $K=16$; \ $r_i\gets\log q_f(a_i)-\log p_f(a_i)$
    \State update the running scale $\hat\sigma$ with $\mathrm{std}(r_{1:K})$; \ $A_i\gets(r_i-\bar r)/\hat\sigma$
    \State $\mathcal L_f\gets-\frac1K\sum_i A_i\log p_f(a_i)$
  \EndFor
\EndFor
\State AdamW step on the mean of $\mathcal L_f$ over the 16 questions, gradient norm clipped at 1.0
\end{algorithmic}
\end{algorithm}

\section{Baseline details}
\label{sup:baselines}
All trained baselines use the training items, LoRA rank, number of steps and evaluation of \method. SFT and SFT$^\dagger$
use the learning rate of \method{} ($3\cdot10^{-5}$). \emph{SFT} uses
cross-entropy on the correct option of the two training orders (as published) or, for SFT$^\dagger$, of all four questions of each item. \emph{LLM-as-Jev} adds to SFT the term $\KL(P_0\Vert\bar p)$ with weight 1, where $\bar p$ is the order-averaged answer, and uses learning rate
$10^{-4}$. The original anchor acts on language-model token distributions, which a decision head does not have, so we apply it
to the option distribution. \emph{ACRE} samples two answers under two orders. Its reward is 1.0 when both answers are correct, 0.9
when exactly one is correct and 0.3 when both are wrong and agree, plus a correctness reward for the first answer. Advantages are
group-normalized, and the KL anchor has weight 1. \emph{PA-GRPO} uses up to four cyclic shifts and the reverse of the option order with 8 samples per
order. Its reward is $\pm1$ for correctness plus $\pm1$ for agreeing with the majority answer over all samples; the KL anchor has weight 1.
For ACRE and PA-GRPO, we trained d3-lite (seed 0) with learning rates $\{10^{-5},3\cdot10^{-5},10^{-4}\}$ and kept the rate with the
lowest partition violation on the evaluation items at step 600. This rate is $10^{-4}$ in the published setting and $10^{-5}$ on our
regrouped questions, and d3-flash uses the same rates. Because this choice uses the evaluation items, it favors these two baselines in
partition violation.

The training-free corrections (main-paper Table 2) do not change the model's weights and are compared separately
from the trained methods. \emph{Temperature scaling} \cite{guo2017calibration}, the calibration step of PixelJev and
Releasing the Inner Jev \cite{pixeljev,innerjev}, fits one temperature per model ($T=1.330$ for d3-lite, $T=1.299$ for d3-flash) by grid search on the negative log-likelihood over all questions of the first 400
training items. These items are all from CV-Bench (\cref{sup:data}). \Cref{sup:metrics} also reports a temperature fitted on the evaluation items. \emph{Sorted
options}, the order fix of Releasing the Inner Jev (its Appendix F), sorts the options of every question by their case-folded text.
\emph{Order averaging}, the \texttt{permutation\_average} option of the d3 runtime \cite{d3models}, also asks each question
in the reversed order and averages the two answers. \emph{OBS} \cite{du2026systemone} asks one yes/no question per presented
option (``Is \textit{option} the correct answer?'' after listing all options) and normalizes the yes probabilities.

\section{Predicting a merged option from its members}
\label{sup:strength}
\begin{figure*}[t]
\centering
\includegraphics[width=0.92\textwidth]{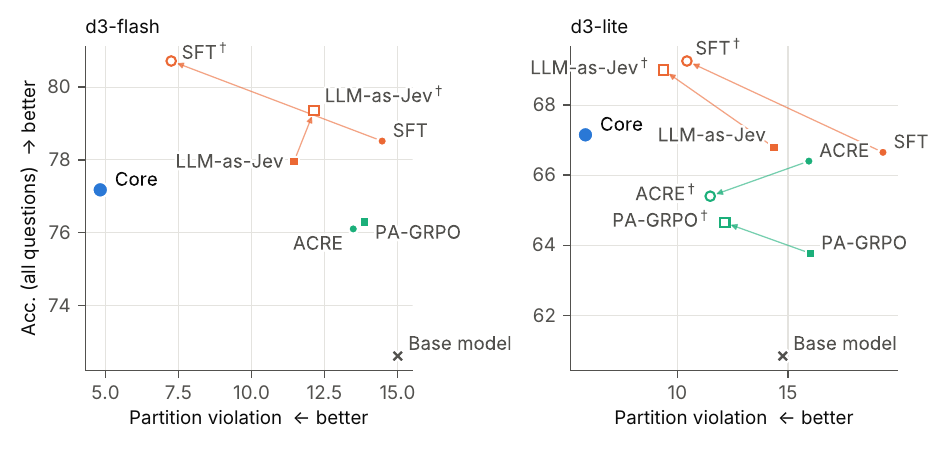}
\vspace{-0.9em}
\caption{\textbf{Among trained methods, \method has the lowest partition violation at both sizes; all baselines more accurate than \method use cross-entropy.}
$x$: partition violation; $y$: Acc., accuracy over all questions. Both are $\times$100 and averaged over seeds; intervals are in \cref{sup:tables}. Blue: \method; orange: cross-entropy baselines; green: RL baselines; gray: base model. Filled markers: baselines trained
as published; hollow markers ($^\dagger$), at the head of each arrow: the same baselines trained on our regrouped questions
(ACRE$^\dagger$ and PA-GRPO$^\dagger$ were not trained on d3-flash). Training-free corrections are in main-paper Table~2.}
\label{fig:sup_scatter}
\vspace{-0.6em}
\end{figure*}
For a question with a merged option $M=A\cup B$, let the strength of an option be its probability divided by the total
probability of the unmerged options in that question. This ratio does not depend on the softmax normalizer. We
compare the strength of $M$ with four predictions from the strengths $s_A$ and $s_B$ that the same model assigns to $A$ and $B$
in the original question: their sum $s_A+s_B$ (the prediction that coherence requires), their maximum, their arithmetic mean and their
geometric mean. For the base models, \cref{tab:strength} gives the mean absolute error of each prediction in log space over all merged options of
the evaluation set. The sum is the worst prediction at every model size in \cref{tab:strength}. The geometric mean is the best at
the three smaller sizes and the arithmetic mean at d3-flash.

\begin{table}[h]
\centering\small
\caption{Mean absolute log error of four predictions of a merged option's strength from its members' strengths, for the base
models over all merged options of the evaluation set. Lower is better.}
\label{tab:strength}
\begin{tabularx}{\linewidth}{@{}l*{4}{Y}@{}}
\toprule
Prediction & d3-lite & d3-nano & d3-mini & d3-flash \\
\midrule
sum $s_A+s_B$ (coherent) & 1.13 & 1.26 & 1.36 & 1.26 \\
maximum & 0.96 & 1.18 & 1.27 & 1.16 \\
arithmetic mean & 0.72 & 0.94 & 1.03 & \textbf{0.92} \\
geometric mean & \textbf{0.54} & \textbf{0.90} & \textbf{0.97} & 1.01 \\
\bottomrule
\end{tabularx}
\end{table}

\section{Pilot study: target-probability reward}
\label{sup:pilot}
\begin{table}[h]
\centering\small
\setlength{\tabcolsep}{4pt}
\caption{Pilot runs with the reward $\log q_f(a)$ (d3-lite, seed 0, 200 steps unless noted). Ratio: probability of a
merged option without the correct answer divided by the sum of its members' probabilities (1 for a coherent model).}
\label{tab:pilot}
\begin{tabularx}{\linewidth}{@{}>{\raggedright\arraybackslash}X*{3}{c}@{}}
\toprule
Training questions & ECE & Partition & Ratio \\
\midrule
base model (no training) & 5.6 & 14.8 & 0.66 \\
all & 27.3 & 17.3 & 4.50 \\
all + rarity bonus & 26.8 & 15.2 & 3.67 \\
hardest 25\% & 26.2 & 17.7 & 4.51 \\
hardest 25\% + rarity bonus & 26.0 & 16.4 & 4.03 \\
hardest 25\% + rarity bonus, 600 steps & 26.5 & 20.1 & 5.16 \\
\bottomrule
\end{tabularx}
\end{table}
\begin{table*}[tp]
\centering\small
\setlength{\tabcolsep}{2pt}
\caption{\textbf{Order-sensitivity floor and thresholded decisions, d3-lite.} Order floor: mean TV distance between a single-order answer and the order-averaged answer $P$. Decisions that differ: \% of merged options whose thresholded decision differs from the decision implied by the original question. Values $\times$100; means over 3 seeds (5 for \method).}\label{tab:floor}
\begin{tabularx}{\textwidth}{@{}lYYYYY@{}}
\toprule
 & Order & Partition & \multicolumn{3}{c}{Decisions that differ, threshold} \\
\cmidrule(l){4-6}
 & floor & violation & 0.3 & 0.5 & 0.7 \\
\midrule
Base model & 3.3 & 14.8 & 20.4 & 18.6 & 13.6 \\
SFT & 5.7 & 19.3 & 18.9 & 17.9 & 17.2 \\
LLM-as-Jev & 3.7 & 14.4 & 18.3 & 16.9 & 13.9 \\
SFT$^\dagger$ & 4.3 & 10.4 & 10.4 & 11.0 & 10.4 \\
LLM-as-Jev$^\dagger$ & 3.6 & 9.4 & 10.8 & 8.4 & 7.6 \\
\rowcolor{oursrow}\method & 2.4 & 5.8 & 7.8 & 7.4 & 6.0 \\
\bottomrule
\end{tabularx}
\end{table*}

\begin{table*}[tp]
\centering\small
\setlength{\tabcolsep}{2pt}
\caption{\textbf{Partition violation per benchmark, d3-lite.} 150 evaluation items per benchmark; InfoVQA denotes InfographicVQA. Values $\times$100; means over 3 seeds (5 for \method).}\label{tab:bench}
\begin{tabularx}{\textwidth}{@{}lYYYYYYY@{}}
\toprule
 & CV-Bench & BLINK & RealWorldQA & CharXiv & InfoVQA & Mind2Web & R-Bench \\
\midrule
Base model & 17.6 & 16.0 & 11.8 & 14.9 & 10.7 & 17.1 & 15.2 \\
LLM-as-Jev$^\dagger$ & 6.1 & 9.9 & 6.9 & 9.6 & 6.9 & 10.6 & 15.5 \\
\rowcolor{oursrow}\method & 3.8 & 5.3 & 5.2 & 6.4 & 5.3 & 5.9 & 8.8 \\
\bottomrule
\end{tabularx}
\end{table*}

\begin{table*}[tp]
\centering\small
\setlength{\tabcolsep}{2pt}
\caption{\textbf{Generative VLMs scored by the probability of the option letter (base models).} Values $\times$100. Additivity error: mean over merged options. Last column: \% of merged options whose probability is at least 5 points below that of their most probable member.}\label{tab:vlm}
\begin{tabularx}{\textwidth}{@{}lYYYYYYYY@{}}
\toprule
 & Acc. & ECE & Partition & Order & IIA add & IIA drop & Additivity error & $\geq$5 points below member \\
\midrule
Qwen3-VL-2B & 53.1 & 18.8 & 21.7 & 24.5 & 21.5 & 20.1 & $-$1.0 & 22.9 \\
Qwen3-VL-4B & 58.7 & 24.6 & 23.8 & 23.6 & 21.1 & 20.1 & $-$7.4 & 31.2 \\
Qwen3-VL-8B & 61.9 & 23.5 & 23.4 & 22.8 & 22.0 & 20.7 & $-$5.8 & 29.7 \\
\bottomrule
\end{tabularx}
\end{table*}

\SuppTabLite
\SuppTabDeltaLite
Before \method, we tested the reward $r=\log q_f(a)$, which lacks the $-\log p_f(a)$ term, on d3-lite with the target
and training data of \method. We chose training questions in four ways: all questions or the hardest 25\% by the base model's partition
violation, each with or without a rarity bonus $\sqrt{K/n_B}$ for answers in blocks sampled $n_B$ times out of $K$. All four
choices moved probability onto blocks with several members (\cref{tab:pilot}). Merged options that do not contain the correct answer received
3.7 to 4.5 times the sum of their members' probabilities after 200 steps and 5.2 times after 600 steps, against 0.66 for the
base model, and ECE rose from 5.6 to between 26.0 and 27.3. The shift occurs under every choice of training questions, so we attribute it
to the reward. The log-ratio reward of main-paper Eq.~(3) has expectation $-\KL(p_f\Vert q_f)$, which is maximized only at
$p_f=q_f$.

\renewcommand{\dbltopfraction}{0.95}   %
\renewcommand{\dblfloatpagefraction}{0.85}   %
\section{Additional analyses}
\label{sup:extra}
\SuppTabFlash
\SuppTabDeltaFlash
\SuppTabSizes
\SuppTabDeltaSizes
\paragraph{Accuracy against partition violation.} \Cref{fig:sup_scatter} plots accuracy against partition violation for
every trained method of main-paper Table 1 and of \cref{tab:main_flash}.
\paragraph{Order-sensitivity floor.} The reference $P$ averages three orders, while each regrouped question is asked in one order, so a model that satisfies main-paper Eq.~(1) exactly but is sensitive to order still has a partition violation. The mean TV distance between a single-order
answer and $P$ estimates this order-sensitivity floor: 2.4 for \method and 3.3 for the base model on d3-lite, against partition
violations of 5.8 and 14.8 (\cref{tab:floor}).

\paragraph{Thresholded decisions.} A downstream system that thresholds the probability of a merged option can make a different
decision from the one implied by the original question. At a threshold of 0.5, the two decisions differ for 18.6\% of merged
options with the base model and 7.4\% with \method{} (\cref{tab:floor}).

\paragraph{Per-benchmark results and generative VLMs.} \method's partition violation is lower than that of the base model and LLM-as-Jev$^\dagger$ on each of the seven
benchmarks (\cref{tab:bench}). \Cref{tab:vlm} applies the diagnosis of main-paper Sec.~3 to generative VLMs that answer with an option letter.

\section{Full results with confidence intervals}
\label{sup:tables}
This section reports the results of main-paper Tables~1--3 and Secs.~3 and 5.1--5.5 with 95\% intervals. Intervals
over seeds are Student-$t$ intervals. Paired differences from \method use an item bootstrap with the same resamples for all
methods, after averaging each method's per-item values over its seeds; these intervals reflect item sampling, not seed variation.
\SuppTabAblLite
\SuppTabAbl

\renewcommand{\dblfloatpagefraction}{0.99}
\SuppTabConf
\subsection{d3-lite}
\label{sup:tab-lite}
\Cref{supt:lite} gives all metrics of the methods in main-paper Tables~1 and~2 and Fig.~4, and \cref{supt:delta-lite}
their paired differences from \method. Every other method has a higher partition violation than \method, and no interval
of these differences includes zero; the smallest difference is that of LLM-as-Jev$^\dagger$ (3.5 points, interval
[3.3, 3.8]). In accuracy, SFT$^\dagger$ and LLM-as-Jev$^\dagger$ exceed \method, whereas PA-GRPO, ACRE$^\dagger$,
PA-GRPO$^\dagger$ and the training-free corrections are less accurate; the intervals of the differences of SFT, LLM-as-Jev and ACRE as
published include zero. On the original questions alone (Orig.\ acc.), five baselines are more accurate than \method, with
intervals that exclude zero: SFT (+2.5), LLM-as-Jev (+2.3), ACRE (+1.5), SFT$^\dagger$ (+3.1) and LLM-as-Jev$^\dagger$
(+2.3). \method's Orig.\ acc.\ rises with the label weight: averaged over seeds, it is 61.1, 64.6 and 66.3 with $w=0$, 0.3
and 0.5 (other metrics in \cref{supt:abl-lite}).

\SuppTabGap
\subsection{d3-flash}
\label{sup:tab-flash}
\Cref{tab:main_flash,supt:delta-flash} give these results for d3-flash, where SFT and LLM-as-Jev were also trained on our
regrouped questions. Every other method again has a higher partition violation than \method; the smallest difference is
that of SFT$^\dagger$ (2.4 points, interval [1.9, 3.0]). SFT, SFT$^\dagger$ and LLM-as-Jev$^\dagger$ are significantly more accurate
than \method; ACRE and the training-free corrections are significantly less accurate. On the original questions, SFT (+3.7), LLM-as-Jev
(+2.1), SFT$^\dagger$ (+4.1) and LLM-as-Jev$^\dagger$ (+2.0) are more accurate than \method, with intervals that exclude
zero.

\subsection{Other model sizes}
\label{sup:tab-sizes}
\Cref{supt:sizes,supt:delta-sizes} give the results of the base model and \method at d3-nano, d3-mini and 26B
(main-paper Sec.~5.3 and Fig.~5) and of SFT$^\dagger$ at 26B. At each size, \method has a lower partition violation and a higher accuracy than the base
model, and no interval of these differences includes zero. ECE falls at d3-nano, does not change significantly at d3-mini,
and rises at 26B. At 26B, SFT$^\dagger$ is 2.0 points more accurate than \method, and its partition violation
is 2.2 points higher.

\subsection{Ablations}
\label{sup:tab-abl}
\Cref{supt:abl-lite} adds intervals to the rows of main-paper Table~3, and \cref{supt:abl} gives the ablations run at the
other sizes. On d3-lite, only label-weight changes alter the partition violation by more than seed variation: by $+6.0$ points
with $w=1$, $-0.3$ with $w=0$, $+0.6$ with $w=0.5$ and $+0.7$ with $w$ from 0 to 0.6. The KL anchor changes no metric beyond seed variation
on d3-lite, d3-mini and d3-flash. On d3-nano (2 seeds), the anchor lowers accuracy by 0.8 points and raises the partition
violation on hard items by 0.6 points; at 26B, it raises the order TV by 0.6 points. On d3-flash, the exact KL is 0.3 points more accurate than \method, and its other differences are
within seed variation.

\subsection{Partition violation at matched confidence}
\label{sup:tab-conf}
\Cref{supt:conf} extends the confidence control of main-paper Sec.~5.5 to all sizes. In every bin and at every size,
\method has the lowest partition violation of the methods shown. \method's mean top probability is within 1.1 points of
the base model's at every size, whereas SFT and SFT$^\dagger$ raise it to between 82.8 and 91.7.

\subsection{Base models}
\label{sup:tab-gap}
\Cref{supt:gap} gives the partition violation and the additivity error of the base models (main-paper Sec.~3 and
Fig.~2). At every size, the mean additivity error is negative, and its interval excludes zero.

\renewcommand{\dbltopfraction}{0.6}   %
\section{Comparison with parsing merged options}
\label{sup:parse}
\SuppTabParse
\SuppTabParseDelta
Software can avoid under-additivity by asking the original question and adding the answers, if it knows which original
options a merged option contains (main-paper Sec.~1). Because every merged option of our evaluation set lists its members word for word,
we can compare this alternative with \method{} on this set. The alternative needs a parser for
each wording, option texts that list the members, and access to the original question; with 3 orders, it needs three
forward passes per question. It does not apply to add questions. \method{} answers each question in one forward pass and needs neither a
parser nor the original question at inference. Merged options that name a broader event without listing their members,
such as ``left half'', are outside the scope of this paper (main-paper Sec.~6).

\paragraph{Parse-and-sum.} A rule-based parser splits the text of each presented option at the separators of the five
wordings (``X or Y'' and the four unseen wordings of \cref{sup:data}) and matches each part to the original option with the
same text. The original question is then asked, and each presented option receives the sum of its members' probabilities.
With 1 order, the original question is asked in one order instead of the regrouped question (one forward
pass). With 3 orders, the original question is asked in three orders (three forward passes) and the answers are averaged;
this average also replaces the model's answer to each original question. An added option is not among the original options, so add questions keep the model's own answer.
Of the 5{,}100 regrouped questions other than reorderings, parse-and-sum applies to \NPsApplicable\%. The parser recovers the members of all \NPsMerged{}
merged options in the ``X or Y'' wording and of at least \NPsWordMin\% in each unseen wording. The parser fails only on an option
whose own text contains a comma. A parser with only the ``X or Y'' rule recovers \NPsOrOnlyMerged{} merged options in the
unseen wordings, so each new wording needs its own rule. Where parsing fails, parse-and-sum keeps the model's own answer.
\method{}, trained on ``X or Y'' only, needs no new rule for the unseen wordings (main-paper Table~4).

\paragraph{Results.} Parse-and-sum improves the accuracy, ECE and Brier score of the base model (\cref{supt:parse}). With 3 orders, accuracy rises from
\NPsLiteBaseAcc{} to \NPsLiteBasePsAcc{} and ECE falls from \NPsLiteBaseEce{} to \NPsLiteBasePsEce{} on d3-lite; on d3-flash
accuracy rises from \NPsFlashBaseAcc{} to \NPsFlashBasePsAcc{} and ECE falls from \NPsFlashBaseEce{} to \NPsFlashBasePsEce.
\method is more accurate than the base model with parse-and-sum over 3 orders, by \NPsLiteDAccAbs{} points
\NPsLiteDAccCI{} on d3-lite and \NPsFlashDAccAbs{} \NPsFlashDAccCI{} on d3-flash, and its Brier score is lower on both models
(\cref{supt:parse-delta}). Its ECE is lower on d3-lite (\NPsLiteDEce{} \NPsLiteDEceCI); on d3-flash the interval of the difference
(\NPsFlashDEce{} \NPsFlashDEceCI) includes zero. On d3-lite, \method is more accurate on every question type, with the largest
difference on the original questions (\NPsLiteCoreAccOrig{} against \NPsLiteBasePsAccOrig), whose answers parse-and-sum
reuses for every regrouped question. On d3-flash, the intervals of the accuracy differences on pair merges and splits include zero. Applied to \method, parse-and-sum over 3 orders raises accuracy by
\NPsLiteDCAccAbs{} points \NPsLiteDCAccCI{} on d3-lite and lowers the Brier score on both models, but raises ECE by
\NPsLiteDCEceAbs{} points \NPsLiteDCEceCI{} on d3-lite and \NPsFlashDCEceAbs{} \NPsFlashDCEceCI{} on d3-flash.

With correctly parsed members, parse-and-sum over 3 orders has a partition violation of 0 by construction. Over 1 order, its
partition violation on d3-lite (\NPsLiteBasePsOnePart{} for the base model and \NPsLiteCorePsOnePart{} for \method) is at
most the order floor of \cref{tab:floor}, because the TV distance between two distributions cannot grow when both are
summed over the same blocks. These partition violations describe the procedure rather than the model, so \cref{supt:parse} does not
compare them across rows.

\section{Sampled reward on large answer sets}
\label{sup:large}
\SuppTabLarge
\SuppTabLargeMass
The reward of main-paper Eq.~(3) needs only the probabilities of the answers drawn. We test it on a model whose answer
set contains every token string: Qwen3-VL-2B-Instruct answers by writing the text of an option. The prompt contains the images, the
question, one option text per line and the line ``Answer with the text of one option from the list above, copied
exactly.'' For a presented option $B$, let $s_B$ be its text followed by the end-of-turn token. The model's probability $p(s)$ of
a string $s$ is the product of its token probabilities, and the frozen model's is $p_0(s)$. $P_0$ is the frozen model's
probability of the original option strings, renormalized over the options and averaged over the two training orders, and
$q_f$ is the target of main-paper Eq.~(2) with $w=0.3$ and $\epsilon=0.05$. Over strings, the target is
\begin{equation*}
q(s) = 0.95\,q_f(B)\,[s=s_B] + 0.05\,p_0(s),
\end{equation*}
which sums to one over all strings. \method draws $K=16$ strings of at most 48 tokens per question from the model at temperature 1
and uses the reward $\log q(s)-\log p(s)$, both terms computed by teacher forcing. A string truncated at 48 tokens is
rewarded as the event that the answer starts with it. The KL divergence to $q$ cannot be computed, because it sums over all
strings. The comparison method minimizes the cross-entropy $-\sum_{B}q_f(B)\log p(s_B)$ over the presented option strings;
this loss is computable because $q_f$ is supported on these strings, and it omits the $0.05\,p_0$ part of $q$. Both methods use
3 seeds and the training items, steps, LoRA settings and learning rate of \cref{sup:impl}. For evaluation, the probabilities
$p(s_B)$ of the presented options are renormalized over these options, and the metrics of the main paper are computed from
these renormalized probabilities. For the base model, strings sampled from the model equal an option string as often as the summed probability of the
option strings predicts: on the 825 questions of every tenth evaluation item, 78.2\% of 13{,}200 samples
are option strings, against a summed probability of 78.2\%.

\paragraph{Results.} The base model assigns \NJBaseMass\% of its probability to the presented option strings, but only
\NJBaseMassPair\% on pair merges and \NJBaseMassPart\% on splits (\cref{supt:large-mass}). In the samples we inspected, it
writes the text of a single member instead, such as ``8'' for the option ``5 or 8''. Both trained models assign 97 to 99\% to
the option strings for every question type (means over seeds), so after training the metrics cover almost all of the model's probability, and part
of the drop in partition violation from the base model comes from this shift. With the sampled reward, \method lowers the
partition violation from \NJBasePart{} to \NJCorePart, its part above the order floor from \NJBaseAbove{} to
\NJCoreAbove, and ECE from \NJBaseEce{} to \NJCoreEce, and raises accuracy from \NJBaseAcc{} to \NJCoreAcc{}
(\cref{supt:large}). The cross-entropy on the option strings is better on every metric: \method's partition violation
is higher by \NJDCoreCePart{} points \NJDCoreCePartCI, its ECE by \NJDCoreCeEce{} \NJDCoreCeEceCI, and its accuracy difference
is \NJDCoreCeAcc{} \NJDCoreCeAccCI. The sampled reward thus trained the model over all token strings, a set too large to sum over, using only
the drawn strings and the option strings. An exact alternative nevertheless existed in this setting, because the target assigns at least 95\% of its
mass to the option strings, and this alternative was better on every metric. The two methods also differ in their targets, since the
cross-entropy omits the $0.05\,p_0$ part of $q$. We did not test a target whose support cannot be enumerated. These results agree
with main-paper Sec.~5.4: the target, not the sampled estimator, reduces the violation, and where an exact objective is
available it works at least as well.

\section{\method on generative VLMs}
\label{sup:vlmtrain}
\SuppTabVlm
\SuppTabVlmDelta
\begin{figure*}[t]
\centering
\includegraphics[width=\textwidth]{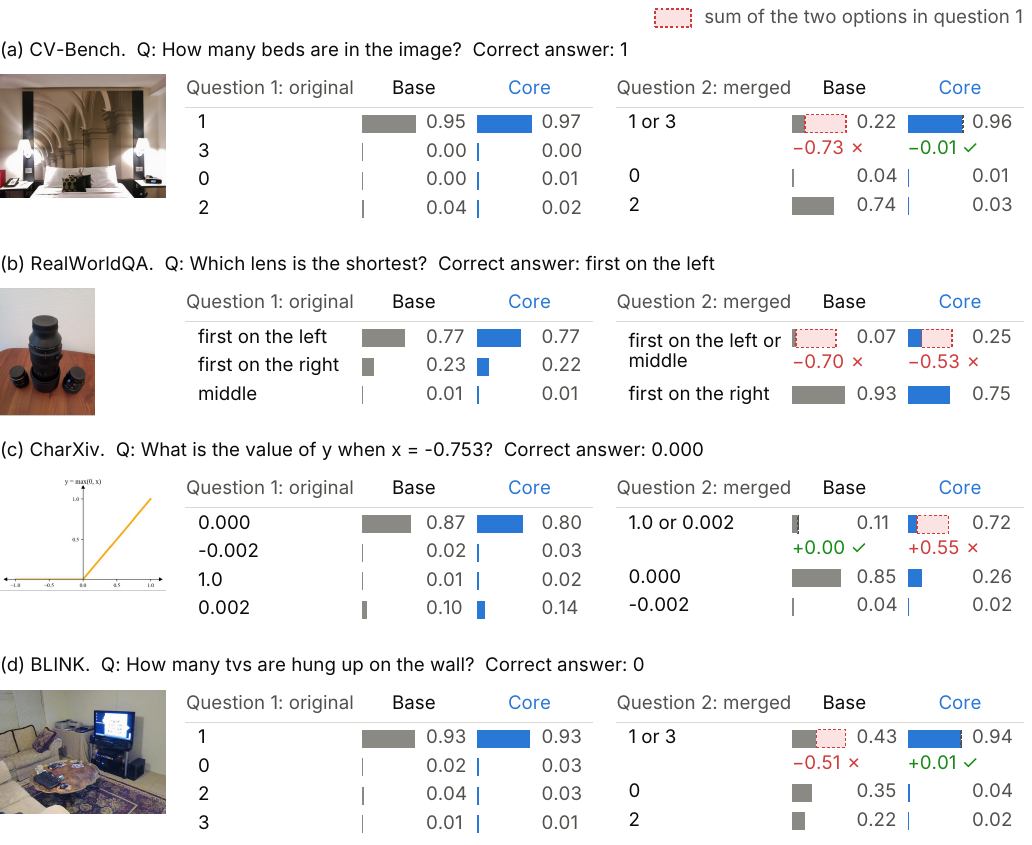}
\vspace{-1.4em}
\caption{\textbf{One success and three failure cases of \method on d3-lite, selected by fixed rules.} Each row shows an
evaluation item, its original question (question 1, averaged over three option orders) and one pair merge
(question 2), answered by the base model and by \method (the seed whose partition violation on question 2 is the median
of five seeds). (a) Largest reduction of the partition violation among pair merges on which the base model is
under-additive and \method is coherent. (b) Most negative additivity error under \method. (c) Most positive additivity
error under \method. (d) Highest probability of a wrong option among pair merges on which \method is coherent and wrong on
question 1 in all seeds. \Cref{sup:qual} gives the rules. Dashed boxes span from the merged option's probability to the sum of its members' probabilities in
question 1; the number below each box is the additivity error, marked $\times$ when its absolute value exceeds 0.1 and
\checkmark{} otherwise. Option texts in (b) are abbreviated.}
\label{fig:sup_qual}
\vspace{-0.6em}
\end{figure*}
We train two generative VLMs, Qwen3-VL-2B-Instruct and Qwen3-VL-8B-Instruct, with \method, SFT and SFT$^\dagger$. The
models answer as in \cref{tab:vlm}: the prompt labels the options with the letters A, B, \dots{} and asks for one letter,
and the answer distribution is the softmax over the logits of the option letters at the first answer token. Training
items, steps, LoRA settings, optimizer, targets and losses are those of \cref{sup:impl,sup:baselines}; LoRA is applied to
all linear layers of the language model, and only the LoRA adapters are trained. All three methods use 3 seeds and the d3-lite learning rate
$3\cdot10^{-5}$, without tuning on the VLMs. The 2B models were trained and evaluated on RTX 4090 GPUs and the 8B models on H100 GPUs. We
re-evaluated both base models with the code and GPUs of the trained models, so their rows in \cref{supt:vlmtrain} differ from
those in \cref{tab:vlm} by at most 0.3 points.

\paragraph{Results.} \Cref{supt:vlmtrain} gives all metrics and \cref{supt:vlmtrain-delta} the paired differences from
\method. On both models, \method has the lowest partition violation of all rows and is less accurate than SFT and
SFT$^\dagger$. \method's partition violation is \NKTwoCorePart{} against
\NKTwoSftDPart{} for SFT$^\dagger$ on the 2B model and \NKEightCorePart{} against \NKEightSftDPart{} on the 8B model. On
the 2B model, \method also has the lowest ECE, order TV and IIA TV, and the intervals of its differences from SFT$^\dagger$
in partition violation, ECE and order TV exclude zero. On the 8B model, the differences between \method and SFT$^\dagger$ are smaller. The partition
violation of SFT$^\dagger$ is higher by \NKDEightSftDPart{} points \NKDEightSftDPartCI, and SFT$^\dagger$ has the lower Brier score
(\NKDEightSftDBrier{} \NKDEightSftDBrierCI). The intervals of the differences in ECE (\NKDEightSftDEce{} \NKDEightSftDEceCI),
in partition violation on the hard items and in order TV include zero.

\method's accuracy (\NKTwoCoreAcc{} and \NKEightCoreAcc) is above that of the base models (\NKTwoBaseAcc{} and
\NKEightBaseAcc). SFT$^\dagger$ exceeds \method by \NKDTwoSftDAcc{} points
\NKDTwoSftDAccCI{} on the 2B model and \NKDEightSftDAcc{} \NKDEightSftDAccCI{} on the 8B model, and SFT exceeds it by
\NKDTwoSftAcc{} \NKDTwoSftAccCI{} and \NKDEightSftAcc{} \NKDEightSftAccCI. On d3-lite, SFT$^\dagger$ also exceeds \method
in accuracy, but SFT does not (\cref{supt:delta-lite}).

\paragraph{Order sensitivity.} The base VLMs are sensitive to the option order (order TV \NKTwoBaseOrder{} and
\NKEightBaseOrder), so a large part of their partition violation is the order floor of \cref{tab:floor}. A model
that satisfies main-paper Eq.~(1) exactly but has the same order sensitivity also has this floor. The part above the floor is
\NKTwoBaseAbove{} of \NKTwoBasePart{} points for the 2B base model and \NKEightBaseAbove{} of \NKEightBasePart{} for the
8B base model. \method lowers both parts: its order floor is \NKTwoCoreFloor{} and \NKEightCoreFloor{} points, and its
partition violation above the floor is \NKTwoCoreAbove{} and \NKEightCoreAbove, the lowest of all rows (last column of
\cref{supt:vlmtrain-delta}). SFT lowers the order floor to \NKTwoSftFloor{} and \NKEightSftFloor{} points, but its
partition violation above the floor, \NKTwoSftAbove{} and \NKEightSftAbove{} points, is close to or above that of the base
models. SFT's mean additivity error, \NKTwoSftGap{} and \NKEightSftGap{} points against \NKTwoBaseGap{} and
\NKEightBaseGap{} for the base models, shows that SFT makes both models more under-additive.

\section{Qualitative examples and failure cases}
\label{sup:qual}
\renewcommand{\dbltopfraction}{0.9}   %
\SuppTabMetrics
\begin{table*}[t]
\centering\small
\setlength{\tabcolsep}{4pt}
\caption{Training and inference cost. Prompts per training item: questions passed through the frozen model without gradient (frozen) and forward and backward through the LoRA adapters (trained), with 4 items per step. The $K$ answers of the RL methods are sampled from the computed distribution and need no further pass. Training time: median over seeds of one 600-step run without the intermediate evaluations, divided by the time of a run with the passes of \method on the same GPU type under the same load. \method takes 29 min on d3-lite (one RTX 4090) and 56 min on d3-flash (one H100). Inference: forward passes per question.}
\label{supt:cost}
\begin{tabularx}{\textwidth}{@{}>{\raggedright\arraybackslash}X*{4}{c}@{}}
\toprule
& Prompts per item & \multicolumn{2}{c}{Training time ratio} & Passes per \\
\cmidrule(lr){3-4}
Method & frozen / trained & d3-lite & d3-flash & question \\
\midrule
\method & 2 / 4 & 1.00 & 1.00 & 1 \\
SFT & 0 / 2 & 0.45 & 0.57 & 1 \\
SFT$^\dagger$ & 0 / 4 & 0.72 & 0.84 & 1 \\
LLM-as-Jev & 2 / 2 & 0.63 & 0.74 & 1 \\
LLM-as-Jev$^\dagger$ & 2 / 4 & 1.11 & 1.00 & 1 \\
ACRE & 2 / 4\textsuperscript{a} & 0.99 & 1.03 & 1 \\
ACRE$^\dagger$ & 2 / 4 & 1.05 & -- & 1 \\
PA-GRPO & 4.9 / 4.9 & 1.45\textsuperscript{b} & 1.30 & 1 \\
PA-GRPO$^\dagger$ & 4.9 / 6.9 & 1.72\textsuperscript{b} & -- & 1 \\
exact $\KL(p_f\Vert q_f)$ & 2 / 4 & 1.00 & 1.01 & 1 \\
cross-entropy to $q_f$ & 2 / 4 & 0.99 & -- & 1 \\
\midrule
Temperature scaling & -- & --\textsuperscript{e} & --\textsuperscript{e} & 1 \\
Sorted options & -- & -- & -- & 1 \\
Order averaging & -- & -- & -- & 2 \\
OBS & -- & -- & -- & $c$ (3.75)\textsuperscript{c} \\
Parse-and-sum, 1 order & -- & -- & -- & 1\textsuperscript{d} \\
Parse-and-sum, 3 orders & -- & -- & -- & 3\textsuperscript{d} \\
\bottomrule
\end{tabularx}
\par\smallskip\raggedright\footnotesize \textsuperscript{a}The ACRE loss uses only the two original orders; our implementation also passes the two regrouped questions of each item, which a standalone implementation of ACRE would skip (2 / 2 prompts per item). \textsuperscript{b}PA-GRPO ran with gradient checkpointing (one extra forward pass in the backward pass), but its d3-lite reference runs on the same GPU did not; on d3-flash, all runs used gradient checkpointing. \textsuperscript{c}OBS asks one yes/no question per presented option; $c$ is the number of presented options, with its mean over the 8{,}250 evaluation questions in parentheses. \textsuperscript{d}For each regrouped question, the original question is asked in 1 or 3 orders instead; add questions keep the model's own answer (1 pass). \textsuperscript{e}The temperature is fitted with one pass over 400 training items.
\end{table*}
\Cref{fig:sup_qual} shows four evaluation items answered by d3-lite before training (Base) and after training with
\method. Each row shows the original question and one pair merge of it. The four pair merges are selected from the 2,100
pair merges of the evaluation set by fixed rules: (a) the largest reduction of the partition violation among pair merges on
which the base model is under-additive and \method is coherent (mean partition violation over seeds at most 0.05); (b) the
most negative additivity error under \method; (c) the most positive additivity error under \method; (d) the highest
probability of a wrong option among pair merges on which \method is coherent and its most probable original option is wrong
in all seeds.
Each rule takes the highest-ranked pair merge from a benchmark not used by an earlier rule, and the item of main-paper
Fig.~1 is excluded. Rankings use the mean over the five seeds of \method; the figure shows the seed whose partition
violation on the shown question is the median of the five.

\paragraph{Success.} In (a), the base model assigns 0.22 to the merged option ``1 or 3'', although its members receive 0.95
in the original question. \method assigns it 0.96, with an absolute additivity error below 0.03 in every seed.

\paragraph{Remaining under-additivity.} In (b), the merged option contains the correct answer, and its members receive
0.75 to 0.78 in the original question under both models. The base model assigns it 0.07 and \method 0.25 (\method's additivity error is $-$0.48 to $-$0.54 over
seeds). Both models answer the original question correctly but choose the unmerged wrong option in the regrouped question.

\paragraph{Over-additivity.} In (c), the base model is coherent (additivity error $+$0.00). \method assigns 0.72 to the merged
option ``1.0 or 0.002'', although its members receive 0.16 in the original question (additivity error $+$0.40 to
$+$0.57 over seeds). \method answers the original question correctly with probability 0.80 but assigns 0.26 to the correct option in the
regrouped question. Over all 2,100 pair merges, the base model and \method{} have mean absolute additivity errors of 0.13 and 0.05, and
1,013 and 305 merges with an error beyond $\pm$0.1 (means over seeds). The base model's large errors are almost all negative (984 below $-$0.1, 29 above 0.1); \method's
are split between 160 below $-$0.1 and 145 above 0.1 (this count ranges from 78 to 198 over seeds), so \method{} over-corrects some merged
options.

\paragraph{Coherent but wrong.} In (d), the television stands on a cabinet, and the correct answer is 0. Both models assign
0.93 to ``1'' in the original question. The base model assigns 0.43 to ``1 or 3'' and \method
0.94, while its members receive 0.94 under both models. Coherence constrains how a model's probabilities relate across regroupings of the options; it
does not make the answer to the original question correct.

\section{Robustness of the metrics}
\label{sup:metrics}
Two metric choices could affect the conclusions of the main paper: ECE uses 15 bins of equal width, and the partition
violation pools pair merges and splits. \Cref{supt:metrics} varies both. With 10 or 20 bins of equal width, or with 15
bins that hold equal numbers of questions (adaptive ECE, the equal-count binning of~\cite{nixon2019measuring} applied to the
most probable option), no trained baseline changes its rank relative to \method. With these binnings, as with 15 bins, the difference
between \method and the trained baseline with the lowest ECE has an interval that includes zero.
For the training-free corrections of main-paper Table~2, the comparison depends on the binning: temperature
scaling on d3-lite and order averaging on d3-flash have a lower ECE than \method with some binnings and a higher one with
others, and all eight of these differences have 95\% intervals that include zero. Fitted on the
evaluation questions themselves, which favors it, temperature scaling has ECE 1.9 (d3-lite) and 2.4 (d3-flash) and partition violation 14.6 on both models;
fitted on the training items as in main-paper Table~2, it has ECE 1.9 and 2.7 and partition violation 14.6 and 14.3.

On both models, \method has the lowest partition violation of all methods in \cref{supt:metrics} on pair merges and on splits,
and for each regrouping type its difference from the best trained baseline has an interval that excludes zero.

\section{Computational cost}
\label{sup:cost}
\method adds no inference cost: like every trained baseline, it answers a question with one forward pass. Order averaging needs 2 passes, OBS needs one per presented option (3.75 on average on our evaluation set), and parse-and-sum needs a parser,
access to the original question and 1 pass per regrouped question with 1 order or 3 passes with 3 orders. On d3-flash (one H100), evaluating the 8{,}250 questions, including model loading, takes 12 min with one pass per question (measured with sorted options), 22 min with order averaging and 73 min with OBS. In training, \method passes the two training orders of each item through the frozen model without gradient and all four questions of the item through the adapters. A d3-flash run of \method trains for 56 min on one H100 without the intermediate evaluations (68--69 min with them, \cref{sup:impl}); LLM-as-Jev$^\dagger$ and ACRE take about as long (time ratios 1.00 and 1.03). SFT and LLM-as-Jev, trained on the original questions only, take 0.57 and 0.74 times as long, and PA-GRPO, which passes 4.9 option orders per item on average, takes 1.30 times as long. Training and inference costs of all methods are in \cref{supt:cost}.
\FloatBarrier

\end{document}